\PassOptionsToPackage{sort&compress}{natbib}
\documentclass{baai}
\usepackage{adjustbox,xspace,xcolor}
\usepackage{multirow,makecell}
\usepackage{float}
\usepackage{algorithm,algorithmic}
\usepackage{needspace}
\usepackage{amssymb}
\usepackage{marvosym}
\usepackage{pifont}
\usepackage{placeins}
\usepackage{colortbl}
\usepackage{fancyhdr}
\usepackage[nameinlink,noabbrev]{cleveref}
\crefname{section}{Section}{Sections}
\Crefname{section}{Section}{Sections}
\crefname{table}{Table}{Tables}
\Crefname{table}{Table}{Tables}
\crefname{figure}{Figure}{Figures}
\Crefname{figure}{Figure}{Figures}
\crefname{equation}{Equation}{Equations}
\Crefname{equation}{Equation}{Equations}
\Crefname{algorithm}{Algorithm}{Algorithms}
\setcitestyle{numbers,square,comma}
\title{Innata-0: A Native Embodied World System with Model-Agent Co-Design}
\author{Innata Team}
\reportaffiliation{Beijing Academy of Artificial Intelligence}
\reporturl{https://baai-eai.github.io/innata-0}
\reportshorttitle{Innata-0 Technical Report}
\reportlogo{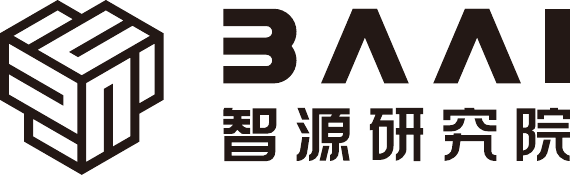}
\reportdate{November 2026}
\reportabstract{The physical world is open-ended, dynamic, and partially observable.
While foundation models have advanced embodied intelligence, standalone models struggle with
long-horizon execution, and simply integrating models with agents may lead to inconsistencies
between planning and physical control.
To support reliable and adaptive behaviors in the physical world, we introduce \textbf{Innata-0}, a
native embodied world system. Innata-0 jointly designs an embodied world-action model and an agent
to combine their complementary strengths.
The model, \textbf{Innata-0-WAM}, adopts a tri-tower Mixture-of-Transformers architecture that
integrates understanding, prediction, and action generation.
The agent, \textbf{Innata-0-Agent}, coordinates long-horizon execution through memory, verification,
and recovery.
A co-design protocol is further proposed to align model training with agent execution and leverage
validated experiences to support subsequent adaptation and model improvement.
Extensive evaluations demonstrate strong embodied capabilities across understanding, world
prediction, and robotic manipulation.
Innata-0 represents our first step toward physical intelligence, and we see it as the beginning of a
much longer journey.
\vskip4ex
\label{sec:codesign_motivation}
\begin{figure}[h]
  \centering
  \includegraphics[width=\linewidth]{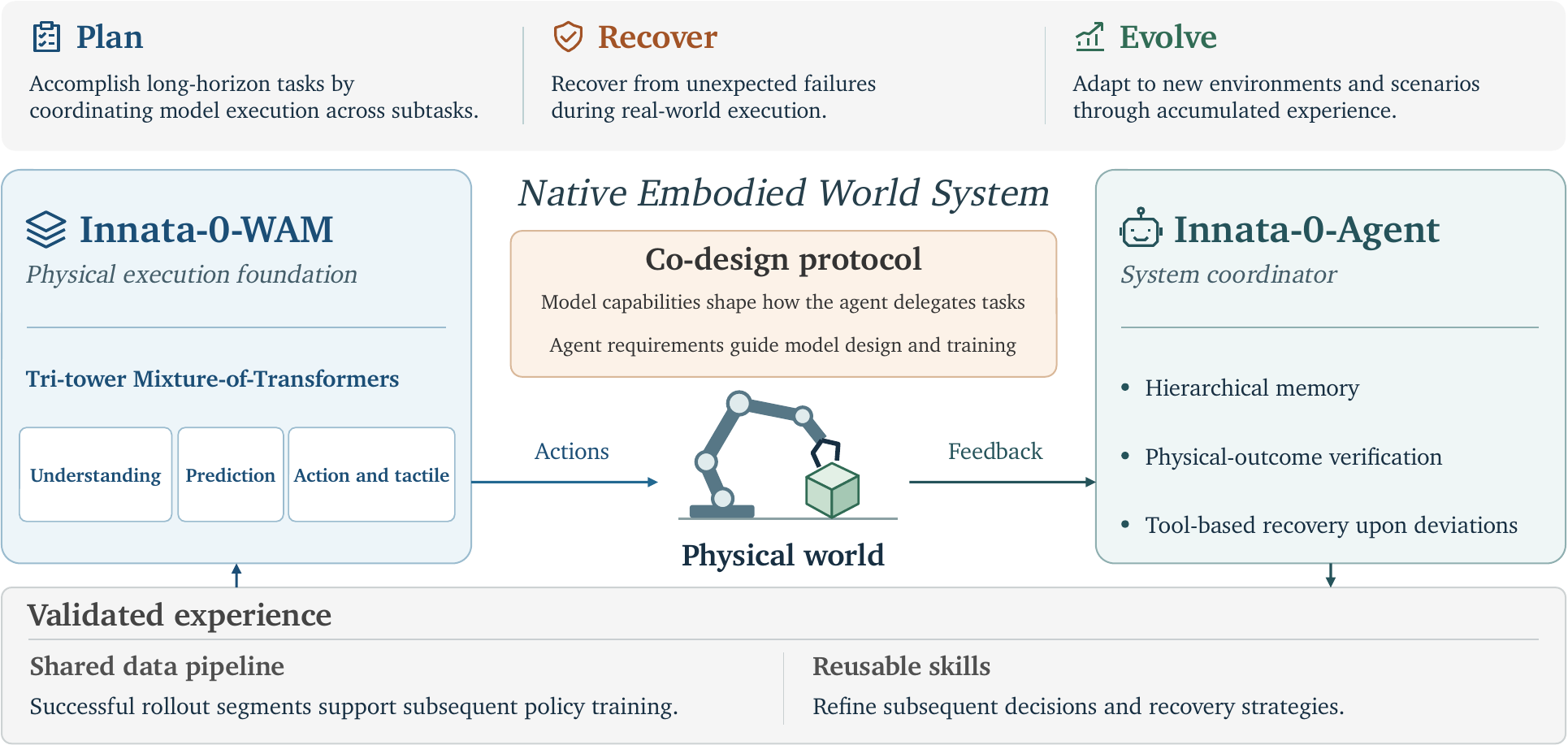}
  \vskip1ex
  \caption{\textbf{Native model–agent co-design in Innata-0.} The embodied model and agent play complementary roles in realizing Plan, Recover, and Evolve. A shared interaction protocol aligns model capabilities with agent requirements. Successful rollout segments support subsequent policy training, while verified interactions and experiences are consolidated into reusable skills.}
  \label{fig:native_codesign}
\end{figure}
}
\AtBeginDocument{\hypersetup{pdfauthor={Innata Team, BAAI}}}

\begin{document}

\reportfrontmatter
\section{Native Model-Agent Co-Design}
\label{sec:introduction}
The physical world is open-ended, partially observable, and dynamically evolving, where actions have
uncertain consequences and failures may incur irreversible costs.
While foundation models have substantially advanced embodied intelligence, standalone foundation
models still struggle to sustain reliable behavior over long horizons, particularly when execution
requires self-correction and adaptation.
Extending their capability boundaries therefore requires coupling native model capabilities with
persistent agentic reasoning~\citep{yao2023react,schick2023toolformer},
memory~\citep{nvidia2026gr00t2,lu2026aspire}, and
interaction~\citep{zitkovich2023rt2,kim2024openvla,gemini2025robotics}.
However, simply integrating models and agents may not fully realize their combined potential.
Inconsistencies between planning and control may lead to unreliable physical execution and limited
adaptation from accumulated experience.
This motivates a shift from a \emph{standalone embodied model} toward a
\emph{native embodied world system}.

To this end, we characterize \emph{native} through three core capabilities:
\emph{Plan} lays the foundation for the system to accomplish long-horizon tasks by coordinating model execution across subtasks;
\emph{Recover} further bolsters the system, enabling it to handle uncertainties and recover from unexpected failures during real-world execution; and
\emph{Evolve} extends the system to continuously adapt to new environments and scenarios through accumulated experience.
To jointly realize these capabilities, we introduce \textbf{Innata-0}, a model-agent co-designed
system that advances \emph{native embodied intelligence}.
As illustrated in \figref{fig:native_codesign}, our co-design establishes complementary roles for
the embodied model and agent, allowing the model to fully leverage its native capabilities while the
agent extends its effective operating scope.

On the model side, \textbf{Innata-0-WAM} serves as the physical execution foundation of the system.
We adopt a tri-tower Mixture-of-Transformers architecture~\citep{liang2024mixture0of0transformers0},
which provides understanding, prediction, and action generation capabilities.
The model is conditioned on the complete instruction, active subtask, and verified task history. By
incorporating current- and history-plan annotations during training, Innata-0-WAM aligns its learned
conditioning interface with the agent's evolving representation of task progress.

On the agent side, \textbf{Innata-0-Agent} serves as the system coordinator, supporting long-horizon
planning, failure recovery, and continual evolution.
It orchestrates model execution across subtasks while maintaining hierarchical memory.
Grounded in observations and execution feedback, the agent verifies against intended physical outcomes and
invokes tool-based recovery upon deviations.
Verified interactions and experiences are consolidated into reusable skills that refine subsequent
decisions and recovery strategies.
Innata-0-Agent aligns its execution logic with the model's capabilities, enabling coordinated
execution, outcome verification, and experience-driven adaptation.

Realizing \emph{Plan}, \emph{Recover}, and \emph{Evolve} requires the model and agent to work
together and play complementary roles.
We therefore establish a co-design protocol that enables them to operate in concert, leveraging
their respective strengths while guiding their subsequent design.
Crucially, this protocol operates in both directions: the model's execution capabilities shape how the
agent organizes and delegates tasks, while the agent's requirements for planning, recovery, and
adaptation guide model design and training.
Our protocol aligns the model's training data with the agent's task instructions, subtask plans, and
recovery commands, allowing complex tasks to be carried out through coordinated execution.
The agent, in turn, validates execution outcomes and consolidates experiences to refine its task
commands and reusable recovery strategies. It also feeds successful rollout segments, including
corrective actions and post-recovery continuations, into the shared data pipeline for subsequent
policy training.
These principles guide our data construction, model training, and agent design, laying the
foundation of \emph{native embodied intelligence}.

We evaluate the physical capabilities underlying Innata-0 and their use in closed-loop task
execution. Among the compared models, the understanding tower ranks first on nine benchmarks and within the top two on fourteen. On WorldArena~2.0 Track~1~\citep{li2026worldarena2}, Innata-0 achieves an
EWM score of 67.63, the highest among the open-source models. Policy evaluations on
RoboCasa365~\citep{robocasa365} and RoboDojo-Sim~\citep{chen2026robodojo} examine manipulation and
task completion.
On RoboCasa365, Innata-0 achieves 60.2\% overall success and ranks first on Atomic-Seen tasks with
an 88.4\% success rate.
These evaluations characterize the model's grounding, prediction, and control capabilities, as well as its
use within an agent-guided execution loop.

To conclude, we propose \textbf{Innata-0}, a native embodied world system that extends physical
intelligence beyond individual task execution. Our approach jointly designs the embodied model and agent to
leverage their complementary strengths. A shared interaction protocol aligns model capabilities with
agent requirements and guides data construction, model training, and agent design. Together, these
designs enable long-horizon planning, failure recovery, and continual evolution. Our system
co-design establishes a basis for studying how aligned task context, physical-outcome verification,
and validated experience contribute to reliable open-world interaction.

\section{Embodied Data Construction}
\subsection{Overview}
Raw visual, linguistic, action, and tactile data contain rich knowledge about objects, interactions,
and physical processes, yet their existing annotations make only part of this knowledge explicit.
Following the co-design protocol, we make task goals, subtask relationships, and completion evidence
explicit, while preserving the links between corrective behavior and subsequent task execution.
We therefore develop \emph{Task-Oriented Knowledge Refinement}, a unified data construction pipeline
that extracts shared annotations from heterogeneous sources and prepares supervision tailored to
each learning objective and training stage.
Each source can support multiple training tasks wherever the necessary observations and annotations
are available.
The resulting task annotations align the model's training context with the agent's subtask
objectives for \emph{Plan}.
Corrective trajectories and task continuations provide supervision for \emph{Recover}, while
validated rollout segments enter the same pipeline for subsequent policy training, supporting
\emph{Evolve} through experience reuse.

We assemble our initial source pool mainly from open-source datasets and web-accessible corpora,
including image--text pairs and other vision--language corpora, general videos, egocentric human
videos, real-robot recordings, and robot simulation data.
Video provides the main source of physical experience, complemented by image--text supervision,
native state and action channels, and tactile observations.
Together, these sources cover diverse scenes, objects, embodiments, and manipulation tasks.

\begin{figure}[t]
  \centering
  \includegraphics[width=\linewidth]{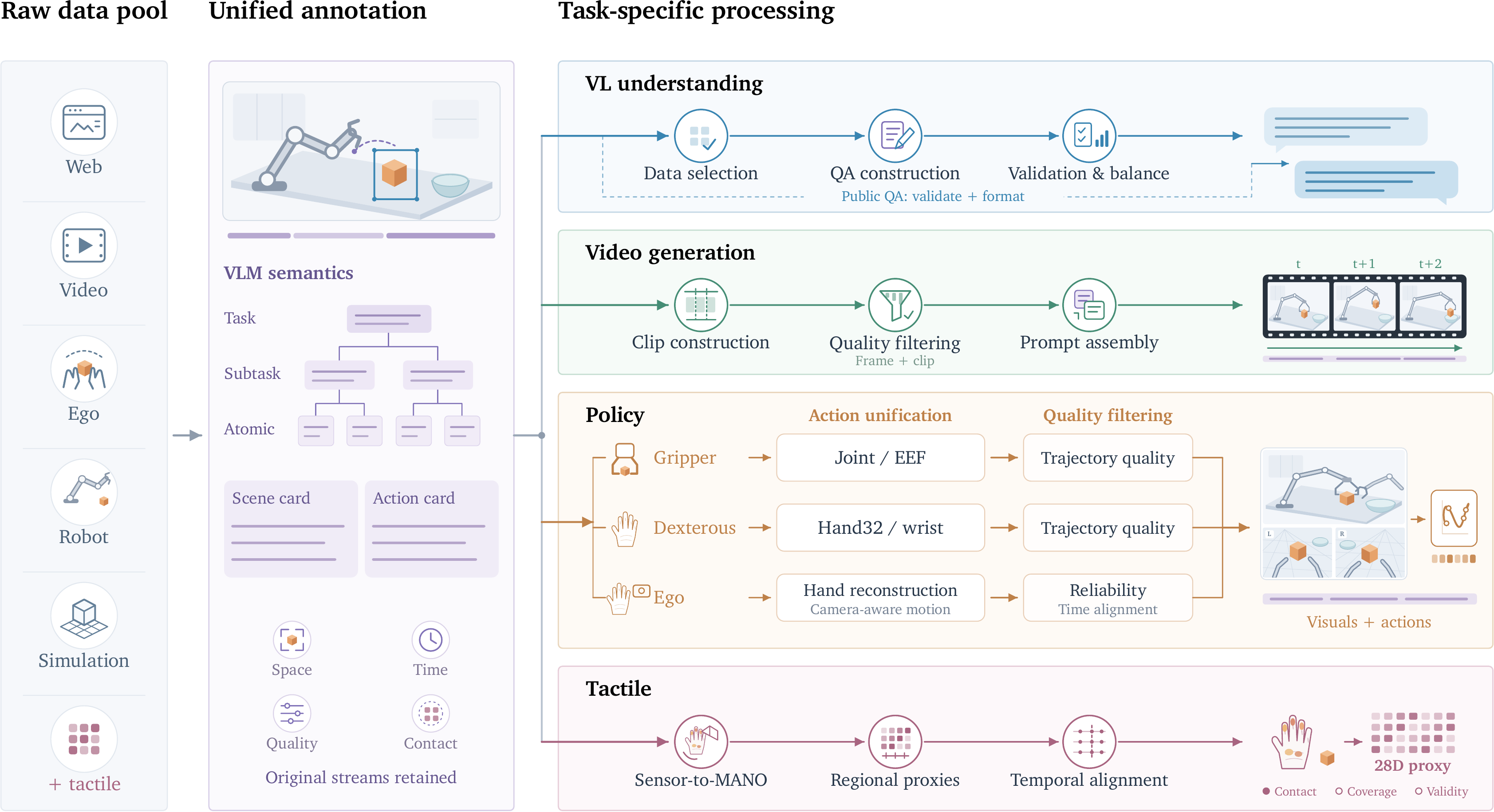}
  \caption{\textbf{Task-oriented knowledge refinement.}
    Heterogeneous sources undergo unified annotation without sample filtering.
    Task-specific pipelines reuse the shared annotations to prepare supervision for vision--language understanding, video generation, policy training, and tactile generation, applying their own quality criteria.
    A source may support multiple tasks according to its available observations and annotations.}
  \label{fig:embodied_data_pipeline}
\end{figure}

\textbf{The knowledge available in a source can exceed the supervision made explicit by its original annotations.} A video may record objects, spatial relationships, intermediate goals, and state changes while its caption names only the overall activity.
Relevant evidence may also occupy only a brief interval within a longer recording.
Each learning objective requires a different view of this information.
For example, the same video of putting a cup into a cabinet can support a grounding question about
the cup, an instruction-conditioned prediction of its placement, and an action-learning example when
corresponding recorded or reconstructed trajectories are available. When synchronized tactile
observations are available, the same interaction can also support prediction of contact and loading
during grasping or placement.
Constructing these views requires us to locate the event, identify its participants and task
structure, and align the context and target at the appropriate granularity.
A frame may suffice for a spatial question, whereas visual prediction requires a coherent transition
and action prediction additionally requires aligned motion supervision.
\textbf{We therefore make the relevant knowledge explicit in the form each objective needs}, while retaining the context needed to interpret it.
This lets one experience contribute to several learning tasks, with each example focusing on a
specific capability.

Figure~\ref{fig:embodied_data_pipeline} summarizes \emph{Task-Oriented Knowledge Refinement} in
three stages: source acquisition, unified annotation, and task-specific data preparation.
The unified annotation stage performs semantic interpretation once per sample to create a shared
record of hierarchical task labels, event boundaries, and scene and action descriptions.
It preserves the original observations and available state, action, and tactile streams, together
with spatial, contact, and quality information, without filtering samples.

The vision--language pipeline uses shared spatial and task annotations to select observations and
construct instruction--target pairs through task-specific rewriting, followed by validation and
distribution balancing.
Existing public QA is validated and reformatted without regeneration.
The video-generation pipeline uses shared temporal annotations for clip construction, applies frame-
and clip-level quality criteria for each training stage, and assembles descriptive or instructional
prompts from shared scene and action descriptions.
The policy pipeline pairs native or reconstructed motion with shared instructions and task
intervals, unifies action semantics, and uses task, outcome, and quality annotations for trajectory
filtering and multi-view assembly.
For eligible human videos, this includes hand reconstruction and camera-aware motion alignment.
The tactile pipeline combines sensor observations with shared contact, coverage, and validity
records, followed by sensor-to-MANO alignment, regional proxy construction, and temporal alignment
with visual and motion streams.
One source can thus yield multiple training forms, depending on its available observations and
supervision.

\subsection{Unified Information Extraction and Annotation}

We attach a shared annotation record to each source, preserving its original timestamps, camera
views, source metadata, native state and action channels, and tactile observations where available.
The record combines source-level context with frame-level and interval-level annotations.
We prompt Qwen3.8-Flash-Next~\cite{qwen3.8flashnext} once per sample to produce the applicable
semantic, relational, temporal, and perceptual-quality fields in a single structured response.
Specialized perception models add dense masks to this record, alongside source-provided depth and
camera calibration.
Thus, one VLM pass supplies a broad description of the experience, while dedicated visual modules
recover spatial measurements and available tactile observations provide contact evidence for
downstream tasks.

\noindent\textbf{Spatial and Geometric Annotation.}
We combine specialized perception models with source-provided depth and camera calibration to obtain
object localization and spatial information.

\begin{itemize}
  \item \textbf{Detection, segmentation, and tracking.}
  We use Grounding DINO~\cite{liu2024groundingdino} for object detection and SAM~3.1~\cite{carion2025sam3,meta2026sam31} for instance segmentation and video tracking. VLMs provide additional object detection and localization for data requiring further annotation. We retain boxes, masks, prediction scores where available, and cross-frame object identities linked to the shared entity record.
  \item \textbf{Depth and 3D structure.}
  We use source-provided depth-camera measurements, together with available camera intrinsics and extrinsics, to obtain spatial inputs and supervision signals.
\end{itemize}

\phantomsection\label{sec:video-task-semantics}
\noindent\textbf{Semantic and Temporal Annotation.}
The shared VLM response describes scenes, interactions, and task structure through the following
groups.

\begin{itemize}
  \item \textbf{Scene and entity context.}
  We annotate scene categories, activity settings, workspace types, and viewpoint tags, including egocentric, external, and robot-mounted views.
  Entity records identify participating humans or robots, relevant objects and tools, observable object attributes, and their initial states.
  Scene tags describe the environment at multiple levels, such as an indoor setting, a kitchen, and a countertop workspace.
  We retain both categorical tags and concise natural-language descriptions.

  \item \textbf{Object relations and interaction states.}
  The shared VLM response explicitly annotates object--object and actor--object relationships, complementing localized entities with language-grounded spatial descriptions~\cite{bai2025qwen30vl}.
  These include directional relations such as left/right, above/below, and in front of/behind; topological relations such as inside, on, and supported by; and visible interaction states such as approaching, holding, touching, or releasing.
  Each relation identifies its participants and the frame or interval in which it holds.
  We also record changes in these relations, such as an object moving from a supporting surface into a container, and retain the viewpoint or coordinate context of each spatial description.

  \item \textbf{Hierarchical tasks and verifiable subgoals.}
  We annotate activity families, overall tasks, subtasks, and atomic goals at the granularity supported by each recording.
  Each subtask identifies the acting entity, relevant objects, and intended state change, together with a concise instruction and an observable condition for judging completion.
  Links to the parent task and temporal ordering preserve its role in the overall activity.
  These annotations provide explicit, actor-centered units for execution and progress assessment.

  \item \textbf{Temporal boundaries and execution stages.}
  We annotate the start and end of tasks, subtasks, and atomic intervals, together with their ordering and execution stages.
  For selected robot demonstrations and egocentric recordings, the shared pass groups movements that serve one manipulation goal and marks a new interval when that goal changes.
  Stage labels describe preparation, approach, interaction, and completion or termination where these phases are visible.
  We also identify pauses, repeated attempts, and transitions between goals, and associate observed state changes with the corresponding intervals.

  \item \textbf{Action types and motion characteristics.}
  We assign action-type labels such as reaching, grasping, lifting, transporting, placing, pushing, pulling, rotating, inserting, opening, closing, pouring, and releasing.
  Task labels describe the intended goal; action labels describe the motion or interaction used to pursue it.
  For each interval, we annotate the active hand or end effector, single-arm or bimanual execution, motion direction, and apparent motion magnitude.
  Magnitude tags describe qualitative amplitude, such as small, moderate, or large motion, separately for the manipulator and the manipulated object.
  We distinguish these visual assessments from metric displacement or joint-motion measurements available in native trajectories, and describe camera or viewpoint motion separately.

  \item \textbf{Task outcomes, failures, and recovery.}
  We annotate task and subtask outcomes with the visible state changes supporting them, recording partial completion or an undetermined result when the footage is incomplete.
  For recovery events, we link the preceding unsuccessful attempt to its corrective interval and describe the corrective behavior, such as regrasping a slipped object, realigning an insertion, or retrying an opening motion.
  We record whether the correction restores the state needed to continue the original task and, where visible, identify the subtask pursued afterward.
  The outcomes of the initial attempt, recovery, and subsequent execution remain distinct.
  Native success flags are retained alongside video-based assessments.
\end{itemize}

\noindent\textbf{Static and Dynamic Description Cards.}
We organize the scene and action descriptions from the same structured response into \emph{static}
and \emph{dynamic cards}.
When a recording stays in one setting, its clips share a static card describing the workspace,
lighting, and objects that remain in place; when the setting varies, we retain a scene description
for each clip.
Each clip has a dynamic card containing a brief action summary and a detailed description of its
participants, movement, and state changes.
For continuous actions, the description follows the movement; for discrete changes, it records the
initial state, action, and visible outcome.
In egocentric footage, we distinguish viewpoint changes from hand and object motion.
The dynamic card also includes an instruction when a manipulation is visible.
Figure~\ref{fig:caption-cards} in the appendix shows how three clips from one recording share a
static card while retaining distinct action descriptions.

\noindent\textbf{Visual Quality, Visibility, and Temporal Integrity.}
We retain technical attributes and perceptual assessments in separate groups.

\begin{itemize}
  \item \textbf{Recording attributes.}
  We extract resolution, aspect ratio, frame rate, duration, codec, and bitrate from source metadata.
  We retain the correspondence between derived clips and original frame timestamps.

  \item \textbf{Perceptual quality and temporal continuity.}
  The shared VLM response assesses defocus and motion blur, exposure, lighting, compression artifacts, camera shake, and jitter.
  It also identifies shot changes, apparent freezes or interruptions, abrupt viewpoint changes, and discontinuities in the visible action.
  We associate localized quality issues with their affected intervals.

  \item \textbf{Embodiment and object visibility.}
  We explicitly annotate whether the robot or human body, arms, hands or grippers, manipulated objects, and interaction regions remain visible.
  The record distinguishes partial cropping, complete departure from the frame, and occlusion by another entity, with entry or exit intervals where observable.
  We track body visibility separately from the visibility of the active manipulator and target object: a robot base outside the frame can coexist with a fully visible manipulation.
  These fields capture whether the evidence required for understanding an action or reconstructing its motion is present.
\end{itemize}

\noindent\textbf{Tactile Observations and Contact Evidence.}
Tactile observations provide local contact and force information for understanding physical
interactions.
The reliability of these observations, however, depends on signal quality and sensing coverage,
which vary across sensing technologies and sensor
configurations~\cite{higuera2024sparsh,yuan2026ftp}.
We therefore apply source-specific baseline correction and denoising to suppress spurious responses
caused by sensor noise and baseline drift.
For complex sensor layouts with incomplete coverage, we record coverage and measurement validity
separately from contact labels, so that missing or unreliable readings are not treated as evidence
of no contact.
The shared record retains these observations and annotations together with their source timestamps
and correspondence to visual and motion streams.

We populate fields according to the source modalities: images provide scene and spatial annotations,
videos additionally provide temporal and motion evidence, and robotics records can supply native
state and action channels, tactile observations, calibration, and outcome information.
Each preparation branch applies its own quality requirements to these attributes.

\noindent\textbf{From Shared Annotations to Focused Supervision.}
The resulting annotations support non-exclusive routing into four preparation branches.
The VLM branch constructs task-specific questions and targets from entity locations, spatial
relations, task hierarchies, execution stages, and outcomes.
The video-generation branch prepares captioned sequences using shared scene context, interval-level
action descriptions, and stage-dependent quality filtering.
The action branch combines the relevant intervals and instructions with native trajectories or
reconstructed motion, then performs alignment, retargeting, trajectory filtering, and multi-view
assembly as required.
The tactile branch combines tactile observations with contact annotations and validity information
to construct aligned tactile targets in a common representation.
A source may contribute to several branches, with each selecting the observations, annotations, and
supervision appropriate to its learning objective.

\subsection{Data for Embodied Vision--Language Understanding}
\label{sec:embodied_vlm_data}

\begin{figure*}[t]
  \centering
  \includegraphics[width=\textwidth]{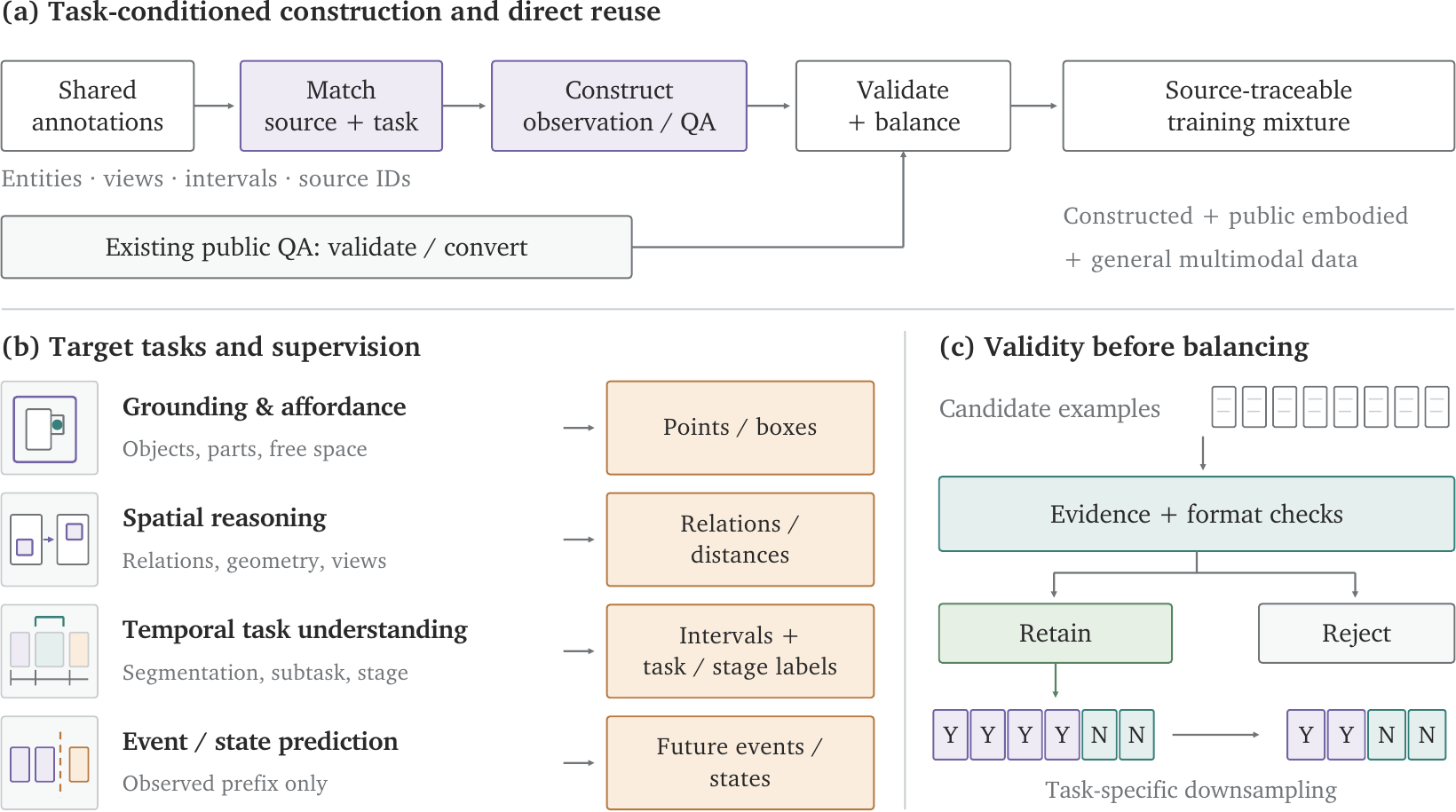}
  \caption{\textbf{The embodied understanding data pipeline.} Shared public sources and source-level annotations support task-conditioned screening, observation--target construction, and quality control. Existing public QA follows a separate validation and format-conversion path. The resulting corpus combines constructed task-focused supervision, public embodied data, and general multimodal data.}
  \label{fig:understanding_data_pipeline}
\end{figure*}

The VLM branch converts the shared source annotations into task-focused supervision for embodied
perception and reasoning. As illustrated in Figure~\ref{fig:understanding_data_pipeline}, it selects
task-compatible observations, constructs questions and targets grounded in the available evidence,
and combines the resulting examples with public multimodal training corpora.

\paragraph{Task Coverage and Specification}
We organize supervision around object grounding, spatial reasoning, and temporal task understanding,
including subtask identification and execution-stage recognition. Temporal segmentation and
future-event or state prediction are included when sufficient evidence is available. Each task
defines its observation requirements, relevant entities or events, and target format (e.g., text,
discrete choices, spatial coordinates, or temporal intervals), guiding source selection and
annotation.

\paragraph{Task-Conditioned Source Selection}
We draw from general videos, egocentric human videos, real-robot recordings, simulation data, and
public vision--language corpora. Shared spatial, semantic, temporal, and quality annotations support
task-conditioned screening for observable entities, interactions, and sufficient context. Selection
criteria vary by task: a clip unsuitable for precise localization may still support semantic
recognition, while one source may serve multiple tasks.

\paragraph{Observation and Supervision Construction}
For each selected source--task pair, we assemble an image, multi-view image set, or video segment
and construct an instruction--target pair. Supervision comes directly from reliable annotations or
through task-specific rewriting and model-assisted labeling. We preserve observation--annotation
correspondence and standardize target formats and coordinate conventions. For future prediction,
targets may derive from subsequent observations, but model inputs exclude the future information
being predicted.

\paragraph{Quality Assurance and Distribution Control}
We verify media availability, instruction--target consistency, annotation support, and task-specific
formatting, while retaining source provenance for deduplication and benchmark-overlap audits. To
reduce distributional shortcuts, we inspect task coverage and answer distributions, downsample
overrepresented binary or directional answers, and rebalance eligible multiple-choice options while
preserving their semantics and grounding. These measures complement annotation validity checks.

\paragraph{Integration with Public Multimodal Corpora}
Constructed examples are combined with public embodied datasets and general-purpose multimodal
corpora. Existing question--answer data enter through validation and format conversion without
regeneration. Public corpora provide broad visual and semantic coverage, while task-focused examples
strengthen embodied supervision.

\paragraph{Data Sources and Coverage}
We assemble a heterogeneous corpus covering general multimodal understanding, spatial perception,
multi-view and video reasoning, language-conditioned grounding, affordance understanding, and
robot-centric question answering. General-purpose data from Cambrian and LLaVA-OneVision provide
broad visual and instruction-following supervision. Embodied data supply complementary supervision
for spatial relations, metric estimation, viewpoint changes, object localization, and
interaction-relevant regions. Table~\ref{tab:understanding_data} in
Appendix~\ref{app:understanding_details} summarizes the public datasets and our task-focused
constructed collection represented in the understanding-tower training mixtures.

\subsection{Data for Video Generation}
\label{sec:video-data-pipeline}

The video-generation branch builds the visual dynamics foundation of the WAM from general videos,
egocentric human videos, and robot videos. It prepares temporally coherent training sequences,
generates captions matched to their content and temporal extent, and applies quality criteria
appropriate to each training stage. General videos provide broad visual and motion coverage, while
human and robot interaction videos focus learning on embodied events.

\noindent\textbf{Clip construction.}
We screen recordings before decoding, checking duration, frame dimensions, and frame rate against
criteria suited to each data source. We extract each selected interval once from its designated
camera view and store the clip in a common video encoding. Quality scoring, captioning, and latent
encoding then read the same clip file, reducing the need for format-specific preprocessing.

\phantomsection\label{sec:video-curation}%
\noindent\textbf{Frame- and clip-level scoring.}
On sampled frames, we measure brightness, saturation, and sharpness with classical image processing
methods on the CPU. Brightness and saturation help flag poor exposure or loss of color.
Gamma~\cite{zhou2025gamma} rates perceptual quality, the LAION aesthetic
predictor~\cite{schuhmann2022aestheticv2} scores visual appeal, and
PaddleOCR~\cite{du2020ppocr,paddleocr2023ppocrv4} locates text. We average LAION scores over the
sampled frames of each clip. For OCR, we compare text locations across frames to flag fixed corner
overlays. Persistence alone does not disqualify a clip: a product label may stay visible throughout
an action.

Models that score a clip as a whole include DOVER~\cite{wu2023dover} for technical and aesthetic
quality and VTSS~\cite{wang2025koala36m} for suitability for video training. We retain each score
with its clip and use stricter quality criteria to select the quality-tuning subset from the
continued-pretraining pool.

\Needspace{5\baselineskip}
\phantomsection\label{sec:video-captioning}%
\noindent\textbf{Prompt assembly.}
In each epoch of continued pretraining, we sample one prompt template compatible with each clip's
annotation fields. Descriptive prompts pair the scene with the observed action; instructions can
appear with or without the scene. The same clip can thus be paired with a description in one epoch
and an instruction in another, as illustrated in \figref{fig:caption-assembly}.

\subsection{Data for Action Generation}

The policy data pipeline prepares paired visual and motion supervision for Innata-0-WAM from public
real-robot and simulation trajectories, together with motion reconstructed and retargeted from
eligible human videos.
Shared task and temporal annotations associate each trajectory segment with its execution objective.
For caption variants that represent task progress, we pair the complete instruction with the active
subtask and previously completed tasks, matching the task context supplied by the agent under the
co-design protocol.

Preparation unifies action semantics, refines trajectory quality, and assembles synchronized camera
views.
Trajectories collected during system rollouts follow the same annotation and preparation procedures.
Successful execution segments, including corrective actions and post-recovery continuations, are
retained for subsequent policy training, allowing validated system experience to contribute
supervision through the shared pipeline.

\subsubsection{Gripper Manipulation Data}
\label{sec:gripper-action-data}

Gripper demonstrations span robot embodiments, tools, and control interfaces, so identical vector
shapes can encode different physical quantities. A pose may describe the wrist flange or grasp
center, and an action field may contain a controller command or a subsequent measured state. We
unify supervision around its physical meaning, enabling demonstrations from different embodiments to
provide compatible learning targets while retaining their kinematic differences. Preparation
therefore proceeds in two stages:
\textbf{semantic unification followed by trajectory quality filtering}. The first establishes what
each channel represents and when it applies; the second evaluates the reliability of the resulting
trajectories.

\paragraph{Stage 1: Unifying Action Semantics}
For each embodiment, we first determine what each field in the source data means and then match each
observation to its action target.
These pairs form the basis for two representations of robot motion: one in a predefined unified
end-effector (EEF) space and the other in joint space.

\noindent\textbf{Canonical end-effector geometry and motion.}
We define a fixed tool center point (TCP) at the center of the effective finger-pad grasp region in
the closed reference configuration. Its +Z axis points from the palm toward the fingertips, +X
follows the opening direction toward a specified finger, and +Y completes a right-handed frame.
Matched tool and mounting transforms map source poses to this TCP; forward kinematics provides poses
when measured joints are available but direct poses are unsuitable. Canonical geometry without
historical calibration defines an explicitly identified virtual TCP.

EEF State records the current TCP pose in a declared reference frame, such as a camera frame or
robot base/body frame~\cite{meeussen2010rep105}. Both arms of a bimanual embodiment share a
reference.
\textbf{We derive the pose component of EEF Action from consecutive measured states, expressing the current-to-next TCP transform in the current TCP frame.}
The current TCP defines both the origin and axes of this increment. The canonical pose and local
increment are
\begin{equation}
  \begin{aligned}
    T^{B}_{\mathrm{TCP},t}
    &= T^{B}_{F,t} H^{F}_{\mathrm{TCP}}, \\
    \Delta T_t
    &= \left(T^{B}_{\mathrm{TCP},t}\right)^{-1}
    T^{B}_{\mathrm{TCP},t+1}.
  \end{aligned}
  \label{eq:canonical-tcp-action}
\end{equation}
Here, $F$ denotes the recorded end-effector frame, $H^{F}_{\mathrm{TCP}}$ is the fixed pose of the
canonical TCP frame relative to $F$, including the tool mounting transform, and $B$ denotes a common
reference frame over the transition. We compute relative rotation before encoding it, rather than
subtracting rotation encodings~\cite{lynch2017modern}. Both poses must share a common reference over
the transition.
We compensate for camera or body motion when the necessary information is available and document any
stationary-reference assumption.

Each arm contributes ten channels. State contains three TCP position components, six
absolute-rotation components, and measured gripper opening. Action contains three local translation
components, six relative-rotation components, and an absolute gripper target. The 6D rotation
encoding concatenates the first two columns of the corresponding rotation
matrix~\cite{zhou2019continuity}.

\noindent\textbf{Joint states and targets.}
We order joints by name within each embodiment, using radians for revolute positions and meters for
prismatic positions. For ordinary joint-position channels, State contains current measured positions
and Action contains the next measured absolute positions, independently of the native controller
interface. Available head, torso, and other joints remain represented, while mobile-base channels
retain their declared pose or displacement semantics.

\noindent\textbf{Gripper opening and control signals.}
We normalize observed gripper opening to a scale from 0 (fully closed) to 1 (fully open), using the
limits specified in the URDF or manufacturer documentation.
Some sources provide separate observation and control signals, with commands sometimes encoded as
discrete values such as $\{-1,0,1\}$.
We prioritize observed gripper states during pretraining, then use control signals during supervised
fine-tuning to match the target robot's control interface.

\noindent\textbf{Layouts and storage.}
Joint uses a fixed 128-channel layout covering arms, grippers, hands, head, torso, mobile base,
legs, and extensions. EEF uses a separate 20-channel layout, with ten channels per arm. Single-arm
data only use the first ten. Unavailable channels are zero-filled and masked, with independent State
and Action validity masks. LeRobot v3 episodes~\cite{huggingface_lerobot_v3} retain source
timestamps, row mappings, per-view video offsets, and task references. Explicit timestamp mappings
establish observation--target alignment; normalization remains separate from semantic unification.
Original values, transformation metadata, proxies, and assumptions remain traceable.

\paragraph{Stage 2: Trajectory Quality Filtering}
We validate the standardized trajectories for non-finite active values, missing observations,
invalid masks or rotations, and inconsistent episode or media boundaries. Motion diagnostics examine
displacement, angular change, velocity, acceleration, jerk, local trend residuals, and oscillation
where the timebase and reference frames support them. Thresholds depend on embodiment,
representation, field, and sampling regime. Command--state checks use only independent, semantically
comparable commands; state-derived targets cannot independently validate their source states.

We combine these diagnostics with shared task, subtask, outcome, and visual-quality annotations to
reject defective trajectories or intervals.
Trajectory validity is assessed separately from task outcome: successful corrective segments remain
eligible even when earlier attempts failed, with their links to the failed attempt and subsequent
task continuation preserved.
When source outcome labels are reliable, they take precedence over model assessments.
Low motion or high jerk alone does not establish corruption: task context distinguishes redundant
idling from informative holding, contact, recovery, and waiting.

Removing an interval splits a trajectory into continuous episodes. We synchronize cuts across camera
views and paired Joint/EEF representations, retain complete subtask descriptions with adjusted
timestamps, and recompute or invalidate derived targets at boundaries. Targets never cross resets or
removed gaps; terminal observations can remain without valid next-step targets. Paired
representations count as one physical demonstration.

Selective smoothing applies only to validated noise candidates. It operates on continuous position
or joint sequences and treats rotations in rotation space, protecting contact events, gripper
transitions, and task boundaries. We recompute observation-derived targets, preserve native
commands, and retain a correction only if consistency and event-preservation checks pass.

\subsubsection{Dexterous Manipulation Data}
\label{sec:dexterous-action-data}

Dexterous demonstrations capture coordinated finger articulation for grasp adaptation and in-hand
manipulation, extending the opening-and-closing supervision provided by gripper data.
Differences in hand mechanisms, joint encodings, and EEF frames make this supervision difficult to
combine across embodiments.
We inherit the two-stage pipeline described for gripper data in
Section~\ref{sec:gripper-action-data}, reusing its visual processing, semantic annotations, and
general trajectory checks.
Here we detail the adaptations needed to align articulated hands and their motion.
Table~\ref{tab:egocentric_data} in Appendix~\ref{app:ego-action-data} lists the action-learning
sources, including complementary instrumented and robot demonstrations.

\newcommand{\dexdatasubsection}[1]{%
  \par\addpenalty{-250}\addvspace{8pt plus 2pt minus 1pt}%
  \noindent{\normalfont\bfseries #1}\par\nobreak}

\noindent\textbf{Unified hand representation.}
Robot hands differ in joint counts, axes, and link arrangements, so their native coordinates do not
provide a common articulation standard.
Despite these structural differences, hands exploit common object affordances to accomplish similar
goals, such as grasping a handle or turning a lid.
These shared interactions motivate aligning hand motion by its functional role, allowing
demonstrations from different embodiments to be related to common task semantics and contact
evidence.
We therefore design \texttt{Hand32} around human-hand anatomy, following the functional alignment
principle of FAAS~\cite{zhang2026unidex}.
Each coordinate identifies a finger, joint level, and motion function: flexion, abduction, or twist.
The thumb contributes seven angular coordinates and each other finger six, yielding 31 shared
coordinates and one reserved for a declared embodiment-specific extension.
Robot and human demonstrations thus share an articulation interface while retaining the motions
supported by each hand.

\noindent\textbf{Mapping diverse hand mechanisms.}
The human-hand reference does not imply a one-to-one joint correspondence: mechanical joints can
occupy different locations, several axes can realize one finger function, and a single motor can
drive several joints.
Mapping therefore accounts for both the intended motion and the mechanism that produces it.
We first identify corresponding finger functions from the source kinematic chain and reconcile joint
axes, neutral positions, and units.
Where direct correspondence is insufficient, inverse kinematics matches aligned finger endpoints to
support mappings between multiple source and reference degrees of freedom.
Three- or four-finger hands use explicit functional finger correspondences, with missing digits and
motions masked.
Known mimic relations remain coupling constraints, while whole-finger drive values use declared
equivalent-angle conversions.
Measured and derived articulation retain their distinction, and independent control availability is
recorded separately from observation and target validity.

\noindent\textbf{Joint and end-effector alignment.}
As with gripper data, we organize dexterous motion into Joint and EEF representations.
Joint data describe the arm configuration together with \texttt{Hand32} articulation, while EEF data
describe the hand's position and orientation, with finger articulation recorded separately when
available.
Table~\ref{tab:app-hand32-slots} lists the hand-local \texttt{Hand32}
indices, and Table~\ref{tab:app-device-hand32} summarizes the device mappings.
The recorded end-effector pose may refer to a flange, palm, or tracking device, each with a
different origin and orientation.
Using matched mounting transforms and kinematic models, we express these poses at a canonical wrist
whose origin is the wrist joint of the reference hand.
With palms down and fingers forward in the neutral configuration, its $+X$, $+Y$, and $+Z$ axes
point right, down, and forward, respectively, for both hands.
This frame follows rigid wrist motion throughout the trajectory.
Trajectory review also considers finger articulation and task context: a stationary wrist can
accompany grasp adjustment, while a stable hand can represent an informative hold.

\noindent\textbf{Instrumented glove data.}
Instrumented glove data, including PaXini's OmniSharing recordings~\cite{fan2026robopaint}, provide
articulation from joint sensors and known mechanical linkages, without requiring image-based
hand-pose reconstruction.
We process these recordings as a special dexterous embodiment, mapping glove kinematics into
\texttt{Hand32} with the same functional rules.
For PaXini's non-thumb fingers, the two distal glove hinges are summed into the corresponding
distal-flexion coordinate.
Where tracked hand-root poses are available, a source-to-wrist calibration also places them in the
canonical wrist frame for EEF supervision.
Finger articulation and hand placement therefore enter the shared representation through separate,
matched transformations, retaining their correspondence to visual and available tactile
observations.

\subsubsection{Egocentric Human Data}
\label{sec:dexterous-ego-data}
To learn from egocentric human videos, we adapt the two-stage pipeline in
Section~\ref{sec:gripper-action-data} to reconstructed hand motion.
We retain \texttt{Hand32} and the canonical wrist convention introduced for dexterous data.
In Stage~1, we recover and represent hand and camera motion separately to account for changes in
viewpoint.
In Stage~2, we filter trajectories based on reconstruction reliability and adjust their temporal
scale to match the pace of robot trajectories.

\paragraph{Hand Reconstruction and Functional Alignment}
We train a hand reconstruction model on 13 datasets with complementary joint and MANO supervision,
using features from \texttt{Wan2.2-Fun-5B-Control}~\cite{aigc_apps_VideoX_Fun_2026,wan2025wan} and
the decoder architecture of ACE-Ego-Hand~\cite{liu2026ace}.
We apply this model to egocentric videos to estimate hand skeletons, MANO
parameters~\cite{romero2017mano}, and hand existence and visibility scores.
For long recordings, we fuse overlapping predictions, estimate a shared shape for each hand within
each continuous sequence, and refine hand placement with the shape fixed.
We convert the reconstructed articulation to \texttt{Hand32} functional angles and apply a fixed
source-root-to-wrist transform to align the hand with the canonical wrist.
Forward kinematics under the shared shape then maintains consistency between the articulation and
reconstructed geometry.
The resulting canonical wrist poses and \texttt{Hand32} articulations serve as inputs to the
camera-aware motion representation in Stage~1. Appendix~\ref{app:hand-outputs} provides further
details.

\paragraph{Stage 1: Camera-Aware Motion Representation}
We standardize units, handedness, MANO conventions, and optical camera coordinates across sources.
When head-tracker poses are available, we apply a fixed tracker-to-camera transform to obtain the
corresponding camera poses.

Let $G_t={}^WT_{C_t}$ denote the main camera pose in a common metric world frame $W$, and
$P_t^h={}^WT_{E_t^h}$ the canonical wrist pose for hand $h\in\{L,R\}$.
A transform ${}^AT_B$ maps column-vector coordinates from $B$ to $A$, with optical camera axes
pointing right, down, and forward.
We anchor each continuous tracking segment at its first valid camera pose, $A={}^WT_{C_0}$, and
reuse this anchor across all episodes extracted from the segment.

We synchronize camera poses with their source frames and smooth wrist positions in world space using
confidence-weighted splines and wrist rotations using sign-continuous quaternion
splines~\cite{li2026scalable}, separately within each valid continuous interval.
Wrist Actions $D_t^h$ follow the local-increment definition in
Equation~\ref{eq:canonical-tcp-action}, computed from consecutive world-space canonical wrist poses.
We additionally define camera-relative wrist states, anchored camera states, and local camera-motion
actions as
\begin{equation}
  \begin{aligned}
    X_t^h &= G_t^{-1}P_t^h, \\
    S_t^C &= A^{-1}G_t,
    & D_t^C &= G_t^{-1}G_{t+1}.
  \end{aligned}
  \label{eq:ego-state-action}
\end{equation}
Here $X_t^h$ is the camera-frame wrist State, $S_t^C$ the anchored camera State, and $D_t^C$ the
local camera-motion Action.
Together with the world-frame wrist actions, these quantities distinguish hand motion from camera
motion while preserving camera-relative wrist states.

The resulting $128$-channel representation, \texttt{Ego128} (Table~\ref{tab:ego128-layout} in
Appendix~\ref{app:ego-action-data}), combines both wrists, both \texttt{Hand32} articulations, and
the main camera.
Translations are expressed in meters, and rotations use the 6D representation formed by the first
two columns of the rotation matrix, as in Section~\ref{sec:gripper-action-data}.
Hand articulation uses $q_t^h$ as its State and the absolute next-step articulation $q_{t+1}^h$ as
its Action.

\paragraph{Stage 2: Reconstruction Reliability and Temporal Alignment}
We assess reconstruction reliability using hand existence, visibility, source confidence, valid
duration, and temporal coverage, with source-specific thresholds and joint- or hand-level
reliability masks. We also apply image quality checks used for robot data.
We filter and segment the smoothed world-space wrist trajectories, removing abrupt motion
discontinuities and intervals where both hands are untracked to retain usable interaction
segments~\cite{bi2026motus2}.
Task and temporal annotations from Section~\ref{sec:video-task-semantics} further guide the
selection of complete manipulation subtasks.

Finally, we align human trajectories with the robot training data using a global slowdown factor of
$2$ and a 30\,Hz training timebase~\cite{yuan2026motiontrans,kareer2025egomimic,pmlr-v305-qiu25a}.
We jointly resample wrist and camera poses and hand articulation, using continuous interpolation for
translations and valid joint-angle trajectories, and spherical interpolation for rotations.

\FloatBarrier

\begin{figure}[t]
  \centering
  \includegraphics[width=\linewidth]{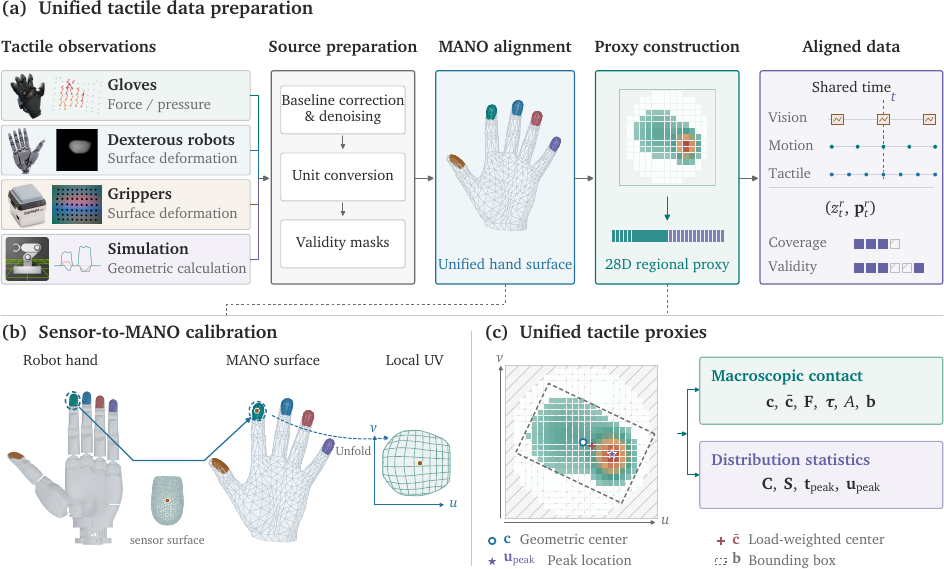}
  \caption{\textbf{Unified tactile data preparation.} Heterogeneous tactile observations undergo source preparation, alignment to a unified MANO hand surface, and regional proxy construction. The resulting presence--proxy pairs are temporally aligned with visual and motion observations, with sensor coverage and feature validity recorded separately.}
  \label{fig:unified-tactile-data}
\end{figure}

\subsection{Data for Tactile Generation}
\label{sec:tactile-data}

Tactile observations provide direct evidence of contact and loading that visual observations alone
may not resolve.
Their form, however, varies with the sensing principle: resistive, magnetic, and optical sensors can
produce pressure readings, directional force arrays, or images of surface deformation, with
different physical observability~\cite{yuan2026ftp}.
Sensor density, coverage, and placement also vary across embodiments, from isolated fingertip pads
to arrays over fingers and palms.
After source preparation, we align these heterogeneous observations through a common geometric
reference and a unified tactile proxy for contact geometry and loading.
Figure~\ref{fig:unified-tactile-data} outlines the preparation pipeline.

\noindent\textbf{Sensor-to-MANO alignment.}
We use MANO~\cite{romero2017mano} as a unified hand surface.
To accommodate differences in hand structure and sensor mounting, we develop a sensor-to-MANO
calibration tool.
Guided by the functional alignment principle of FAAS~\cite{zhang2026unidex}, it aligns each sensor's
footprint and local directions with a corresponding MANO region, such as the index fingertip for a
single gripper pad.
Calibration produces a fixed correspondence for each sensor layout, reused across recordings to
place its measurements consistently on the hand surface.
A fixed regional UV chart provides two-dimensional coordinates for describing contact location and
shape on the hand surface.
The mapped sensor footprint identifies which part of that surface is observable, so unsensed areas
remain masked.

\noindent\textbf{Unified tactile proxies.}
To accommodate different signal dimensions and spatial densities, we aggregate the aligned contact
field into regional tactile proxies describing macroscopic contact and distribution statistics.
Where sensor calibration supports absolute force, measurements are converted to newtons; pressure is
integrated over its physical support area, and directional forces are expressed in the common local
frame.
Surface integrals of traction (force per unit area) give the resultant force and its moment about
the load-weighted contact center, forming a six-dimensional wrench.
Uncalibrated sources retain their relative-response semantics and component-validity masks.

We compute contact area and moment arms from three-dimensional surface geometry, and describe
contact centers and footprint shape in the regional UV coordinates established above.
The stored surface correspondence anchors these UV features back to the MANO hand.
The regional proxy is

\begin{equation}
  \begin{aligned}
    \mathbf{p}_t^r = \big[&
    \mathbf{c}_t^r;\,\bar{\mathbf{c}}_t^r;\,
    \mathbf{F}_t^r;\,\boldsymbol{\tau}_t^r;\,A_t^r;\,\mathbf{b}_t^r;\\
    &\operatorname{symvec}(\mathbf{C}_t^r);\,
    \operatorname{symvec}(\mathbf{S}_t^r);\,
    \mathbf{t}_{\mathrm{peak},t}^r;\,
    \mathbf{u}_{\mathrm{peak},t}^r\big]\in\mathbb{R}^{28}.
  \end{aligned}
  \label{eq:tactile-proxy}
\end{equation}

The first group describes the macroscopic contact: geometric and load-weighted UV centers
$\mathbf c$ and $\bar{\mathbf c}$, resultant $\mathbf F$, moment $\boldsymbol\tau$, contact area
$A$, and bounding-box scale, anisotropy, and orientation $\mathbf b$.
The remaining fields retain finer distribution information: the six independent entries of traction
covariance $\mathbf C$, the three entries of load-weighted UV covariance $\mathbf S$, a
three-dimensional peak traction vector $\mathbf t_{\mathrm{peak}}$, and its two-dimensional location
$\mathbf u_{\mathrm{peak}}$.
Here $\operatorname{symvec}$ packs the independent entries of a symmetric matrix.
These statistics describe load variation, spatial concentration, and local peaks without tying the
representation to a fixed taxel count.

Each proxy is paired with a contact-presence label at the same MANO region and timestamp, forming
synchronized pairs $(z_t^r,\mathbf p_t^r)$.
Sensor coverage and feature validity remain separate from this label, distinguishing missing
observations from measured absence of contact.
Source-specific normalization preserves the shared scales of coupled directional components.
The pairs remain aligned with visual and motion observations on a shared time axis across sampling
rates, supplying tactile inputs and prediction targets.

\FloatBarrier

\section{World Action Model}
\label{sec:world-model-architecture}
We develop Innata-0-WAM as a unified World Action Model (WAM) that connects vision--language
understanding, prediction, and embodied control.
Under the co-design protocol, WAM generates action chunks conditioned on observations and the task
context maintained by the agent.
Its execution capabilities, in turn, guide how the agent organizes and delegates tasks.
Training on this shared task context provides the execution foundation for \emph{Plan}, while
corrective trajectories teach the model to address deviations and continue task execution for
\emph{Recover}.
Validated experience contributes to \emph{Evolve} through the agent's refinement of task context and
the reuse of curated rollouts in subsequent policy training.

Visual observations provide \emph{semantic evidence} for grounding language in scenes, objects, and
their relationships.
They also serve as \emph{predictive context} for learning temporal dynamics through video
pretraining, including observable action consequences when actions are supplied.
For policy learning, observation history provides \emph{control context} for interpreting the
current situation and generating executable action sequences.

We integrate these roles within a Mixture-of-Transformers (MoT) architecture that jointly models
actions, visual dynamics, and tactile feedback.
We support this architecture with a distributed training and inference framework for variable-length
multimodal sequences.
After adapting the understanding tower to embodied tasks, we train the world prediction tower
through video generation and action-conditioned prediction.
Policy pretraining then introduces action generation, followed by robot-state conditioning and joint
action--tactile learning.

\subsection{World Model Architecture}
\label{sec:world-model}

\subsubsection{World Understanding Tower}

\label{sec:understanding_tower}

The understanding tower adopts a decoder-only vision--language architecture initialized from
Qwen3-VL-8B~\citep{bai2025qwen30vl}. It consists of a vision encoder, a vision--language merger, and
an autoregressive language model. The vision encoder extracts visual representations from images and
video frames, which are projected into the language model's embedding space through the merger.
Given visual observations and language instructions, the model produces natural-language responses
and structured spatial predictions, such as points and bounding boxes.

\subsubsection{World Prediction Tower}

The world prediction tower uses the 8B backbone of Cosmos~3 Nano~\citep{nvidia2026cosmos3}, with 36
Transformer layers and a hidden width of 4096. A video variational autoencoder (VAE) encodes
observations into spatiotemporal latents, which are patchified and projected into video tokens.
Conditioned on language and available visual context, the tower predicts target video latents; when
a time-aligned action trajectory is also supplied, it predicts the corresponding visual changes.
Video generation serves as a pretraining objective for learning temporal dynamics, with an emphasis
on spatial displacement and object interactions relevant to manipulation.

We train the tower with conditional flow matching in video latent space. Gaussian noise is
interpolated with clean target latents, and the tower predicts the velocity along this path,
conditioned on the noise level. The loss applies to target video latents; observed conditioning
frames remain clean and are excluded from the loss. This formulation accommodates both prediction
from visual and language context and action-conditioned world modeling.

\subsubsection{Action and Tactile Generation Tower}
\label{sec:action-generation-tower}

The action and tactile generation tower is a separately parameterized Transformer with approximately
1B parameters. It retains the 36-layer depth of the two 8B towers while reducing the hidden width to
1280 and the feed-forward width to 3840. Each layer uses 32 query heads, 8 key/value heads, and a
head dimension of 128, preserving compatibility with the larger towers' attention interfaces. These
compatible attention dimensions allow the towers to exchange information despite their different
hidden widths.

\paragraph{Cross-tower interaction}
Video and action generation tokens, together with tactile tokens when present, interact through full
attention within each example, while the understanding tower supplies conditioning features to the
generation streams. Attention among generation tokens is bidirectional, enabling joint prediction
across modalities. Observed frames, together with available robot states and tactile observations,
remain clean conditioning inputs; future video, action, and tactile tokens are prediction targets.
Each generation stream is conditioned on its own noise level. This shared computation connects
action generation with predictions of the accompanying visual changes and contact feedback.

\paragraph{Domain-aware projections}
We use domain-aware affine maps at the input and output of the action tower to connect heterogeneous
control spaces to a shared Transformer. For projection domain $d$, each map takes the form
\begin{equation}
  f_d(x)=xW_d+b_d,
  \label{eq:domain-aware-action-projection}
\end{equation}
with a separate matrix and bias for each domain. The input map projects normalized, padded
128-dimensional state or action representations into the 1280-dimensional hidden space; the output
map returns 128-dimensional action-space flow velocities. EEF states and action targets use separate
projections because absolute observations and relative motion targets have different semantics. A
separate shared pair serves EEF-containing dexterous-hand and egocentric representations, including
wrist motion and hand articulation. Joint-space representations use embodiment-specific or shared
projection families according to their control semantics and joint layouts.
Table~\ref{tab:wam-domain-projections} summarizes these families and their implementation
identifiers.

\begin{table}[t]
  \centering
  \small
  \caption{\textbf{Domain-aware projection interfaces.} The implementation provides 64 projection slots. Compatible sources may share an interface. Projection IDs select input/output maps and are separate from the source identifiers used to select normalization statistics.}
  \label{tab:wam-domain-projections}
  \begin{tabularx}{\linewidth}{@{}Xcc@{}}
    \toprule
    Projection family & State ID & Action ID \\
    \midrule
    Standard EEF representations & 32 & 0 \\
    Shared EEF-containing dexterous-hand and egocentric representations & 62 & 63 \\
    Joint-space representations & \multicolumn{2}{c}{By embodiment or registered family} \\
    \bottomrule
  \end{tabularx}

\end{table}

\paragraph{Action supervision}
We apply conditional flow matching directly to continuous action chunks. Let $a$ denote a clean
normalized action chunk, $\epsilon_a\sim\mathcal{N}(0,I)$ independent Gaussian noise, and
$\sigma_a\in[0,1]$ its noise level, with larger values denoting noisier inputs. We define
\begin{equation}
  a_{\sigma_a}=(1-\sigma_a)a+\sigma_a\epsilon_a,
  \qquad v_a^\star=\epsilon_a-a.
  \label{eq:action-flow-path}
\end{equation}
Let $c$ collect clean conditioning inputs, including visual history, language conditioning, and
available robot states and tactile observations.
The language input contains the task instruction and, for the corresponding caption variants, the
active subtask and completed-subtask history (Section~\ref{sec:wam-multimodal-alignment}).
Let $\widetilde z_v$ and $\widetilde z_\tau$ denote the video and tactile streams at their
respective noise levels, when present.
The action objective is
\begin{equation}
  \mathcal{L}_{\mathrm{action}}=
  \mathbb{E}\!\left[
  \frac{\left\|M_a\odot\left(\hat v^a_\theta(a_{\sigma_a},\sigma_a\mid c,\widetilde z_v,\widetilde z_\tau)-v_a^\star\right)\right\|_2^2}
  {\sum M_a}
  \right],
  \label{eq:action-flow-matching}
\end{equation}
where $M_a$ selects valid target action entries, excluding conditioning states and padded
dimensions. At inference, integrating the predicted velocity field from the noisy endpoint toward
the clean endpoint produces an action chunk.

Within the agent loop, successive action chunks pursue the active subtask using updated
observations.
A subtask may span multiple chunks; the agent verifies its physical outcome before activating the
next subtask.
Once recovery is verified as successful, WAM resumes execution from the repaired state under the
updated task context.

\paragraph{Tactile generation and supervision}
Manipulation actions affect objects through contact, whose occurrence and loading determine how
robot motion is transmitted to the environment.
Tactile observations provide direct feedback on these interactions.
Predicting them alongside actions therefore gives the action tower supervision of both the motion
and its contact consequences.
Contacts are sparse and can appear or disappear abruptly; we therefore pair each regional proxy
$\mathbf p$ from \secref{sec:tactile-data} with a presence state $z$, modeling both contact
occurrence and the attributes of an active contact.
For each future pair, both components follow the action flow path,
$x_\sigma=(1-\sigma)x+\sigma\epsilon_x$, with independent Gaussian noise $\epsilon_x$.
Separate input projections combine $\mathbf p_\sigma$ and $z_\sigma$ into one token per region and
timestamp, which the shared action tower processes jointly with action tokens.
Two output projections predict their flow velocities.
As these velocities update the pair, the evolving presence state is encoded again at the next
denoising step, allowing proxy prediction to use the model's current estimate of contact occurrence.
To train the pair within the same continuous generation process, we apply a velocity MSE loss to
each branch.
Presence also defines a discrete contact decision at abrupt transitions, so we add a small BCE term
on its reconstructed clean estimate.
The prediction objective is
\begin{equation}
  \begin{aligned}
    \mathcal L_{\mathrm{tactile}}^{\mathrm{pred}}
    &=\frac{
      \operatorname{MSE}_{M_p}(\hat v_p,v_p^\star)
      +\lambda_z\operatorname{MSE}_{M_z}(\hat v_z,v_z^\star)
    }{1+\lambda_z}\\
    &\quad+\lambda_{\mathrm{BCE}}\operatorname{BCE}_{M_z}(\hat q_\sigma,y),
  \end{aligned}
  \label{eq:tactile-prediction-loss}
\end{equation}
where $v_x^\star=\epsilon_x-x$, $y$ is the binary contact label, and
$\hat q_\sigma=\operatorname{sigmoid}\!\left(2[z_\sigma-\sigma\hat v_z]\right)$ reconstructs the
clean contact probability at the sampled noise level.
Each term is averaged over its valid target entries.
Since contact location and shape are undefined without contact, $M_p$ selects valid proxy features
at contact targets; $M_z$ also includes valid no-contact labels, preserving supervision of contact
absence while excluding missing measurements.
At inference, integrating the predicted velocities from Gaussian noise yields the future pair
$(\hat{\mathbf p},\hat z)$.
For tactile readout, we compute $\hat q_{\mathrm{out}}=\operatorname{sigmoid}(2\hat z)$ and return
$\widetilde{\mathbf p}=\mathbb{1}[\hat q_{\mathrm{out}}\geq 0.5]\hat{\mathbf p}$, retaining contact
attributes only where the model predicts contact.

\subsubsection{Multimodal Conditioning and Alignment}
\label{sec:wam-multimodal-alignment}

We condition the model on camera observations and structured task instructions, with recent state
and tactile histories when available.
Video, action, and tactile tokens share a physical time axis, while spatial coordinates distinguish
camera views and tactile regions.

\paragraph{Single- and multi-view observations}
The policy accepts single-camera or synchronized multi-camera observations. For multiple cameras, we
decode frames on a common temporal grid and resize them into one composite canvas before VAE
encoding. Two views are stacked vertically. With three views, the primary view occupies the upper
region, and the remaining views occupy the lower-left and lower-right regions. Thin black separators
distinguish neighboring camera regions. Camera assignments remain consistent throughout each sampled
clip. Views captured at the same timestamp share a temporal position and occupy distinct spatial
regions of the video token grid.

\paragraph{Structured task conditioning}
Language conditioning provides the shared task interface through which the co-design protocol
connects agent decisions to model execution.
Caption variants that represent task progress pair the complete instruction with the active subtask
and a history of completed subtasks.
The complete instruction preserves the overall goal, the active subtask specifies the immediate
execution objective, and the history records completed stages.
The history records completed stages.
Training annotations populate these fields; during execution, the agent maintains the same task
representation and updates subtask progress according to verified physical outcomes.

This interface lets the agent organize long-horizon tasks into objectives suited to the model's
execution capabilities, while the corresponding training supervision teaches WAM to act on those
objectives in context.
Validated experience can further refine the task commands supplied in subsequent executions through
the same interface, without an immediate change to model parameters.

Structured captions also retain the task title, embodiment identity, clip duration, video FPS,
action frequency, canvas resolution, and aspect ratio.
For multi-view observations, they specify each camera's semantic role, canvas location, and region
dimensions.
These fields make the sampling rates and camera layout explicit in the conditioning text; the
positional encoding below represents their temporal and spatial organization numerically.

\paragraph{FPS-aligned rotary positions}
We use three-axis multimodal RoPE with temporal, height, and width coordinates. Text tokens advance
along all three axes, while video tokens follow a spatial patch grid and a temporal coordinate
determined by physical time. With reference FPS $f_{\mathrm{ref}}=24$ and reference VAE temporal
compression $c_{\mathrm{ref}}=4$, a token at time $\tau$ seconds receives the temporal coordinate
\begin{equation}
  p_t(\tau)=b+\frac{f_{\mathrm{ref}}}{c_{\mathrm{ref}}}\tau
  =b+6\tau,
  \label{eq:wam-fps-rope}
\end{equation}
where $b$ is the shared temporal origin of the associated video, action, and tactile streams. On a
regular video latent grid, index $k$ corresponds to $\tau=k c_v/f_v$, with video FPS $f_v$ and VAE
temporal compression $c_v$. Action tokens are temporally uncompressed: their coordinates follow the
action sampling rate and sequence alignment offset, or explicit timestamps when supplied. The
128-dimensional state/action pathway uses explicit timestamps, allowing the current state and first
action to share a temporal coordinate when they correspond to the same time. Floating-point position
IDs preserve fractional offsets. Equal elapsed times therefore receive equal temporal displacements
across streams with different sampling rates. These physical-time coordinates are separate from the
noise levels used for flow matching.

\paragraph{Spatial positions for stitched views}
Spatial coordinates reset for each video segment. For multi-view canvases, we express vertical
position relative to the seam between the upper and lower camera regions:
\begin{equation}
  p_h=s-h,
\end{equation}
where $h$ is the patch-row coordinate and $s$ is the seam location in patch units, computed from the
actual camera-layout geometry. The seam location may be fractional. Width coordinates retain their
canvas positions, distinguishing the two lower views in the three-camera layout. Single-view
observations retain the ordinary spatial grid. This transformation leaves temporal coordinates
unchanged and preserves physical-time alignment across modalities.

\paragraph{Tactile alignment}
A no-contact reading can occur both before first touch and after release, leaving the stage of an
interaction ambiguous from a single observation.
Recent tactile and state histories help resolve this ambiguity by recording preceding contacts and
the motion that led to the current state.
We retain these histories as clean conditioning for joint generation of future tactile, action, and
video tokens, with losses applied only to future targets.
To preserve when contact changes occur relative to motion, tactile mRoPE uses actual timestamps on
the shared temporal axis of \eqref{eq:wam-fps-rope}.
Tactile samples remain temporally uncompressed, and all regions at the same instant share a temporal
coordinate, preserving their timing relative to action and video across sampling rates.
The two spatial axes encode the canonical functional region and hand side, so each region keeps the
same anatomical coordinates across time and embodiments; contact position within the region remains
part of the predicted proxy.
Because these coordinates identify fixed anatomical slots, observed state embeddings additionally
provide their pose context: historical tokens use the corresponding observed states, while future
queries use the current observed state.

\FloatBarrier
\subsection{Infrastructure}

\phantomsection\label{sec:infra}

We extend NVIDIA's \texttt{cosmos-framework} and Cosmos~3 execution
stack~\citep{infra_framework,nvidia2026cosmos3} to train the world model described in
\secref{sec:world-model}. Variable-length video, action, and tactile sequences constrain device
memory, graph reuse, and input supply. Our training framework addresses these constraints through
hybrid sharding, fixed-capacity execution, and shared data preparation, then exports weights and
preprocessing state for inference. The measurements below use historical workloads without tactile
($T=0$); \appref{app:infra:experiments} identifies their configurations and scope.

\begin{figure}[htbp]
  \centering
  \includegraphics[width=\linewidth]{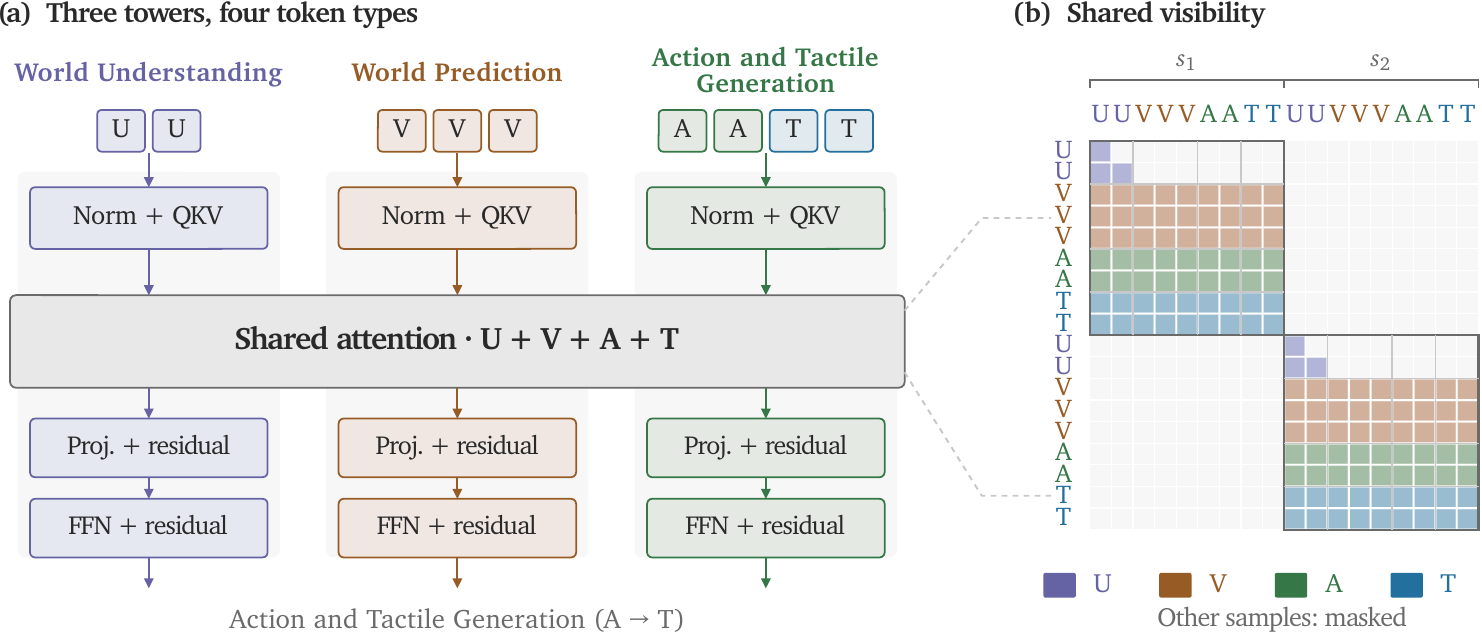}
  \caption{\textbf{Four token types, three towers.} Violet U, ochre V, green A, and
    blue T denote understanding, video, action, and tactile tokens,
    respectively. The World Prediction Tower processes video tokens.
    Each sample illustrates two U, three V, two A, and two T tokens;
    A precedes T within the shared Action and Tactile Generation Tower. U attention is causal;
    V/A/T queries attend to all valid same-sample tokens. Packed samples
    remain isolated; token counts are schematic.}\label{fig:infra:mot}
\end{figure}

\figref{fig:infra:mot} shows the execution contract: understanding
tokens (U) use the World Understanding Tower, video tokens (V)
use the World Prediction Tower, and action tokens (A)
followed by tactile tokens~(T) use the shared Action and Tactile Generation Tower.
Tower-specific projections and FFNs surround joint attention;
masks isolate packed samples and exclude padding from supervision.
Clean conditioning and prediction targets use separate validity
and loss masks.

\subsubsection{Training Framework}\phantomsection\label{sec:infra:training}

Training on variable-length multimodal examples requires fitting model state and token batches in
GPU memory while keeping the devices supplied with data.
We address these needs through distributed sharding, fixed-capacity token buffers, and a data loader
that overlaps batch delivery with computation.

\paragraph{Distributed execution}\phantomsection\label{sec:infra:distributed}
Fully sharded data parallelism (FSDP2) partitions parameters, gradients, and optimizer state; hybrid
sharding (HSDP) synchronizes corresponding shards across replicas~\citep{infra_zero,infra_fsdp}. On
128 GPUs, shard8 $\times$ replica16 confines parameter collectives to eight-GPU nodes. In our target
production recipe, shard8 could not fit the 96K video-token budget. Shard16 $\times$ replica8 halves
persistent state per device from $M/8$ to $M/16$ for total unsharded state $M$, at the cost of
cross-node parameter communication. With two gradient-accumulation steps (GA2), BF16, full
recomputation, and static compilation, the resulting text-to-video (T2V) recipe supports
U3,072/V98,304 tokens per microbatch. The historical 64K-to-96K offline transition reports
\textbf{8.9\%} higher sample throughput; this is a configuration-level comparison.
\appref{app:infra:experiments} separates that recipe from the later shard8/96K kernel benchmark.

\begin{figure}[htbp]
  \centering
  \includegraphics[width=\linewidth]{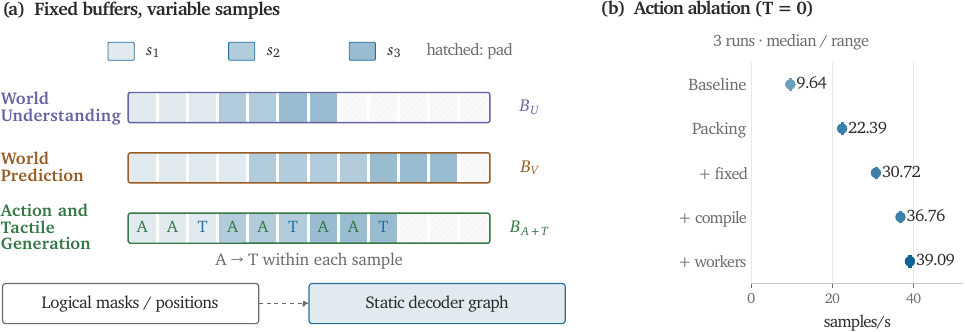}
  \caption{\textbf{Static token buffers and training throughput.} (a) U and V fill the World Understanding and World Prediction buffers. A and T share a buffer of capacity $B_{A+T}$ in the Action and Tactile Generation Tower. Padding is excluded from attention and supervision. (b) Historical DROID Action ablation without tactile ($T=0$): \texttt{width\_only\_1b}, without state conditioning, frozen World Understanding Tower, eight GPUs, BF16, online VAE, full recomputation, and a global batch size of 32. Three runs/configuration, updates 11–40; points invert median run-mean time and ranges invert observed extrema. \tabref{tab:infra:static-config} specifies each configuration.}\label{fig:infra:static}
\end{figure}

\paragraph{Compute optimization}\phantomsection\label{sec:infra:compute}
Fixed U, V, and shared A+T token buffers separate physical capacity from logical sample length.
Positions and masks retain variable-length semantics; dynamic preparation stays outside compiled
decoder blocks, allowing graph reuse~\citep{infra_torch2}. In the DROID Action ablation in
\figref{fig:infra:static}, packing raises throughput from 9.64 to 22.39 samples/s. Fixed shapes and
compilation reach 36.76 samples/s, \textbf{64.2\% above packing alone}; additional workers reach
39.09 samples/s, or \textbf{4.06$\times$ the initial short-window baseline}. The global batch size
is fixed at 32. Startup and compilation are excluded.

For T2V, offline latents remove repeated decoding, preprocessing, and frozen VAE encoding while
preserving frame selection and temporal context. Noise and targets remain constructed online. A
four-GPU profile records 1.069 s of online VAE work per captured rank-0 update and no VAE work in
the offline trace; different catalogs prevent a matched end-to-end speedup estimate. Preprocessing
and latent storage are separate costs.

We extend FlashAttention-3~\citep{infra_fa3} to select the attention mask per segment. A single
dispatch then applies causal attention to understanding queries and full attention to video, action,
and tactile queries within each sample. In the T2V trace, grouping halves main-kernel launches with
little change in update time. Operator checks pass; full-update numerical acceptance remains
unresolved. \appref{app:infra:execution} gives the kernel design, communication and scheduling
experiments, and compute/memory tradeoffs.

\begin{figure}[htbp]
  \centering
  \includegraphics[width=\linewidth]{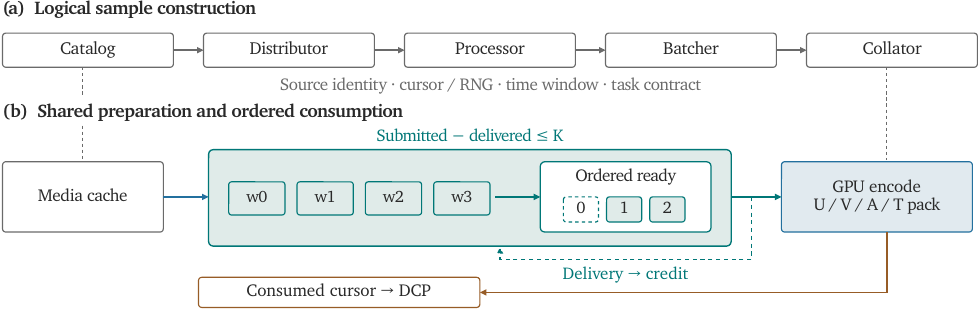}
  \caption{\textbf{Logical construction and physical preparation.} Packing preserves U/V/A/T order, with A/T routed to the Action and Tactile Generation Tower. Per-rank budget $K$ bounds submitted-but-undelivered tasks across workers, transport, optional pinning, and ordered readiness. Delivery returns a credit; consumption commits recoverable progress. Active batches and media caches have separate budgets.}\label{fig:infra:data}
\end{figure}

\paragraph{Data loading}\phantomsection\label{sec:infra:data}
We separate source identity, sampling order, and recoverable cursors from physical workers
(\figref{fig:infra:data}). A shared pool prepares batches under a rank-wide credit budget $K$:
delivery returns a credit, while consumption commits recoverable progress. $K$ bounds
submitted-but-undelivered tasks, not bytes; active batches and media caches have separate budgets,
and ordered delivery can still stall.

\begin{figure}[htbp]
  \centering
  \includegraphics[width=\linewidth]{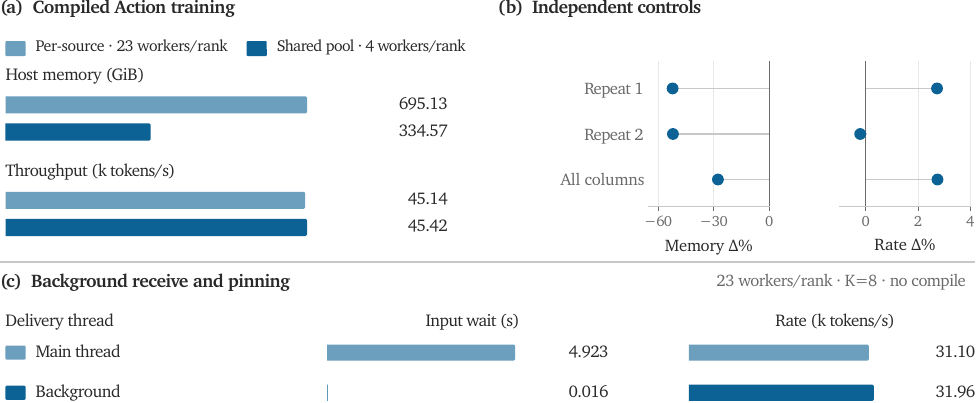}
  \caption{\textbf{Eight GPUs, 23 sources.} (a) Compiled: four warmup/eight timed updates, one run/arm. (b) Two independent repeats: 4/12 updates each; all-column control: 2/6, reading complete episode tables. Each uses its own per-source baseline. (c) Fixed 23 workers per rank and $K=8$, 2/6 updates, one run/arm. Input wait is the six-update rank-0 sum. In (b,c), decoder and ahead-of-time VAE compilation are disabled. Memory is whole-run node anon+shmem maximum, sampled every two seconds, excluding file cache.}\label{fig:infra:loader}
\end{figure}

The per-source loader baseline uses 23 workers per rank with a prefetch factor of 2. The shared pool
uses four workers per rank with a rank-wide credit budget of $K=16$. In the compiled comparison,
node host memory falls from \textbf{695.13 to 334.57~GiB ($-51.9\%$)} at similar measured throughput
(\figref{fig:infra:loader}(a)). Independent repeats save approximately 52\% memory, with throughput
changes of $+2.73\%$ and $-0.21\%$. These compare complete loader/resource configurations; the short
compiled replay covers nine of 23 sources.

\FloatBarrier

Background IPC reception, deserialization, and optional pinning overlap delivery with compute. With
23 workers per rank, $K=8$, and memory pinned in both arms, rank-0 input wait over six updates falls
from \textbf{4.923 to 0.016~s}. Throughput rises from 31.10 to 31.96~k tokens/s, and host memory
rises from 244.92 to 265.91~GiB.

Fingerprint-indexed metadata also reduces 58-source data initialization from 643.39 to 14.59 s with
cache hits. The comparison leaves the OS file cache in place and excludes model loading and
compilation. \appref{app:infra:measurements} reports controls and longer input-wait measurements.

\paragraph{Other optimizations}\phantomsection\label{sec:infra:other}
Selective recomputation trades activation memory for less repeated
computation~\citep{infra_recompute}; precision and VAE compilation studies appear in
\tabref{tab:infra:compute}. On-Demand Communication~\citep{infra_odc} removes layerwise rendezvous,
and cost-aware reassignment reduces predicted attention
imbalance~\citep{infra_hydraulis,infra_bytescale,infra_disttrain}, but neither establishes an
end-to-end gain in the tested workloads (\appref{app:infra:execution}).

For recovery, Distributed Checkpoint~\citep{infra_dcp} saves model/optimizer state together with
logical data-shard identities, consumed cursors, and RNG state. Offline ownership remapping supports
\textbf{$160 \rightarrow 144$-GPU continuation}: all 11,136 logical states are retained and each
rank passes its first ten microbatch sequence checks (\tabref{tab:infra:recovery}). Global batch
composition may change; saved-state restoration does not establish numerical continuation
equivalence (\appref{app:infra:correctness}).

\paragraph{Training efficiency}\phantomsection\label{sec:infra:mfu}
\tabref{tab:infra:configuration} lists historical configurations on 80~GB GPUs. Model FLOPs utilization (MFU) \textbf{excludes activation recomputation and padding}~\citep{infra_palm,infra_recompute}. The 128-GPU online T2V audit yields \textbf{31.04\%} including the required forward pass through the frozen VAE, or \textbf{25.28\%} for the Transformer alone. These are rank-0 analytical estimates over 677 complete updates, not an all-rank average. Full inter-update time retains input and synchronization stalls; \appref{app:infra:accounting} provides the FLOP numerator and dense BF16 peak convention.

\begin{table}[htbp]
  \centering\small
  \setlength{\tabcolsep}{4pt}
  \renewcommand{\arraystretch}{1.10}
  \caption{\textbf{Historical diagnostic configurations.}
    These exclude tactile tokens ($T=0$) and differ from the curricula
    in \secref{sec:wam-training}. Token capacities are per GPU microbatch. Only the first row has a reconstructed useful-work MFU.}\label{tab:infra:configuration}\label{tab:infra:mfu}
  \begin{tabular}{@{}>{\raggedright\arraybackslash}p{\dimexpr0.32\linewidth-1.28\tabcolsep\relax} >{\raggedright\arraybackslash}p{\dimexpr0.20\linewidth-0.80\tabcolsep\relax} >{\raggedright\arraybackslash}p{\dimexpr0.48\linewidth-1.92\tabcolsep\relax}@{}}
    \toprule
    \textbf{Workload} & \textbf{Parallelism} & \textbf{Physical tokens; useful-work MFU} \\
    \midrule
    T2V, online encoding (audited) & 128 GPUs, GA4\newline shard8 \ensuremath{\times} replica16 & U 5,958; V 59,578; \textbf{31.04\%}, including VAE forward \\
    Three-tower Action, online encoding & 128 GPUs, GA2\newline shard16 \ensuremath{\times} replica8 & U 11,776; V 62,592; A 2,944; MFU not audited \\
    T2V, offline encoding & 128 GPUs, GA2\newline shard16 \ensuremath{\times} replica8 & U 3,072; V 98,304; MFU not audited \\
    \bottomrule
  \end{tabular}
\end{table}

\FloatBarrier
\subsubsection{Inference}\phantomsection\label{sec:infra:inference}
Exported weights, preprocessing assets, and normalization state initialize resident inference. The
policy and serving results come from separate tests of historical three-tower models without state
or tactile inputs (\appref{app:infra:experiments}).

\begin{itemize}
  \setlength{\itemsep}{2pt}
  \setlength{\parsep}{0pt}
  \setlength{\parskip}{0pt}
  \item \textbf{Video generation kernels.} Forward attention accounted for 34.0\% of cumulative compute-kernel time on rank~0 in 50-step text-to-video inference. To address this bottleneck, we optimized the attention computation using the FlagGems attention operator~\citep{flaggems2024} from FlagOS, achieving a 1.39–1.41× kernel-level speedup on BF16 tensor shapes extracted from the actual inference trace (\tabref{tab:infra:flaggems}).
  \item \textbf{Condition reuse.} Caching timestep-invariant states and key/value tensors from the World Understanding Tower for each request and classifier-free guidance (CFG) branch saves 72--73 ms on one GPU and 45--47 ms with two-GPU CFG parallelism. Outputs match in the tested cases.
  This cache is already enabled in the CFG parallelism baseline.
  \item \textbf{Two-GPU CFG parallelism.} Concurrent conditional/unconditional branches~\citep{infra_cfg-method} on two GPUs reduce action-only latency \textbf{$2.004 \rightarrow 1.054$~s ($1.90\times$)} at $960 \times 640$, 41 frames/40 actions, and five sampling steps. Allocated GPU-s/request rises from 2.004 to 2.108.
  \item \textbf{World Prediction Tower FFN reuse.} Similar outputs at adjacent steps motivate caching selected FFN outputs~\citep{infra_pab,infra_teacache}. Skipping 20\% of calls reduces prepared-execution latency by approximately \textbf{10\%}, with a 2.42 GiB cache. Mean rotation-6D RMSE rises 3.42\%; worst individual gripper-opening MAE rises 21.03\%. This approximation is disabled by default.
  \item \textbf{Bounded queues.} Each replica serves one request, queues two in FIFO order, rejects overflow, and skips expired requests. With an eight-second deadline, timely goodput rises \textbf{22.20\%} at 0.5 requests/s/GPU and falls \textbf{57.13\%} at 0.75.
\end{itemize}

Offline hand reconstruction combines mixed precision, compilation by input shape, and overlapped
window processing (\appref{app:hand-outputs}).
\FloatBarrier

\subsubsection{Evaluation Infrastructure}

\begin{figure}[H]
  \centering
  \includegraphics[width=\linewidth]{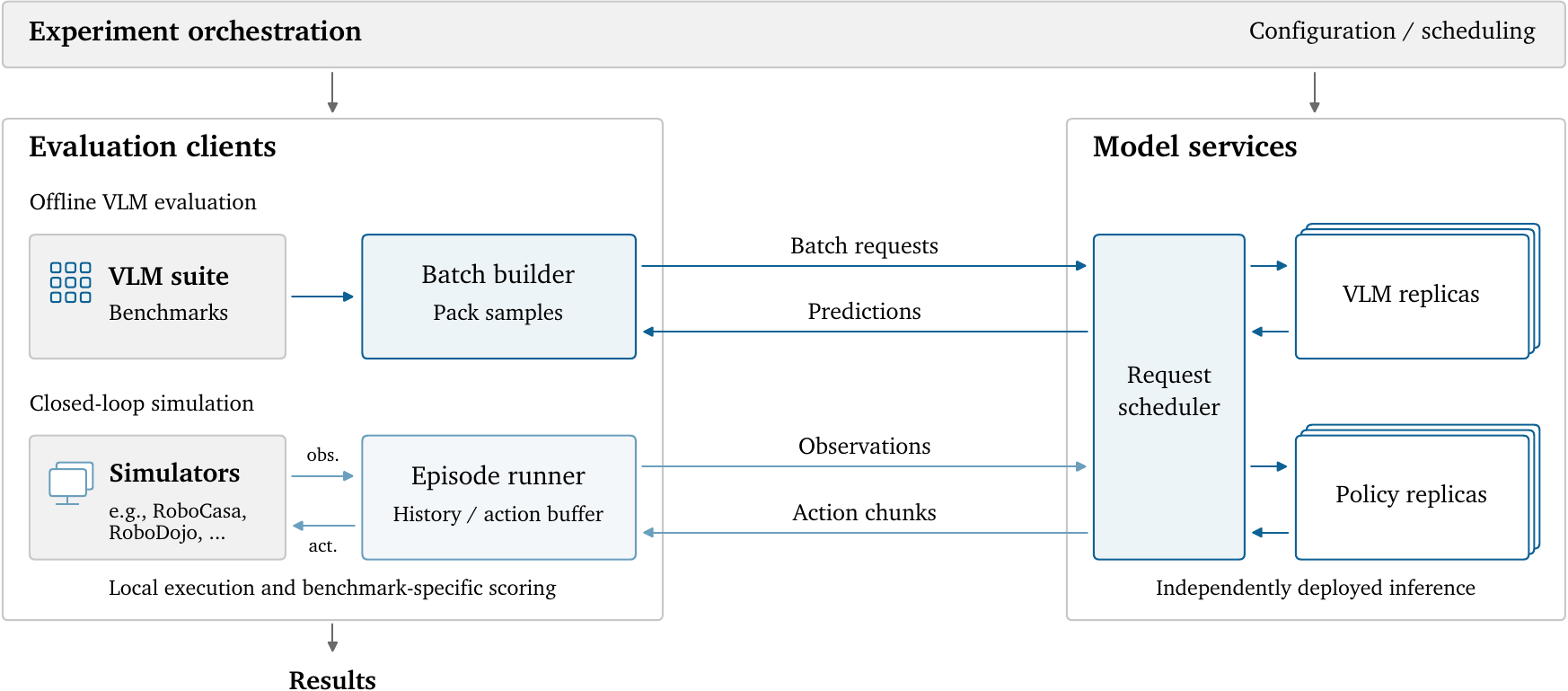}
  \caption{\textbf{Service-based evaluation infrastructure.} Vision--language benchmarks submit batched requests, while simulation clients query policy services as they advance each episode. Each benchmark retains its own scoring procedure.}
  \label{fig:evaluation-infrastructure}
\end{figure}

Our evaluation system supports vision--language benchmarks and closed-loop simulation across
multiple environments, including several simulation benchmarks
(Fig.~\ref{fig:evaluation-infrastructure}).
Evaluation clients execute benchmarks and communicate with independently deployed model services
through request--response interfaces, allowing data preparation, simulation, and inference to run on
separate resources.

For VLM evaluation, benchmark adapters prepare task-specific prompts, media inputs, and generation
settings.
A batch execution pipeline groups requests and distributes them to model services, then associates
returned predictions with the corresponding examples.
The framework supports native benchmarks and integrations with VLMEvalKit and lmms-eval, preserving
task-specific prompt construction and scoring procedures.

Simulation clients maintain environment state, observation history, and action buffers, requesting
policy inference when a new action chunk is needed.
Returned actions are executed locally, and subsequent requests incorporate updated observations.
A request scheduler dispatches inference requests from concurrent environments to available model
replicas.
Policy state is maintained separately for each environment, and responses are routed to the
corresponding episode.
This separation supports parallel simulation without embedding model inference in each environment
process.

Scoring follows each benchmark's protocol, either during execution or in a separate postprocessing
stage.
VLM benchmarks requiring dataset-level evaluation receive the collected predictions for scoring.
Simulation evaluation records task completion and, where defined, partial task progress.
Reports retain benchmark-specific summary metrics and individual evaluation cases.

The VLM backend saves questions, predictions, scores, and run manifests with model and benchmark
configurations, execution settings, and source-version information.
These records support inspection, resuming interrupted runs, and deferred scoring.
The simulation framework retains inference requests and responses, episode traces, and rollout
videos for analyzing policy behavior in relation to the corresponding observations and actions.
For system rollout collection, the same execution interfaces retain the observations, commands,
actions, and outcomes needed to prepare trajectories for policy training.

\subsection{Training}
\phantomsection\label{sec:wam-training}
Training proceeds in two major stages: \emph{video generation pretraining}, which develops visual
prediction through video generation and action-conditioned world modeling (ACWM), followed by
\emph{policy model pretraining}, which introduces action generation and subsequently joint
action--tactile generation. The understanding tower is adapted for embodied understanding before
these stages.

The raw source pool contains approximately 100,000 hours of data before filtering; stage-specific
training mixtures are constructed from this pool. When both end-effector (EEF) and joint-space
representations are available, we use both representations of the same experience: EEF targets
describe end-effector motion, while joint-space targets retain embodiment-specific articulation.
This expands the available supervision without increasing the underlying recorded duration.

\subsubsection{World Understanding Training}

Within the MoT architecture, the understanding tower is designed to develop spatial representations
relevant to robotic manipulation, with an emphasis on 3D spatial grounding, object localization, and
spatial relationships. Rather than pursuing general-purpose visual understanding alone, we
prioritize capabilities that are potentially useful for action generation, providing a perceptual
and spatial foundation for downstream robotic
decision-making~\citep{yuan2026democratizing, geminirobotics15}. We fine-tune the understanding
tower to strengthen these capabilities, with details of the training data and fine-tuning procedure
provided in Appendix~\ref{app:understanding_details}.

\subsubsection{Video Generation Pretraining}
\label{sec:wam-video-pretraining}

Video pretraining teaches the World Prediction Tower to predict how a manipulation scene evolves
from an observed state.
We adapt a general-purpose video generator to embodied interactions, then add examples that link a
supplied action trajectory to the resulting video.

\paragraph{From general video priors to embodied dynamics}
Adapting a pretrained video generator to robotic manipulation  scenarios
requires alignment in both domain and conditioning: the
T2V-centered priors inherited from Cosmos~3 Nano cover broad
visual content, whereas downstream manipulation tasks require modeling embodied interactions from a
specific observed state.

We therefore continue pretraining the model on a 10M-clip mixture of general videos, real-robot
demonstrations, and egocentric human interactions, shifting the emphasis toward embodied dynamics
and
text-and-image-to-video (TI2V) prediction. Each clip is paired with an action description or
instruction, and training is predominantly single-view,
modeling how the scene evolves from one viewpoint.
We retain the original text-only T2V supervision alongside TI2V tasks, preserving text-driven
general generation ability while developing prediction
grounded in observed scenes. All settings use the conditional
flow-matching objective. The understanding
tower remains frozen as the conditioning backbone throughout
this stage.

Later in this stage, we increase the
sampling emphasis on visually reliable examples according to
the quality scores in Section~\ref{sec:video-curation}, while
maintaining coverage across sources.

\paragraph{Combining video generation with action-conditioned prediction}
The final phase of video pretraining combines video generation with ACWM. An ACWM example provides
an initial observation, a language instruction, and a time-aligned action trajectory; the prediction
target is the resulting future video. The supplied actions remain clean conditioning inputs.
Video-generation examples without future action annotations continue to contribute visual prediction
supervision, combining broad visual coverage with explicit action--outcome learning.

\paragraph{Training configuration}
We update the video-generation parameters and their input/output projections using AdamW~\citep{loshchilov2019adamw} with a
learning rate of $10^{-4}$. A 50-step warmup is followed by a constant learning rate. Video noise
levels are sampled using Waver. Training uses BF16 precision, sharded data parallelism, full
activation checkpointing, and separate packing budgets for understanding and video tokens.
Table~\ref{tab:wam-video-training} lists the optimizer and packing settings.

\begin{table}[t]
  \centering
  \small
  \caption{\textbf{Video-generation pretraining settings.} The adapted understanding tower is frozen, while the generation parameters and their input/output projections are updated. Token capacities are packing limits for variable-length examples.}
  \label{tab:wam-video-training}
  \begin{tabularx}{\linewidth}{@{}p{0.48\linewidth}X@{}}
    \toprule
    Setting & Video-generation pretraining \\
    \midrule
    Optimizer & AdamW \\
    Video-tower learning rate & $10^{-4}$ \\
    Learning-rate schedule & 50-step warmup, then constant \\
    AdamW coefficients & $(\beta_1,\beta_2)=(0.9,0.95)$ \\
    AdamW $\epsilon$ & $10^{-6}$ \\
    Weight decay & $0$ \\
    Gradient accumulation & $4$ \\
    Gradient-norm clipping & $0.1$ \\
    Understanding-token capacity per device & $2{,}176$ \\
    Video-token capacity per device & $65{,}536$ \\
    Video noise-level sampler & Waver \\
    \bottomrule
  \end{tabularx}

\end{table}

\subsubsection{Policy Model Pretraining}
\label{sec:wam-policy-pretraining}

Building on video pretraining, we train the policy to generate actions from observations and task
instructions.
The curriculum uses broad interaction data to learn everyday manipulation behaviors before
emphasizing the state and contact information available in robot demonstrations.

\begin{figure}[t]
  \centering
  \includegraphics[width=\linewidth]{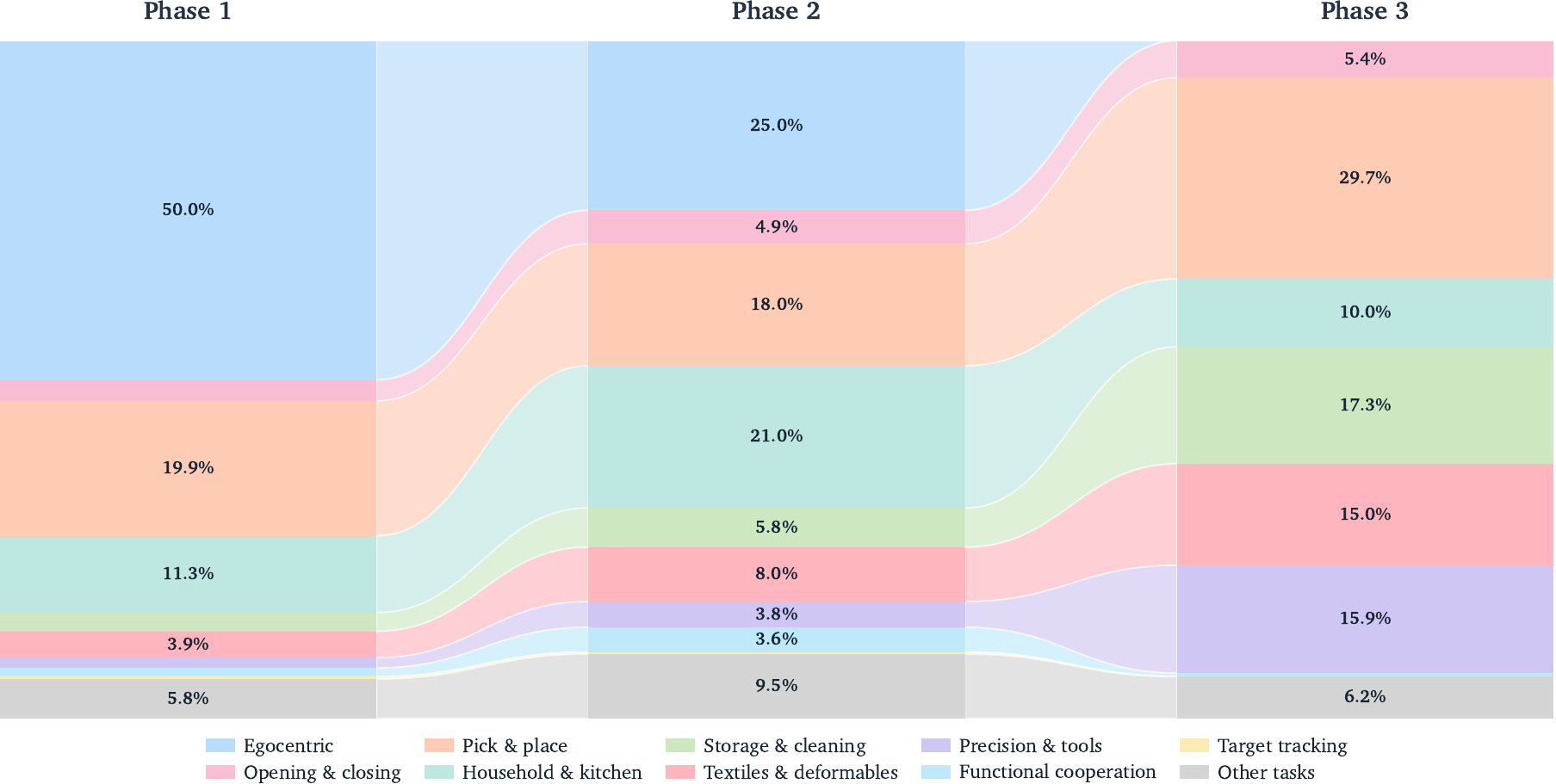}
  \caption{\textbf{Illustrative three-phase policy-training mixtures.} Band heights show estimated shares based on source weights and task annotations. Egocentric experience forms one group. Robot demonstrations, including those performed with dexterous hands, are grouped by task.}
  \label{fig:policy-task-mixture}
\end{figure}

\paragraph{A three-phase curriculum}
Starting from the pretrained world model, policy learning progresses through broad action learning,
state-conditioned action learning, and joint action--tactile learning. Action supervision remains
active throughout, and tactile supervision enters in the final phase.
Figure~\ref{fig:policy-task-mixture} illustrates how the curriculum shifts from egocentric
experience toward robot demonstrations, with pick-and-place and household tasks accounting for most
of the final mixture.

\paragraph{Phase 1: broad action learning}
We initially train action generation without explicit robot-state conditioning or tactile
prediction. Egocentric demonstrations provide broad coverage of everyday interactions, while robot
tasks follow the source distribution. This phase emphasizes language-conditioned motion and visual
cues to object interaction and task progress across diverse scenes.

\paragraph{Phase 2: state-conditioned action learning}
We introduce observed robot state as clean conditioning and reduce the share of egocentric
demonstrations. State conditioning supplies explicit robot-state information that may be ambiguous
from images alone. We also rebalance the remaining sources to increase coverage of task families
such as precision and tool use, storage and cleaning, textiles and deformables, and functional
cooperation. Both EEF and joint-space targets remain in the mixture where available and use the
corresponding domain-aware projections.

\paragraph{Phase 3: joint action and tactile learning}
The final phase retains action supervision and adds tactile prediction for demonstrations with
aligned tactile observations. The action and tactile streams share the action-generation tower and
use the joint supervision described in the architecture section. Examples without tactile
measurements continue to contribute action supervision; tactile losses apply only to valid targets.
Egocentric data are removed in this phase, and the mixture focuses on robot demonstrations,
particularly pick-and-place and household and kitchen tasks. Tactile-label availability is recorded
separately from task category, allowing tactile supervision to contribute across categories.

\paragraph{Recovery supervision and rollout reuse}
Curated system rollouts contribute to policy pretraining and selected supervised fine-tuning
settings. These rollouts often contain recovery processes, supplementing demonstrations with
corrective behavior from states reached during system execution.
The shared annotations in Section~\ref{sec:video-task-semantics} link each recovery segment to the
preceding unsuccessful attempt and, where available, the subsequent task continuation.
We assign higher sampling weights to validated subtask trajectories containing recovery behavior,
supervising corrective actions and continued execution with the same action objective used for other
demonstrations.

Before reuse, rollout trajectories pass through the shared annotation and policy data preparation
pipeline.
Successful task and recovery segments are then incorporated into training.
This reuse complements the agent's refinement of task context by incorporating validated interaction
experience into model training.

\paragraph{From interaction coverage to embodiment alignment}
This curriculum gradually shifts emphasis from diverse interaction experience to supervision aligned
with robot control. Egocentric sources provide broad coverage of objects, scenes, and motion,
whereas robot demonstrations provide action and state supervision in robot-specific control spaces,
together with contact observations where available. This progression is motivated by related
approaches that combine broad manipulation experience with subsequent control learning: VPP2 uses
manipulation-video pretraining before action learning~\citep{guo2026vpp2}, GR00T~2 transfers from
human and handheld demonstrations to robot embodiments~\citep{nvidia2026gr00t2}, and
Xiaomi-Robotics-1 follows broad pretraining on handheld manipulation trajectories with embodiment
and instruction alignment~\citep{xiaomi2026robotics1}. Our curriculum progressively reduces the
egocentric contribution as robot-specific state and contact supervision are introduced.

\begin{figure}[t]
  \centering
  \includegraphics[width=0.72\linewidth]{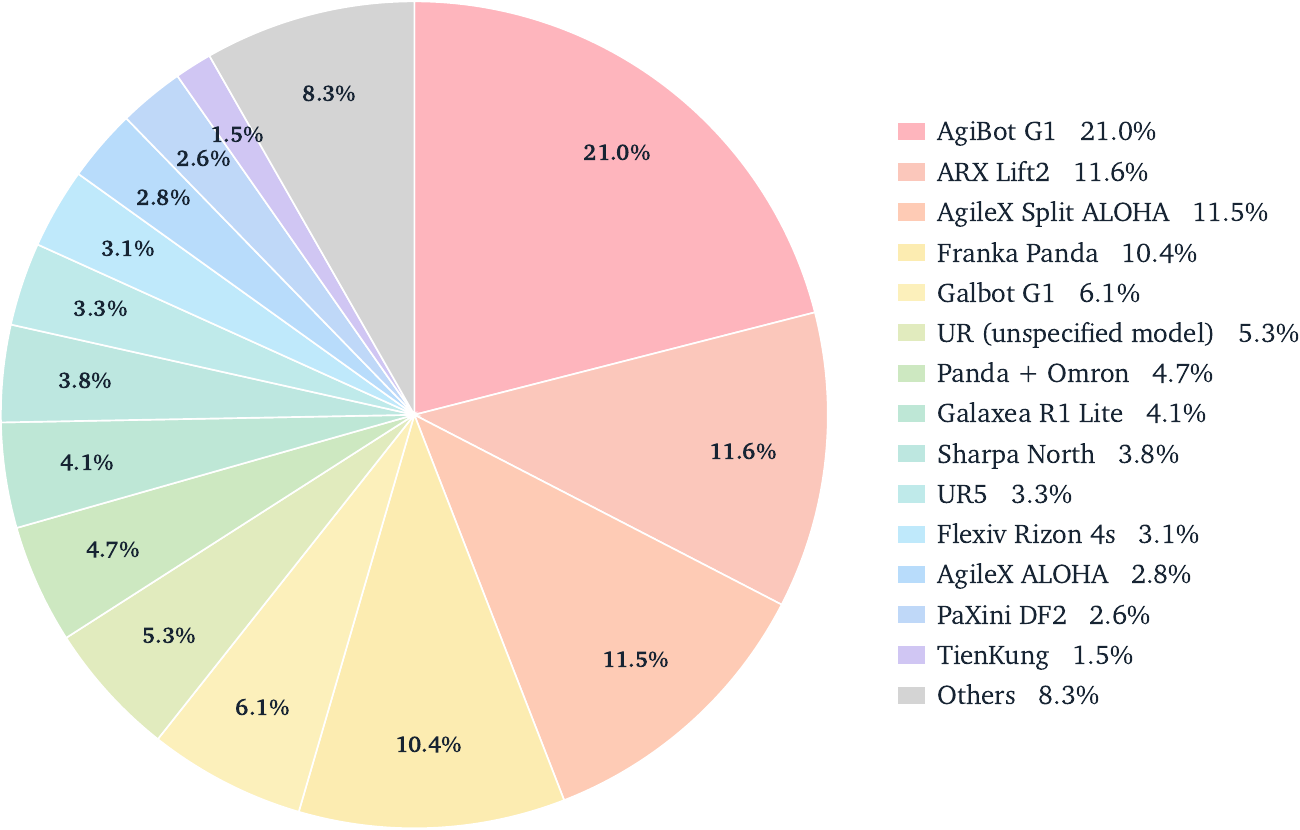}
  \caption{\textbf{Embodiment composition of the joint-space source pool.} Proportions are measured by source duration before filtering. The 14 largest groups are shown separately, with the remainder aggregated as Others. Corresponding real and simulated variants and end-effector variants are grouped according to the source legend. EEF and human egocentric representations are excluded. These proportions describe source-pool coverage, rather than the training-mixture shares in Figure~\ref{fig:policy-task-mixture}.}
  \label{fig:policy-embodiment-mixture}
\end{figure}

\paragraph{Embodiment coverage}
Figure~\ref{fig:policy-embodiment-mixture} describes the joint-space source pool, complementing the
training-mixture view. AgiBot~G1 (21.0\%), ARX~Lift2 (11.6\%), AgileX Split ALOHA (11.5\%), and
Franka Panda (10.4\%) are the largest embodiment groups, alongside mobile manipulators and dexterous
platforms. Domain-aware projections accommodate their distinct control representations within the
shared action Transformer.

\begin{table}[t]
  \centering
  \small
  \caption{\textbf{Settings for the state-conditioned policy configuration.} Video and action learning rates are assigned separately. Token capacities describe sequence-packing limits, so the number of examples per batch depends on sequence lengths.}
  \label{tab:wam-policy-training}
  \begin{tabularx}{\linewidth}{@{}p{0.48\linewidth}X@{}}
    \toprule
    Setting & State-conditioned policy training \\
    \midrule
    Video-tower learning rate & $10^{-5}$ \\
    Action-tower learning rate & $10^{-4}$ \\
    Learning-rate warmup & $100$ optimizer steps \\
    Gradient accumulation & $3$ \\
    Understanding-token capacity per device & $11{,}776$ \\
    Generation-token capacity per device & $62{,}592$ \\
    Reserved action-token capacity & $2{,}944$, within generation capacity \\
    Video / action loss weights & $1.0$ / $1.0$ \\
    Video / action noise-level sampler & Shifted uniform, shift $3$; sampled independently \\
    \bottomrule
  \end{tabularx}

\end{table}
\paragraph{Noise-level sampling}
Noise realizations and noise levels are sampled independently across video, and action targets when
present. For video and action in policy training, we independently draw $u_m\sim\mathcal{U}(0,1)$
and apply a shift of 3:
\begin{equation}
  \sigma_m=\frac{3u_m}{1+2u_m},
  \qquad m\in\{v,a\}.
  \label{eq:wam-policy-noise-sampling}
\end{equation}
This shifted-uniform distribution emphasizes higher noise levels while retaining low-noise coverage.
The emphasis on noisy action inputs is motivated by the timestep-sampling analysis in
$\pi_0$~\citep{physicalintelligence2024pi0} and the high-noise distributions studied in FLUX~3
Action~\citep{bfl2026flux3action}.
In Phase~3, tactile targets share the action stream's RF timestep, with independently sampled
Gaussian noise for each modality.

\paragraph{Hyperparameters}
During policy pretraining, the video and action towers use fixed learning rates of $10^{-5}$ and
$10^{-4}$, respectively. Training uses BF16 precision, full activation checkpointing, and packed
multimodal sequences. The state-conditioned configuration uses a 100-step warmup. Additional
settings are listed in Table~\ref{tab:wam-policy-training}. Token budgets constrain packed sequence
lengths, so the number of examples per batch varies with their lengths.

\section{Embodied Agent System}

\begin{figure}[!htbp]
  \centering
  \includegraphics[width=\linewidth]{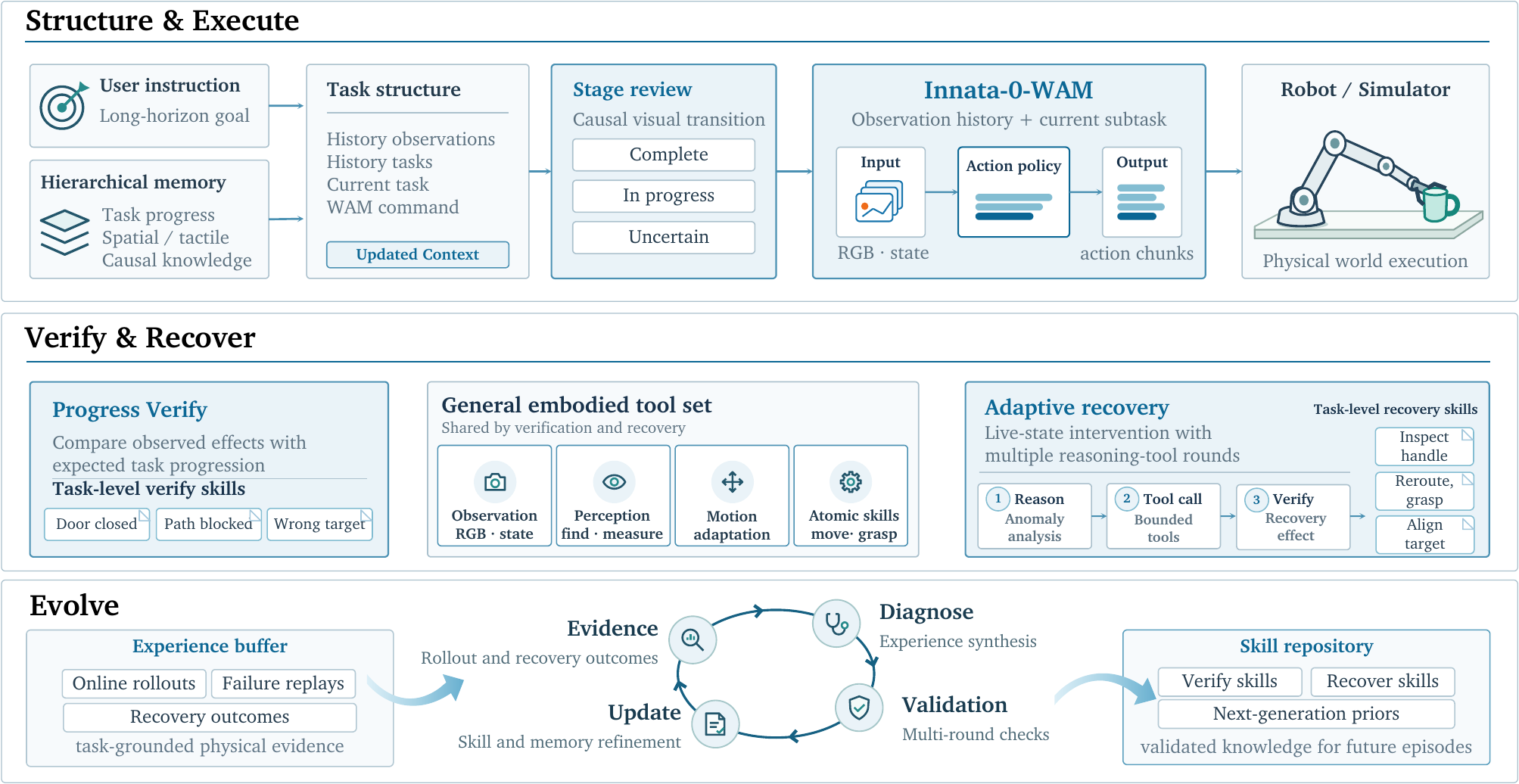}
  \caption{\textbf{Innata-0 agent runtime.}
    The agent grounds a long-horizon goal in hierarchical memory and maintains an updated task structure. Periodic stage reviews evaluate causal visual transitions and update task progress, while WAM predicts action chunks for execution in the robot or simulator. With the task-level verify skills, the verifier compares observed physical effects with the expected task progression using a general embodied tool set shared with recovery. When a deviation is verified, adaptive recovery performs iterative reasoning, bounded tool invocation, and outcome verification. Rollout and recovery evidence is subsequently synthesized and validated to refine task-level skills and provide reusable priors for future episodes.}
  \label{fig:agent_architecture}
\end{figure}

Innata-0-WAM provides the physical capabilities for understanding, prediction, and action
generation. Extending these capabilities to long-horizon tasks requires coordination beyond
individual model executions. We therefore develop \textbf{Innata-0-Agent} to support \emph{Plan},
\emph{Recover}, and \emph{Evolve}, following our co-design protocol.

For \emph{Plan}, the agent decomposes complex instructions into task structure based on the model's
capabilities and coordinates their execution.
It maintains the complete instruction, active subtask, and verified task history, providing
structured task context aligned with the model's training.
For \emph{Recover}, the agent verifies execution outcomes and identifies deviations from the
intended task progress. When failures occur, it invokes recovery tools to restore valid execution
conditions and coordinates subsequent execution to complete the remaining task.
For \emph{Evolve}, the agent accumulates validated interaction experiences and consolidates them
into reusable skills to improve future decisions and recovery strategies.
Successful rollout segments, including corrective actions and post-recovery continuations, are
further collected through the shared data pipeline for subsequent policy training. The overview of
the Innata-0-Agent runtime is shown in Figure~\ref{fig:agent_architecture}.

To support these capabilities within a unified execution framework, we develop an
\emph{Agent Infrastructure} that connects task structure, model execution, and tool-using. It
organizes execution through composable task graphs, provides unified access to models and tools, and
integrates physical feedback and safety mechanisms. Task isolation and resource reuse further
support reliable and efficient system operation.

\subsection{Task Structure and Execution}
\label{sec:task_structure_execution}

To support \emph{Plan}, Innata-0-Agent coordinates model execution across interdependent subtasks to
accomplish long-horizon tasks. Given a user instruction, the agent organizes the task based on the
model's execution capabilities and retrieves relevant hierarchical memory to construct the current
task context and WAM command. WAM then executes the assigned subtask through action chunks in the
robot or simulator. During execution, the agent periodically reviews task progress and updates the
active subtask and WAM command based on observed outcomes. This forms a closed loop of task
planning, physical execution, and progress-driven adaptation.

\subsubsection{Hierarchical Memory Architecture}
Long-horizon interaction requires the system to maintain an evolving task state rather than
repeatedly reasoning from the original instruction ~\cite{he2026most}.
We therefore introduce a \emph{hierarchical memory architecture} that progressively organizes
physical interaction evidence into task-relevant state and reusable knowledge. Together, these
memories provide a shared state substrate for task decomposition, planning, verification, recovery,
and model execution.
\paragraph{Task Memory}
During execution, the agent continuously compresses the interaction history into a task-centric
state that summarizes current progress, completed subtasks, unresolved dependencies, relevant object
states, detected anomalies, and recovery outcomes. This evolving summary provides the primary
context for long-horizon reasoning, avoiding the need to reconstruct the task state repeatedly from
the full execution history. At fixed intervals, the agent reviews the current task stage using the
latest observations and execution evidence, and updates the task memory accordingly.

\paragraph{Physical State Memory}
Physical state memory maintains a shared account of the evolving physical world through
complementary spatial and tactile components.
The spatial component maintains persistent object locations, scene structure, and spatial relations
beyond the current field of view.
It provides stable grounding for task-state tracking and allows planning and recovery to reason
about objects or regions that are temporarily occluded or no longer directly
observed~\citep{su2026mfm,xie2026lte}.
The tactile component contributes sparse but interaction-rich evidence about local contact,
particularly when interactions are visually ambiguous~\citep{bi2025vlatouch,zhou2026touchworld}.
The agent interprets this feedback together with visual and proprioceptive observations to retain
verified interaction states, such as whether an object remains held or has been released.
Together, the two components ground the task state in both the spatial arrangement of the scene and
ongoing physical interactions.

\paragraph{Causal Interaction Memory} At the highest abstraction level, validated interaction experience is consolidated into reusable knowledge about how actions transform physical states. Causal memory represents action-centered preconditions and effects over structured object states and relations~\citep{shen2026causground}:
\begin{equation}
  \text{Precondition}
  \rightarrow
  \text{Action}
  \rightarrow
  \text{Effect}.
\end{equation}
This abstraction allows experience obtained from previous interactions to be grounded to the current
state and reused for action selection, forward prediction, backward reasoning, and recovery without
requiring immediate model updates.
The memory hierarchy therefore supports a bidirectional information flow. During execution, spatial
and tactile evidence updates the agent-summarized task state, while validated state transitions are
progressively consolidated into causal knowledge. During decision making, relevant causal experience
is grounded back into the current task and physical state to guide planning, execution, and
recovery.

\subsubsection{Grounded Perception and Execution}

Building on hierarchical memory, the agent decomposes a long-horizon objective into causally ordered
subtasks and dispatches them to WAM for grounded execution. We further introduce a periodic
stage-review mechanism to automatically govern when the active subtask should be maintained or
advanced. At each review, the agent samples earlier keyframes and recent observations to analyze the
visible state transition after the robot or simulator executes an action chunk produced by WAM. This
temporal evidence is interpreted together with the within-episode task memory, available spatial and
tactile memory, and the subtask dependencies and temporal knowledge retrieved from long-term
hierarchical memory.

The reviewer then determines whether the expected physical effect of the current subtask is visually
supported, classifying its status as \emph{completed}, \emph{in progress}, or \emph{uncertain}. A
subtask transition is accepted only when completion is supported by sufficient evidence; the
completed stage is then added to the task history, and the next subtask and corresponding WAM
command are activated. Otherwise, WAM continues executing the current command while subsequent
reviews accumulate additional evidence. This process couples high-level task progression with
grounded execution without requiring WAM itself to infer the global task structure or decide when to
switch subtasks.

Every review is recorded, including the environment step, the stage index before and after the
review, whether the transition was accepted, and the complete structured decision returned by the
model. These records make the evolution of the task structure auditable and provide reusable
execution evidence for subsequent analysis and system improvement. The overall execution process
therefore alternates between WAM action generation and conservative, evidence-grounded stage review,
enabling long-horizon tasks to progress without relying on privileged simulator state or allowing
the model to advance the plan from unsupported assumptions.

\subsection{Verification and Adaptive Recovery}
\label{sec:verification-recovery}

To support \emph{Recover}, Innata-0-Agent handles uncertainties and unexpected failures during
real-world execution through physical outcome verification and adaptive recovery. Physical actions
may produce unintended outcomes due to partial observability, contact uncertainty, or changes in the
environment. To address these challenges, the agent continuously monitors execution progress through
environmental feedback and verifies whether the intended physical outcomes have been
achieved~\citep{innermonologue,agia2025unpacking}. When deviations occur, the agent analyzes their
causes and selects appropriate recovery strategies~\citep{agia2025unpacking,liu2023reflect}.  It
then invokes complementary embodied tools and skills to restore valid execution conditions.
Depending on the recovery outcome, subsequent execution proceeds through either WAM or
agent-controlled tools to complete the remaining task. This closed-loop process connects physical
outcome verification with adaptive recovery, allowing the system to sustain task progress under
uncertainty.

\subsubsection{Execution Monitoring and Intervention}

During nominal execution, WAM repeatedly predicts and executes action chunks, while the agent
evaluates task progress at scheduled verification checkpoints.
Each verification constructs a compact temporal evidence bundle from sampled historical keyframes
and recent observations, allowing the agent to compare the observed state transition with the
physical effect expected under the current task context.
The resulting decision describes whether an anomaly should be triggered, the corresponding failure
type, the supporting evidence, and the discrepancy between the observed and expected physical
changes. Slow progress or visual ambiguity alone is not treated as failure; intervention is
initiated only when positive temporal evidence indicates that execution has deviated from the
intended task progression.

In addition to periodic verification, the system adaptively increases the review frequency around
potentially high-risk interactions, such as changes in gripper state during grasping or object
transport. This event-driven scheduling concentrates reasoning on contact-sensitive phases without
continuously querying the reasoning model. Temporal cooldowns suppress repeated triggers caused by
the same persistent visual symptom, preventing redundant interventions. Once an anomaly has been
verified, WAM execution is suspended and control is transferred to the agent. This handoff occurs
directly in the current live environment. The agent inherits the same scene instance, object
configuration, robot joint and gripper states, environment step, and remaining execution budget.

\subsubsection{Tool-based Verification and Recovery}

To support recovery across different tasks and anomaly modes, we collect a general and extensible
tool set. Rather than directly generating low-level control code, the recovery agent selects tools,
instantiates their bounded parameters, and composes them into short closed-loop intervention
sequences.

The tool set is organized into four functional layers. \emph{Observation} tools expose task
information, robot state, synchronized visual observations, camera calibration, and previously
discovered scene entities without changing the environment. \emph{Perception} tools convert these
observations into actionable semantic and geometric representations, including target segmentation,
three-dimensional measurement, and transformations between image and world coordinates.
\emph{Motion adaptation} tools provide parameterized geometric adjustments that allow the agent to
retarget the end-effector, execute contact-aware arc motions, or reposition the mobile base
according to the current scene. Finally, \emph{atomic skills} encapsulate bounded and repeatedly
physical operations, such as actuating the gripper, returning the arm to a safe configuration,
maintaining a pose, or performing a geometry-conditioned rim grasp.

These layers support different stages of the closed-loop workflow. During verification, the agent
primarily uses observation and perception tools to inspect the evolving scene, ground task-relevant
entities, and determine whether the intended physical effect has occurred. Utilizing these tools, the
agent can gather additional evidence and reduce semantic or geometric uncertainty. Their outputs
supplement the temporal evidence used to distinguish normal progress, ambiguous observations, and
verified anomalies.

Once an anomaly has been verified, the same observation and perception interface provides the
grounded diagnosis and geometric targets required for recovery. The agent can then introduce
motion-adaptation tools and atomic skills to modify the live physical state. Candidate motions may
first be checked for feasibility before a bounded physical primitive is executed through the native
robot or simulator interface. Verification and recovery therefore share a consistent representation
of objects, geometry, and robot state, avoiding a separate perception pipeline at the point of
intervention.

Tool use follows an iterative observe--reason--act--verify process. The agent acquires evidence,
selects a tool according to the current task or diagnosed failure, executes the tool, and
incorporates its result into the next decision.
Fresh observations are collected after physical intervention to determine whether the intended local
change has occurred.
Unsuccessful actions, geometric constraints, and tool-level failure feedback are retained in the
interaction history, enabling the agent to revise parameters, select a different tool, or terminate
an ineffective strategy rather than repeatedly issuing the same intervention.

\subsection{Experience-Driven Evolution}
\label{sec:experience_evolution}
\emph{Evolve} preserves and builds on experience from earlier runs so that the agent can recognize similar failures and respond to them in later episodes. Execution records can help diagnose failures and guide corrective actions~\citep{liu2023reflect,ding2026zetta}, while skill libraries allow agents to retain successful behaviors for later use~\citep{wang2024voyager}. In Innata-0, the WAM generates actions for the current subtask. From the resulting observations, the agent learns which visual transitions indicate genuine failures. When the agent intervenes, the physical effects of its tool calls reveal effective tool compositions for that task.
A \emph{verify skill} captures observable signs of failure, while a \emph{recover skill} describes
which tool combinations work and when to use them. The agent maintains the two skills as versioned
artifacts for each task.
During \emph{experience accumulation}, the agent diagnoses failures and can replay them to test
candidate interventions. It uses new evidence to refine the existing skill pair and validates each
revision before promotion. During \emph{deployment}, it uses the latest validated pair for online
verification and live recovery and keeps that version fixed throughout the run. Finally, selected
rollout segments also provide supervision for subsequent WAM training.

\paragraph{Experience accumulation and initial skill synthesis}

Experience accumulation is performed on held-out task instances and scene configurations. Each
rollout records synchronized observations, WAM actions, verification decisions, tool calls, physical
interventions, and final outcomes.
Generation zero begins without task-specific verify or recover experience. The system executes tasks
with native WAM, producing an initial collection of successful and failed trajectories. Failed
episodes are then analyzed retrospectively to identify the earliest causally observable divergence,
the physical mechanism underlying the failure, and a state at which intervention remains feasible.
Although the final episode outcome can be used during hindsight diagnosis, the resulting failure
signature must be expressible using only observations available at the proposed verification point.
This ensures that knowledge derived retrospectively remains valid for causal online use.

During experience accumulation, failed trajectories may be replayed to reconstruct the diagnosed
state and test candidate recovery strategies. The diagnostic agent can vary tool selection,
geometric targets, motion parameters, and intervention timing while observing their physical
effects.
A candidate strategy is retained only when it executes valid physical actions and produces the
intended task-consistent outcome. Failed recovery attempts are also preserved because they reveal
unsuitable tools, unreachable geometries, unsafe intervention points, and conditions under which a
strategy
should not be applied.

The resulting evidence is consolidated into the initial task-level skill pair. Repeated failure
observations are summarized into a compact verify skill, while validated intervention traces and
their applicability conditions form the corresponding recover skill. Seed-specific coordinates and
object instances are replaced with semantic references, relative geometry, and
observable preconditions so that the resulting skills can transfer to new scene configurations.

\paragraph{Generational skill evolution}
After the initial skill synthesis, evolution proceeds in generations. At each generation, the verify
and recover skills from the preceding generation are consolidated and used on new task instances.
The verify skill guides online anomaly identification, while the recover skill provides
task-specific references for selecting and composing tools when an anomaly is confirmed. The system
therefore evaluates not only whether an existing skill can reproduce a previously successful
intervention, but also whether it generalizes to new object configurations, robot states, and
failure variations.

New rollouts produce additional evidence. Successful executions refine the range of nominal state
transitions and reduce false interventions, while failed verification and recovery attempts expose
missing signatures, incorrect applicability conditions, or insufficient recovery strategies.
During this accumulation phase, newly failed episodes may again be diagnosed and replayed to develop
candidate revisions. At the end of each generation, the existing task-level verify and recover
skills are consolidated with two sources of newly validated experience: evidence collected during
closed-loop online execution and diagnostic or recovery experience obtained through failure replay.
The resulting revisions form the candidate skill set for the next generation, subject to validation
on fresh task configurations.

Consolidation does not simply append complete trajectories. It extracts the minimal reusable
knowledge needed by the runtime agent, merges equivalent failure signatures and tool flows, records
both positive and negative applicability evidence, and preserves links to the supporting episodes
for audit.
Each skill revision is versioned and evaluated on fresh held-out configurations before promotion.
Once an evaluation instance contributes to skill revision, it becomes development experience and is
not reused as held-out evidence for the next generation.

Through successive generations, verify skills become more precise about when an apparent deviation
warrants intervention, while recover skills acquire a broader repertoire of strategies and clearer
conditions for selecting among them. Skill evolution therefore expands task-level competence without
requiring every contingency to be absorbed into the WAM policy parameters.

\paragraph{Deployment with accumulated skills}
Deployment follows a stricter protocol than experience accumulation. For the current task, the agent
retrieves only the latest validated verify and recover skills and keeps them frozen throughout
execution. The verify skill is conditioned on the current task state and causal observation history
to identify whether the physical execution has deviated from expectation. When sufficient evidence
supports an anomaly, the corresponding recover skill is provided to the recovery agent as
task-specific guidance for composing the general embodied tools.

Recovery during deployment operates directly on the current live environment state. Failed episodes
are not reset or replayed, and alternative recovery strategies cannot be tested by reconstructing an
earlier state. This constraint ensures genuinely online intervention rather than repeated offline
search. Deployment trajectories are still recorded and can later enter a new experience-accumulation
cycle, but they do not modify the active skills within the current evaluation generation. This
separation prevents unvalidated episode-specific changes from altering runtime behavior while
preserving a clear path for subsequent skill evolution.

\subsection{Agent Infrastructure}
\label{sec:embodied-infrastructure}
\begin{figure}[!t]
  \centering
  \includegraphics[width=\linewidth]{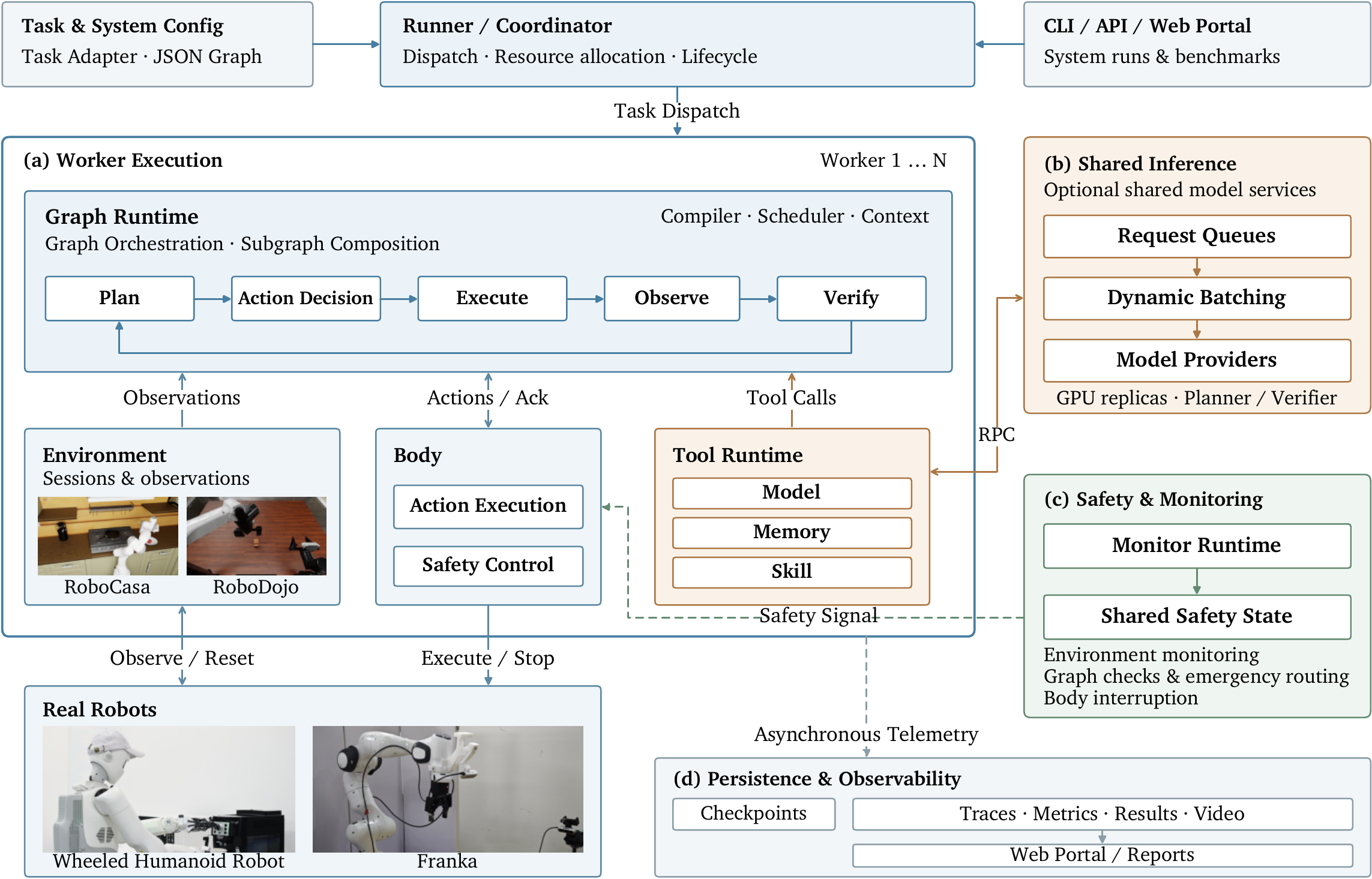}
  \caption{\textbf{Agent infrastructure and execution flow.} The coordinator dispatches tasks to workers. Within a worker, the example graph connects planning, action generation, execution, observation, and verification; feedback determines subsequent decisions. Nodes invoke models, memory, and skills through the tool runtime. The environment supplies observations, while the body executes actions through the physical integration. Shared inference services allow workers to reuse model resources. Safety monitoring affects both graph progression and action execution, and execution records feed the web portal and reports. Solid arrows show the indicated execution, data, or service interactions; dashed paths carry safety signals and telemetry. Platform photographs illustrate simulator and robot categories.}
  \label{fig:architecture}
\end{figure}

As robot actions change the environment, the agent must use their observed effects to decide how to
proceed. We organize this feedback loop as a task graph (Figure~\ref{fig:architecture}), following
the interleaving of reasoning and action in ReAct~\citep{yao2023react} and Inner
Monologue~\citep{innermonologue}. The graph specifies how planning, action generation, execution,
observation, and verification depend on one another.
For \emph{Plan}, the agent selects the current subtask and sends the corresponding command to WAM.
Physical interfaces pass the actions WAM generates to the robot or simulator and return observations
to the agent.
The agent verifies the physical outcome and pauses WAM to intervene with recovery tools if it
confirms an anomaly (\emph{Recover}).
The graph uses the verification result to choose the next step, and the runtime retains the full
execution record to support later skill revisions (\emph{Evolve}).
Our design keeps task logic in the graph so the same workflow can run with different models, skills,
and robot platforms through compatible interfaces.

\paragraph{Task graphs}
We borrow LangGraph's organization of workflows through shared state, nodes, and conditional
transitions~\citep{langgraph}, using our own runtime for embodied execution. Nodes represent
operations such as planning, action generation, and verification. Their connections specify which
results must be available and where different outcomes lead. The compiler resolves the graph
definition, and the scheduler runs nodes according to these dependencies. Task context retains the
overall goal, current progress, and intermediate results so that subsequent decisions can build on
earlier execution. A complex step can be expanded into a subgraph, for example by separating
observation analysis from goal checking during verification. Each subgraph receives its own inputs,
can access the overall task context, and keeps its internal results separate. The parent uses the
subgraph's defined output rather than its internal node results. A verification procedure can
therefore gain an additional check without changing how the parent calls it or uses its judgment.

\paragraph{Tools and memory}
The graph determines when a capability is needed and how its result affects the task; the tool
runtime provides access to that capability. Planning and verification can invoke the same reasoning
model with different inputs and criteria, while an action node can use a policy or an existing
skill. Replacing that policy does not change the node's role in the workflow, provided the
replacement preserves the expected inputs and outputs. Memory participates through the same tool
interface. Task context holds information needed during the current execution, whereas persistent
memory retains information across tasks. Following the use of retrieved experience in
memory-augmented agents~\citep{park2023generative}, a node can retrieve relevant information for its
current decision. Storage, selection, and summarization remain part of the memory implementation.
The task graph specifies when this information is needed and how it is used, without embedding the
memory procedure in every calling node.

\paragraph{Physical feedback and safety}
Once an action is generated, the body executes it and reports its status, while the environment
supplies observations of the resulting scene. Verification combines these two sources because
command completion and task success are different events. A gripper command may finish even though
the object has not been grasped; the next observation is needed to determine whether to continue or
attempt recovery. Verification returns this judgment to the graph, which selects the next step. Body
and environment interfaces expose these interactions consistently across simulators and robots, with
platform adapters handling coordinate frames, sensor formats, and action semantics. Safety
monitoring acts on both graph execution and body control through a shared safety state. Body control
can stop an ongoing action, and the graph can change the subsequent path in response to the same
condition, so that task progression accounts for the interruption.

\paragraph{Runtime and shared resources}
The coordinator dispatches configured tasks to workers. Each worker maintains its own task context
and interaction session with the environment or robot, while models, skills, and memory services can
remain resident for later tasks. This separates the lifetime of a task from the resources it uses:
ending one task need not unload a model that the next task still requires. Concurrent workers can
also use shared model services, which queue and batch compatible inference
requests~\citep{orca,vllm}. The infrastructure maintains the association between requests and tasks,
while the model service handles inference scheduling. Results return to the requesting workers, each
of which continues according to its own observations and feedback. Checkpoints, traces, metrics,
results, and video record this execution process. A visualization web interface provides access to
these records for inspecting task progress and relating decisions to their execution outcomes.

\FloatBarrier
\section{Evaluation}

We evaluate Innata-0's embodied capabilities and its performance in task execution.
Understanding benchmarks assess semantic and spatial grounding, while
WorldArena~2.0~\citep{li2026worldarena2} evaluates action-conditioned video prediction.
On RoboCasa365~\citep{robocasa365} and RoboDojo-Sim~\citep{chen2026robodojo}, we evaluate AGENT+WAM
configurations in closed-loop tasks, examining how task context, physical-outcome verification, and
validated experience guide execution and recovery.
UniVTAC~\citep{chen2026univtac} and Bench2Dex~\citep{yang2026bench2dex} further assess
tactile-conditioned control in gripper manipulation and multi-stage dexterous tasks.

\subsection{Understanding Evaluation}

\begin{table*}[t]
  \centering
  \footnotesize
  \setlength{\tabcolsep}{2pt}
  \renewcommand{\arraystretch}{1.12}

  \caption{\textbf{Understanding performance across 18 benchmarks.}
    We compare our model with seven reference models.
    Bold and underlined values indicate the best and second-best
    scores in each row, respectively.
    All scores are reported on a 0--100 scale, with higher being better.}
  \label{tab:understanding_results}

  \begin{tabular*}{\textwidth}{
      @{}l@{\extracolsep{\fill}}*{8}{c}@{}
    }
    \toprule

    \multirow{2}{*}{Benchmark} &
    Gemma 4 &
    Qwen3-VL &
    Embodied-R1.5 &
    MiMo-Embodied &
    RoboBrain 2.5 &
    RynnBrain &
    Cosmos 3 Nano &
    \textbf{Ours} \\

    &
    E4B~\citep{team2026gemma} &
    8B~\citep{bai2025qwen30vl} &
    8B~\citep{yuan2026embodied0r1050} &
    7B~\citep{hao2025mimo0embodied0} &
    8B~\citep{tan2026robobrain} &
    8B~\citep{dang2026rynnbrain0} &
    8B~\citep{nvidia2026cosmos3} &
    \textbf{8B} \\

    \midrule
    \multicolumn{9}{@{}l}{
      \textit{General visual understanding}
    } \\

    BlinkDepth~\citep{fu2024blink0}
    & 71.77 & 83.87 & 86.29 & 78.23 & 83.06
    & \underline{91.13} & \textbf{93.55} & 89.52 \\

    BlinkSpatial~\citep{fu2024blink0}
    & 77.62 & \underline{83.22} & 75.52 & 76.92
    & \textbf{86.01} & 78.32 & 81.12 & 79.02 \\

    CausalVQA~\citep{foss2025causalvqa0}
    & 43.00 & \underline{65.00} & 56.00 & 54.00
    & 48.00 & 46.00 & 63.00 & \textbf{69.00} \\

    CountBenchQA~\citep{beyer2024paligemma0}
    & 55.65 & 90.35 & 77.21 & 79.47
    & 89.12 & \underline{90.55} & 85.01 & \textbf{92.81} \\

    RealWorldQA~\citep{xai2024grok15vision}
    & 49.02 & 71.11 & 69.02 & 63.27
    & \underline{73.20} & 70.33 & 72.55 & \textbf{73.86} \\

    RefCOCO~\citep{yu2016modeling}
    & 72.09 & \textbf{87.79} & 75.01 & 78.91
    & 78.23 & 83.34 & 84.02 & \underline{86.14} \\

    V-Star Bench~\citep{wu2023v00}
    & 23.56 & 82.72 & 81.68 & 48.17
    & 57.59 & 85.34 & \underline{86.91} & \textbf{89.53} \\

    CV-Bench~\citep{tong2024cambrian010}
    & 52.91 & 82.34 & 86.95 & 77.12
    & 86.06 & \underline{88.15} & 87.96 & \textbf{89.18} \\

    \midrule
    \multicolumn{9}{@{}l}{
      \textit{Embodied understanding and grounding}
    } \\

    3DSRBench~\citep{Ma_2025_ICCV}
    & 49.01 & \underline{56.60} & 56.25 & 54.36
    & 49.48 & 56.14 & 55.05 & \textbf{58.38} \\

    EmbSpatial-Bench~\citep{du-etal-2024-embspatial}
    & 62.75 & 77.94 & 75.16 & 76.07
    & 77.72 & 79.20 & \underline{81.59} & \textbf{84.42} \\

    ERQA~\citep{gemini2025robotics}
    & 38.25 & 43.00 & 42.00 & 38.25
    & 44.75 & 27.75 & \underline{48.00} & \textbf{48.50} \\

    MindCube~\citep{yin2025spatial}
    & 38.36 & 29.70 & 34.48 & 44.80
    & 27.74 & \textbf{57.63} & 40.45 & \underline{50.92} \\

    MMSI-Bench~\citep{yang2025mmsi0bench0}
    & 27.70 & 31.40 & 29.80 & 25.90
    & 28.90 & \textbf{39.30} & \underline{36.10} & 34.70 \\

    MMSI-Video-Bench~\citep{lin2025mmsi0video0bench0}
    & 22.33 & 29.93 & 28.12 & 29.66
    & \textbf{30.56} & 30.02 & 26.40 & \underline{30.47} \\

    RoboAfford~\citep{hao2025roboafford000}
    & 30.78 & 51.67 & 72.57 & 62.02
    & \textbf{77.17} & \underline{76.04} & 70.67 & 73.97 \\

    SPARBench~\citep{zhang2025flatland}
    & 26.44 & 39.29 & 40.33 & 16.27
    & 42.34 & 49.47 & \underline{55.02} & \textbf{69.08} \\

    SWITCH~\citep{lin2025switch0}
    & 24.87 & 23.27 & 21.93 & \textbf{26.53}
    & 26.33 & 21.93 & 20.80 & \underline{26.47} \\

    VSI-Bench~\citep{Yang_2025_CVPR}
    & 31.65 & 55.46 & 56.15 & 46.70
    & 43.77 & \textbf{62.98} & 55.00 & \underline{61.60} \\

    \bottomrule
  \end{tabular*}

\end{table*}

We report results on
18 benchmarks spanning general visual understanding, embodied
reasoning, and spatial grounding. Table~\ref{tab:understanding_results}
compares the understanding tower with seven reference models.
Across the 18 reported benchmarks, our model ranks first on nine
and within the top two on fourteen among the models listed.
It achieves 69.08 on SPARBench, 84.42 on EmbSpatial-Bench, and
48.50 on ERQA. Relative to the Qwen3-VL-8B~\citep{bai2025qwen30vl} initialization, it
improves performance on 16 of the 18 benchmarks, including gains
of 29.79 points on SPARBench, 22.30 points on RoboAfford, and
6.14 points on VSI-Bench. Improvements also extend to general
visual tasks, including CountBenchQA, RealWorldQA, and V-Star Bench.

\ifnum\value{subsubsection}=0\relax\stepcounter{subsubsection}\fi
\subsection{Video Generation}
\label{sec:video-generation-evaluation}

We evaluate video prediction on WorldArena 2.0 Track~1~\citep{li2026worldarena2}. The public test
set contains 1,000 robot manipulation episodes drawn from RoboTwin 2.0~\citep{chen2025robotwin}
trajectories. Each episode provides an initial RGB frame, a language instruction, and a robot action
trajectory. The task is to predict the video that follows the initial frame.

\paragraph{Evaluation metrics}
We report 15 normalized metrics across six categories: visual quality, motion quality, content
consistency, physics adherence, 3D accuracy, and controllability. Scores range from 0 to 100; higher
is better. For the models we evaluate, we cap Dynamic Degree, Flow Score, and Motion Smoothness at
their respective ground-truth values for each episode before aggregating across episodes. We compute
the overall EWM score as the unweighted mean of the 15 component scores.

\paragraph{Baselines and inference protocol}
We evaluate our model and the released RoboTwin~2.0 post-trained checkpoints of
ABot-PhysWorld~\citep{amap2026abotphysworld}, FlowWAM~\citep{chen2026flowwam}, and
BWM~\citep{bwm2026simulator} with the same official Track~1 evaluation code. For all four models, we
use Real-ESRGAN~\citep{wang2021realesrgan} to upscale the initial frame to \(640 \times 480\). Our
model conditions on the action trajectory; each baseline uses the inputs supported by its released
checkpoint. For models whose evaluated checkpoints are closed or unreleased,
Table~\ref{tab:worldarena-comparison} lists the scores from the original papers without recomputing
them.

\paragraph{Results}
Our model achieves an EWM score of 67.63, ranking first among the open-source models and models with
publicly reported results included in our comparison (Table~\ref{tab:worldarena-comparison}). It
exceeds BWM, FlowWAM, and ABot-PhysWorld by 1.48, 1.77, and 6.44 points, respectively. Relative to
the best local baseline on each metric, the largest gains are in Image Quality (67.49, \(+13.18\))
and Aesthetic Quality (51.02, \(+9.17\)). Our model also leads this group in Photometric Consistency
(44.73) and Perspectivity (95.68). Its Trajectory Accuracy of 62.81 is 3.54 points above BWM and
1.25 below FlowWAM. Background Consistency and Instruction Following remain below the strongest
baselines on those metrics.

\begin{table}[!htbp]
  \centering
  \caption{\textbf{WorldArena 2.0 Track~1 video prediction.} We evaluated our model and the released RoboTwin~2.0 post-trained checkpoints (\(\dagger\)); scores marked \(\ddagger\) come from the original papers. Bold and underline mark the best and second-best EWM among the four models we evaluated.}
  \label{tab:worldarena-comparison}
  \begingroup
    \fontsize{7.5}{9.0}\selectfont
    \setlength{\tabcolsep}{1.2pt}
    \renewcommand{\arraystretch}{1.12}
    \begin{tabular*}{\linewidth}{@{\extracolsep{\fill}}>{\raggedright\arraybackslash}p{2.75cm}>{\centering\arraybackslash}p{0.95cm}*{15}{>{\centering\arraybackslash}p{0.72cm}}@{}}
      \toprule
      \multirow{3}{*}{Model} & \multirow{3}{*}{EWM} & \multicolumn{3}{c}{\makecell{Visual\\Quality}} & \multicolumn{3}{c}{\makecell{Motion\\Quality}} & \multicolumn{3}{c}{\makecell{Content\\Consistency}} & \multicolumn{2}{c}{\makecell{Physics\\Adherence}} & \multicolumn{2}{c}{\makecell{3D\\Accuracy}} & \multicolumn{2}{c}{\fontsize{6.8}{8.0}\selectfont Controllability} \\
      \cmidrule(lr){3-5}\cmidrule(lr){6-8}\cmidrule(lr){9-11}\cmidrule(lr){12-13}\cmidrule(lr){14-15}\cmidrule(lr){16-17}
      & & IQ & AQ & JS & DD & FS & MS & SC & BC & PC & IN & TA & DA & PV & IF & SA \\
      \midrule
      \multicolumn{17}{@{}l}{\emph{Evaluated in this work}} \\
      ABot-PhysWorld\textsuperscript{$\dagger$}~\citep{amap2026abotphysworld} & 61.19 & 54.31 & 40.66 & 89.95 & 22.44 & 12.02 & 67.10 & 83.28 & 89.48 & 18.65 & 70.20 & 29.35 & 81.70 & 93.66 & 75.92 & 89.20 \\
      FlowWAM\textsuperscript{$\dagger$}~\citep{chen2026flowwam} & 65.86 & 48.95 & 40.84 & 97.31 & 21.89 & 11.77 & 66.77 & 79.51 & 86.85 & 29.60 & 73.70 & 64.06 & 98.98 & 94.74 & 83.06 & 89.92 \\
      BWM\textsuperscript{$\dagger$}~\citep{bwm2026simulator} & \underline{66.15} & 52.08 & 41.85 & 97.77 & 22.28 & 11.52 & 67.30 & 76.47 & 82.88 & 40.55 & 74.70 & 59.27 & 97.82 & 94.78 & 83.12 & 89.91 \\
      \textbf{Ours} & \textbf{67.63} & 67.49 & 51.02 & 95.37 & 22.01 & 11.48 & 65.18 & 76.35 & 81.14 & 44.73 & 72.28 & 62.81 & 98.29 & 95.68 & 80.80 & 89.88 \\
      \midrule
      \multicolumn{17}{@{}l}{\emph{Reported in the original papers}} \\
      DreamX-Phi 1\textsuperscript{$\ddagger$}~\citep{dreamx2026phi} & 60.65 & 63.25 & 43.38 & 92.93 & 22.90 & 5.81 & 63.26 & 71.93 & 84.52 & 14.29 & 57.36 & 57.15 & 98.55 & 82.24 & 61.62 & 90.53 \\
      CtrlWAM\textsuperscript{$\ddagger$}~\citep{peng2026ctrlwam} & 61.75 & 39.40 & 37.10 & 96.30 & 21.20 & 10.90 & 66.00 & 73.70 & 78.50 & 47.40 & 66.80 & 41.20 & 95.50 & 88.70 & 73.80 & 89.90 \\
      Ropedia W0\textsuperscript{$\ddagger$}~\citep{liu2026ropedia} & 66.92 & 53.57 & 41.04 & 96.00 & 25.48 & 15.75 & 67.55 & 79.51 & 84.17 & 27.98 & 77.42 & 63.78 & 98.56 & 97.72 & 85.50 & 89.73 \\
      \bottomrule
    \end{tabular*}
    \par\smallskip
    \begin{minipage}{\linewidth}
      \raggedright\fontsize{7.5}{9}\selectfont
      \textit{Metric abbreviations.} IQ: Image Quality; AQ: Aesthetic Quality; JS: JEPA Similarity; DD: Dynamic Degree; FS: Flow Score; MS: Motion Smoothness; SC: Subject Consistency; BC: Background Consistency; PC: Photometric Consistency; IN: Interaction Quality; TA: Trajectory Accuracy; DA: Depth Accuracy; PV: Perspectivity; IF: Instruction Following; SA: Semantic Alignment.
    \end{minipage}
  \endgroup
\end{table}
\FloatBarrier

\subsection{Policy Evaluation}
\label{sec:policy-evaluation}

We adapt our pretrained model to RoboCasa365~\citep{robocasa365} and
RoboDojo-Sim~\citep{chen2026robodojo} through benchmark-specific supervised
fine-tuning (SFT) and evaluate the resulting policies in closed-loop
simulation.

\paragraph{SFT setting}
For RoboCasa365, we fine-tune on Human300 demonstrations, rebalancing task
sampling according to the number of available training windows per task:
tasks with more windows are downweighted, while those with fewer are
upweighted. This increases the relative exposure to atomic tasks,
complementing the subtask decomposition and conditioning used during
pretraining. To support both standalone execution and integration with a
planning system using the same policy, we mix 70\% command-only examples
with 30\% examples augmented by \texttt{current\_plan} (string) and
\texttt{history\_plan} (list), derived from VLM subtask segmentation and
annotation. We train for 26k steps with a global batch size of 1,848,
conditioning on the current multi-view RGB observation and robot state
(H1).

For RoboDojo, we initialize from our pretrained WAM checkpoint and fine-tune on RoboDojo-Sim
demonstrations. To reduce task imbalance, we first sample uniformly from 35 task groups and then
sample trajectories within each group. Each example contains the original task command together with
\texttt{current\_plan} and \texttt{history\_plan}. The policy conditions on H5 recent multi-view RGB
observations, as well as the current robot state.
For Memory tasks, the oldest observation is
replaced by the episode-initial observation. The model predicts 16 future video frames and
corresponding 14D joint-control targets at 25\,Hz. We train for 24k steps with a global batch size
of 1,680.
Table~\ref{tab:policy-sft-settings} summarizes the training settings.

\begin{table}[!htbp]
  \centering
  \small
  \setlength{\tabcolsep}{5pt}
  \renewcommand{\arraystretch}{1.15}
  \caption{\textbf{Policy SFT settings.} RoboCasa365 uses Human300 demonstrations only,
    excluding MG60, Target Human50, and other RoboCasa365 demonstration
    collections. JPEG quality refers to stored training images.}
  \label{tab:policy-sft-settings}
  \begin{tabularx}{\linewidth}{@{}>{\raggedright\arraybackslash}p{0.25\linewidth}>{\raggedright\arraybackslash}X>{\raggedright\arraybackslash}X@{}}
    \toprule
    Setting & RoboCasa365 & RoboDojo \\
    \midrule
    SFT data
    & Human300
    & RoboDojo-Sim trajectories \\

    Initialization
    & Pretrained model
    & Pretrained model \\

    Training steps / global batch size
    & 26k / 1,848
    & 24k / 1,680 \\

    Conditioning
    & Language, multi-view RGB, robot state
    & Language, multi-view RGB, robot state \\

    Image storage
    & JPEG, quality 95
    & JPEG, quality 95 \\

    Multi-view processing
    & Views composed into one RGB canvas
    & Views composed into one RGB canvas \\

    Observation history
    & H1: current images and robot state
    & H5: multi-view RGB at five recent time points and current robot state; oldest observation replaced by the episode-initial observation for Memory tasks \\

    Prediction horizon
    & 16 future actions and corresponding video
    & 16 future actions and corresponding video \\

    Action interface
    & 12D control commands; masked 128D internal representation
    & 14D joint targets; masked 128D internal representation \\

    Conditioning loss mask
    & Observed images and robot state excluded
    & Observed images and robot state excluded \\

    Video / action learning rate
    & $10^{-5}$ / $10^{-4}$
    & $10^{-5}$ / $10^{-4}$ \\

    LR decay
    & Cosine
    & Constant through 10k steps, followed by linear decay \\

    EMA
    & Full MoT network; momentum $0.995$
    & Full MoT network; momentum $0.995$ \\

    Video timestep sampling
    & Waver; resolution-dependent shift
    & Uniform shifted distribution; shift $5$ \\

    Action timestep sampling
    & Uniform base distribution; shift $3$, independent of video
    & Shared with the video timestep \\

    Training objective
    & Video and action flow matching
    & Video and action flow matching \\

    Text dropout
    & $0.1$
    & $0.0$ \\
    \bottomrule
  \end{tabularx}
\end{table}

\paragraph{Evaluation setting}
On RoboCasa365, we evaluate the 26k-step EMA checkpoint on 50 tasks with
50 fixed initial conditions per task, totaling 2,500 episodes: 900 Atomic-Seen, 800
Composite-Seen, and 800 Composite-Unseen. The policy receives a detailed
task command expanded using the initial observation; neither
\texttt{current\_plan} nor \texttt{history\_plan} is supplied, and no online
planner updates the guidance during execution. Each camera image undergoes
JPEG quality-95 encoding and decoding before view composition. Inference
uses 30 denoising steps, CFG~2, and timestep shift~5, with independent video
and action noise and an updated sampling seed at each decision. The policy
executes 16 actions at 20\,Hz before receiving the next observation. Success
is checked at every control step, with task-specific episode horizons.

On RoboDojo-Sim, we evaluate a VLM + WAM system on 42 official
tasks represented by 54 evaluation entries. A VLM first generates
an ordered sequence of subtask plans from the initial observation
and task instruction. During execution, a second VLM assesses
whether the current subtask has been completed, allowing the
system to advance by at most one plan per check; it does not
generate, rewrite, or reorder the plans.
The WAM receives the original command, current plan, completed-plan history, H5 multi-view RGB
observations, and the current robot state. Images undergo JPEG quality-95 encoding and decoding
before view composition. Inference predicts 16 actions and executes 8 at 25\,Hz, using 30 UniPC~\citep{zhao2023unipc}
denoising steps, guidance scale~1, timestep shift~5, and a new sampling seed for each action chunk.
We report Score and success rate, averaging each metric equally across the five capability
dimensions.

For full-system evaluation (AGENT+WAM), we collect skills through four generations of
development-time interaction on both RoboCasa365 and RoboDojo-Sim. For each task, we randomly sample
five scene initializations per generation, identified by $\mathrm{seed}$ or
$(\mathrm{seed}, \mathrm{layout})$. The layouts, object placements, and other randomized scene
properties are instantiated according to the benchmarks from these identifiers. Note that these
scenes are disjoint from those formal deployment ones. After all episodes in a generation terminate,
their successful executions, observable failure patterns, hindsight diagnoses, and recovery outcomes
are consolidated into a new pair of task-level verify and recover skills. Deployment uses the
validated skill pair from the final generation. Following the RoboDojo benchmark, we exclude those episodes from both score and success rate due to simulation instability or scene initialization failures.

\begin{table}[!htbp]
  \centering
  \small
  \setlength{\tabcolsep}{4pt}
  \renewcommand{\arraystretch}{1.12}
  \caption{\textbf{RoboCasa365 success rates (\%).} Reference scores follow their
    published protocols and are taken from the
    \href{https://robocasa.ai/leaderboard.html}{public leaderboard}
    (November 18, 2026), except the GR00T~2 results, from Table~19 of
    its report~\citep{nvidia2026gr00t2}.
    The shaded row ($\dagger$) uses the WAM as the default executor for
    23 of 50 tasks and ASPIRE-generated programs for the remaining 27,
    with code-as-policy recovery available in both cases.}
  \label{tab:robocasa-policy-results}

  \begin{tabularx}{\linewidth}{
      >{\raggedright\arraybackslash}X
      lrrrr
    }
    \toprule
    Model & Type & Overall & Atomic-Seen & Composite-Seen & Composite-Unseen \\
    \midrule

    GR00T~2--SFT~\citep{nvidia2026gr00t2}
    & WAM & \textbf{67.2} & 87.4 & \textbf{73.6} & 38.2 \\

    Awomo-0.5V~\citep{awomo2026technicalreport}
    & WAM & 62.5 & 85.0 & 60.4 & \textbf{39.4} \\

    Axiom-0~\citep{shu2026axiom0}
    & VLA & 61.6 & 86.5 & 59.4 & 35.8 \\

    RPent~\citep{zhang2026harnessvla}
    & AGENT+VLA & 59.2 & 87.8 & 43.8 & 42.5 \\

    WorldDreamer v5~\citep{wu2026worlddreamerv5}
    & WAM & 57.5 & 82.0 & 56.0 & 31.3 \\

    Xiaomi-Robotics-1~\citep{xiaomi2026robotics1}
    & VLA & 57.4 & 80.2 & 57.1 & 32.1 \\

    Meituan-Robotics-0~\citep{meituan2026robotics0}
    & VLA & 55.0 & 77.3 & 57.8 & 27.1 \\

    ABot-M0.5~\citep{chen2026abotm05}
    & WAM & 40.3 & 75.6 & 37.7 & 3.3 \\

    $\pi_{0.5}$~\citep{physicalintelligence2025pi05}
    & VLA & 16.9 & 39.6 & 7.1 & 1.2 \\

    \rowcolor[gray]{0.94}
    GR00T~2$^{\dagger}$~\citep{nvidia2026gr00t2}
    & AGENT+WAM
    & 81.8
    & 88.2
    & 86.9
    & 69.5 \\

    \midrule
    \textbf{Ours}
    & WAM & 60.2 & \textbf{88.4} & 56.1 & 32.4 \\

    \textbf{Ours}
    & AGENT+WAM & 65.4 & \textbf{90.6} & 58.9 & \textbf{43.6} \\

    \bottomrule
  \end{tabularx}
\end{table}

Stage review is triggered every 80 simulator steps during WAM execution. The agent verifies
execution every 256 simulator steps during normal operation, every 128 steps within a critical
window, and immediately after an explicit high-risk event. For online recovery, the agent takes
control of the current live simulator state without replaying the preceding trajectory. It can
interact for at most 20 reasoning--tool rounds. When calling the tools, perception-only operations
do not advance the simulator, whereas executed motion operations consume control steps.
Under our co-design protocol, WAM serves as the
default executor in both settings. The agent verifies execution
and intervenes for recovery. We do not synthesize task-level
coding-control to serve as the default executor.

\begin{table*}[t]
  \centering
  \caption{
    \textbf{Evaluation on RoboDojo-Sim.} Each entry reports Score / Success Rate (\%) from the
    \href{https://robodojo-benchmark.com/leaderboard}{public leaderboard} as of November 18, 2026. The reported Average value is the equal-weight mean of the scores for Generalization, Precision, Long-Horizon, Memory, and Open. The Cosmos3-Nano$^{\dagger}$ result refers to the independently reproduced and fine-tuned evaluation of Cosmos3-Nano~\citep{nvidia2026cosmos3} reported by the FluxVLA project~\citep{li2026fluxvlaengineonestopvla}. Best results are shown in bold.
  }
  \label{tab:robodojo_policy_agent_evaluation}

  \scriptsize
  \setlength{\tabcolsep}{2.4pt}
  \renewcommand{\arraystretch}{1.08}

  \begin{tabularx}{\textwidth}{
      @{}
      >{\raggedright\arraybackslash}p{0.20\textwidth}
      >{\centering\arraybackslash}p{0.11\textwidth}
      *{6}{>{\centering\arraybackslash}X}
      @{}
    }
    \toprule
    \textbf{Method}
    & \textbf{Type}
    & \textbf{Average}
    & \textbf{Generalization}
    & \textbf{Precision}
    & \textbf{Long-Horizon}
    & \textbf{Memory}
    & \textbf{Open} \\
    \midrule

    \multicolumn{8}{@{}l}{
      \textbf{Model Track}
      \hfill
    } \\
    \addlinespace[1pt]

    InternW0-$\Delta$~\citep{miao2026internw0delta}
    & WAM
    & \textbf{30.77} / 23.91
    & 30.09 / 22.78
    & 31.98 / 23.25
    & \textbf{46.29} / 29.33
    & 34.67 / 34.00
    & 10.84 / 10.17 \\

    Liber-0 Preview~\citep{liberai2026liber0}
    & VLM + WAM
    & 30.74 / \textbf{25.52}
    & 24.99 / 18.61
    & \textbf{38.28} / \textbf{33.17}
    & 45.98 / \textbf{33.17}
    & 37.77 / 37.33
    & 6.68 / 5.33 \\

    Meituan-Robotics-0~\citep{meituan2026robotics0}
    & VLA
    & 21.68 / 15.73
    & 16.31 / 10.50
    & 21.24 / 11.17
    & 27.07 / 13.92
    & \textbf{39.52} / \textbf{39.22}
    & 4.28 / 3.83 \\

    ME-Brain-1.0~\citep{he2026me0brain01000}
    & VLA + WAM
    & 21.67 / 15.99
    & 21.81 / 15.34
    & 24.23 / 15.92
    & 32.96 / 20.58
    & 23.53 / 22.78
    & 5.83 / 5.33 \\

    GalaxeaVLA (G0.5)~\citep{liu2026g05}
    & VLA
    & 20.23 / 14.88
    & 18.46 / 12.83
    & 28.25 / 20.42
    & 44.12 / 32.25
    & 8.61 / 7.33
    & 1.73 / 1.58 \\

    Xiaomi-Robotics-1~\citep{xiaomi2026robotics1}
    & VLA
    & 20.07 / 13.93
    & 23.54 / 17.00
    & 26.69 / 18.83
    & 38.39 / 23.67
    & 7.81 / 6.56
    & 3.94 / 3.58 \\

    MachEmbodied-U0~\citep{wen2026machembodiedu0}
    & WAM
    & 17.78 / 11.51
    & 16.07 / 8.78
    & 24.20 / 16.67
    & 36.66 / 22.25
    & 8.16 / 6.67
    & 3.78 / 3.17 \\

    Hy-Embodied-0.5-VLA~\citep{zhang2026hy0embodied00050vla0}
    & VLA
    & 13.07 / 8.80
    & 11.78 / 8.39
    & 13.81 / 8.00
    & 25.74 / 14.92
    & 13.37 / 12.11
    & 0.65 / 0.58 \\

    Cosmos3-Nano~\citep{nvidia2026cosmos3}$^{\dagger}$
    & WAM
    & 3.03 / 0.93
    & 2.58 / 1.00
    & 2.29 / 0.00
    & 8.74 / 2.75
    & 1.23 / 0.67
    & 0.30 / 0.25 \\

    \midrule
    \textbf{Ours}
    & \mbox{VLM + WAM}
    & 23.04 / 15.98
    & 30.88 / 23.67
    & 21.58 / 11.04
    & 32.31 / 18.56
    & 19.90 / 17.00
    & 10.53 / 9.63 \\

    \textbf{Ours}
    & \mbox{AGENT + WAM}
    & 30.02 / 24.17
    & \textbf{39.01} / \textbf{33.20}
    & 27.07 / 19.20
    & 32.99 / 20.00
    & 23.58 / 21.78
    & \textbf{27.46} / \textbf{26.66} \\

    \bottomrule
  \end{tabularx}
\end{table*}

\paragraph{Results Analysis}
On RoboCasa365 (Table~\ref{tab:robocasa-policy-results}), Innata-0-WAM achieves 60.2\% overall
success (1,504/2,500).
Success reaches 88.4\% on Atomic-Seen, compared with 56.1\% on Composite-Seen and 32.4\% on
Composite-Unseen.
The gap between atomic and composite tasks highlights the difficulty of translating individual
manipulation skills into reliable multi-stage execution, particularly for unseen task compositions.

With Innata-0-Agent, overall success increases to 65.4\%, a gain of 5.2 percentage points.
The largest improvement occurs on Composite-Unseen, where success rises from 32.4\% to 43.6\%,
gaining 11.2 percentage points. Both configurations use the same WAM checkpoint and full-task
instructions.
The agent verifies physical outcomes and invokes tool-based recovery, using skills refined from
validated interaction experience.
These results illustrate how outcome verification and experience-informed recovery jointly improve
task completion, extending the model's effective operating scope through agent--model collaboration.
GR00T~2's AGENT+WAM directly uses ASPIRE-generated programs as the default executor on 27 of 50
tasks, while Code-as-Policy is additionally available for recovery~\citep{nvidia2026gr00t2}.
Instead, in our system, WAM remains the nominal action policy across tasks, whereas the agent
provides task structuring, causal verification, and tool-composed live recovery. We therefore mark
this result as a reference under a different execution setting.

On RoboDojo-Sim (Table~\ref{tab:robodojo_policy_agent_evaluation}), our VLM + WAM system achieves an
overall score of 23.04 and a success rate of 15.98\%. Its strongest results are obtained on tasks
that require generalization. In the Generalization dimension, it achieves 30.88 / 23.67\%, exceeding
InternW0-$\Delta$ by 0.79 score points and 0.89 percentage points, respectively. It also reaches
10.53 / 9.63\% on Open tasks. These results indicate that the WAM provides a strong and transferable
physical execution foundation, particularly when the task requires adapting previously learned
manipulation behaviors to unfamiliar objects, layouts, or instructions. With full system involve, Innata-0-Agent can further enlarge the capabilities of Innata-0-WAM. Adding Innata-0-Agent improves both score and success rate in all five dimensions. The average score rises from 23.04 to 30.02, and the average success rate increases from 15.98\% to 24.17\% (+8.19 percentage points).
In addition to the substantial gains on Open tasks, Innata-0-Agent improves success rates in Generalization and Precision to 33.20\% and 19.20\%, respectively.

\paragraph{Visual Analysis}
\begin{figure}[htpb]
  \centering
  \includegraphics[width=\linewidth]{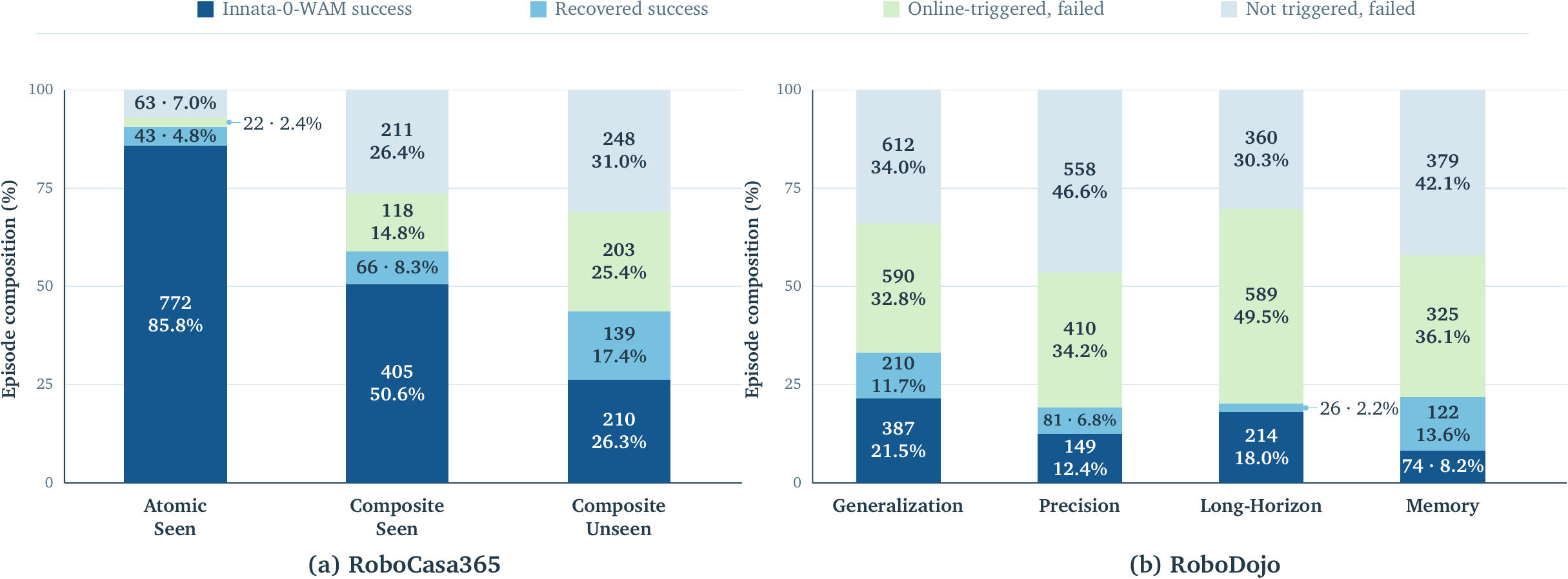}
  \caption{\textbf{Recovery capability analysis on RoboCasa365 and RoboDojo.} Each bar partitions episodes into four mutually exclusive final outcomes: Innata-0-WAM success, completed by the Innata-0-WAM without recovery; Recovered success, completed after a recovery intervention; Online-triggered, failed, where the online verification invoked recovery but the episode remained unsuccessful; and Not triggered, failed, where no recovery was triggered.}
  \label{fig:recovery_analysis}
\end{figure}

We evaluate the recovery capability of our system by analyzing how agent intervention changes
episode outcomes across RoboCasa365 and RoboDojo. As shown in Figure~\ref{fig:recovery_analysis}, on
RoboCasa365, recovery involvement increases with task difficulty. Atomic-Seen relies predominantly
on Innata-0-WAM execution, with 85.8\% Innata-0-WAM success and the smallest recovery involved. This
involvement increases for Composite-Seen and is highest for Composite-Unseen, while Innata-0-WAM
success decreases from 85.8\% to 50.6\% and 26.3\%, respectively. These results suggest that the
system adaptively balances the faster but less capable base model against the slower, more capable
agent: straightforward episodes are completed by Innata-0-WAM, whereas the agent participates more
frequently as task complexity increases.

On RoboDojo, recovery contributes differently across task dimensions. Generalization accounts for the largest number of recovered episodes, while Memory shows the largest relative gain over Innata-0-WAM execution. Memory has the lowest Innata-0-WAM success rate (8.2\%), but recovery adds 122 successful episodes, raising the final success rate to 21.8\%. One advantage of the agent is that it can check the current scene against a task requirement established earlier in the episode, such as whether the object being picked up matches the previously shown target. This check can surface a wrong-object grasp even when the grasp itself appears successful.

\begin{figure}[!htpb]
  \centering
  \includegraphics[width=\linewidth]{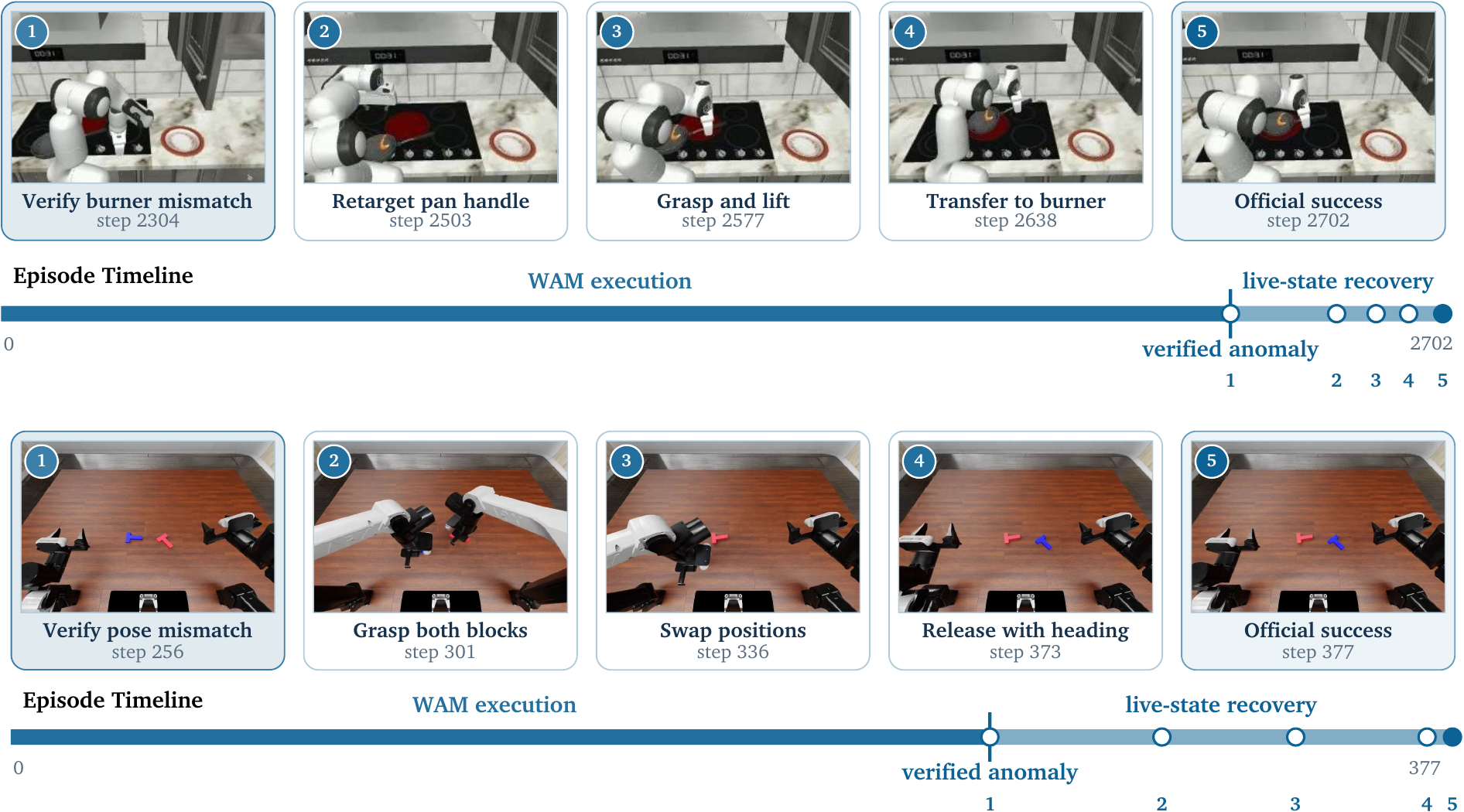}
  \caption{\textbf{Successful online recovery cases in RoboCasa365 and RoboDojo.}
    After verifying task-state mismatches during WAM execution, the agent intervenes directly in the live environment. In RoboCasa365, it regrasps a misplaced pan and transfers it onto the active burner (top). In the RoboDojo memory task, it identifies an incorrect arrangement of two T-shaped blocks, grasps both objects, swaps their positions, and releases them with the required headings (bottom). Both recoveries ultimately satisfy the corresponding official success conditions.}
  \label{fig:case_study}
\end{figure}

The two cases shown in Figure~\ref{fig:case_study} demonstrate online recovery across different
benchmarks, task horizons, and failure structures. In the RoboCasa365 searing task, the agent
verifies an incorrect object--fixture relation: the pan is not positioned on the active burner. It
retargets the pan handle, establishes a new grasp, lifts the pan, and transfers it to the correct
heat source. In the RoboDojo memory task, the agent verifies that the two T-shaped blocks do not
match the required pose arrangement. It then grasps both blocks, exchanges their positions, and
releases them with the appropriate headings.

The resulting interventions are failure-specific: one corrects an object--fixture relation through
regrasping and transport, whereas the other performs coordinated bimanual rearrangement to restore a
remembered spatial configuration. These examples show how the recovery module complements WAM by
converting verified visual discrepancies into targeted, state-dependent action compositions that
bring otherwise incomplete executions to the official success states.

\subsection{Policy Evaluation with Tactile Feedback}
\label{sec:tactile-evaluation}

In this evaluation, we focus on how tactile feedback helps the WAM handle contact interactions that
visual observations alone may not resolve, from atomic manipulation skills to multi-stage execution
within the embodied system.
To test the generality of our unified tactile representation, we choose two benchmarks with
different types of tactile sensors and manipulators.
UniVTAC~\citep{chen2026univtac} provides eight gripper-based tasks for evaluating contact control
and tactile perception.
Bench2Dex~\citep{yang2026bench2dex} combines tactile feedback with bimanual dexterous manipulation
and long-horizon tasks, allowing us to assess contact handling and finger coordination within the
system's multi-stage execution.

\paragraph{SFT setting}
For both benchmarks, we initialize task-specific policies from the final WAM checkpoint after joint
action and tactile pretraining.
We fine-tune for 20k steps on UniVTAC and 10k steps on Bench2Dex, with a global batch size of 128.
Each policy conditions on H9 synchronized RGB, robot-state, and tactile observations, comprising the
current observation and eight past observations.
The model jointly predicts 16 future video frames together with the corresponding actions and
tactile outputs.
The objective combines video and action flow matching with tactile presence and proxy supervision as
defined in \eqref{eq:tactile-prediction-loss}.
Learning rates are $10^{-5}$ for video generation and $10^{-4}$ for the action and tactile modules.

For UniVTAC, we fine-tune a separate policy for each of the eight tasks using 50 officially released
demonstrations per task (400 in total), conditioning on the task command without a plan.
For Bench2Dex, we use 50 expert demonstrations per selected task--embodiment setting and augment the
task command with \texttt{current\_plan} and \texttt{history\_plan}, derived from segmented expert
demonstrations.

\paragraph{Evaluation setting}

For both benchmarks, we execute all predicted actions before requesting the next chunk.

On UniVTAC, the WAM runs as a standalone closed-loop policy with tactile feedback over 100 official
test rollouts per task, without seed filtering or replacement.
Simulation runs at 120\,Hz, holding each action for two physics steps (60\,Hz control).
Inference uses non-EMA checkpoints with 10-step Euler sampling, with flow-time shift and guidance
scale both set to 1.0.
The causal H9 history is initialized by repeating the first observation; tactile inputs comprise
proxies and presence masks.

\Needspace{3\baselineskip}
For Bench2Dex, we exclude tasks with recorded trajectories that contain no tactile feedback and
select the ten most challenging tasks among those retained.
Each task is evaluated over 200 rollouts, with 50 for each of the four official settings.
We evaluate the WAM within the embodied system.
The agent invokes VLM-based progress checks using recent visual evidence, tactile-event summaries,
and task context.
Once completion is confirmed, it updates the current plan and completed-subtask history supplied to
the WAM.
We report composite Score, $0.45\,\mathrm{SR}+0.45\,\mathrm{LSCR}+10\,\mathrm{Eff.}$, combining
stable success rate, latched stage completion rate, and success-conditioned efficiency.
For the ten-task aggregate, SR and LSCR average all selected tasks, while efficiency averages only
tasks with successful rollouts.

\begin{table}[H]
  \centering
  \small
  \caption{\textbf{UniVTAC task success rates (\%).} Avg. is the unweighted mean across all eight tasks. ACT$^{\mathit{Tac}}$ uses the UniVTAC Encoder; GW-Policy and Xiaomi-R0 abbreviate GigaWorld-Policy and Xiaomi-Robotics-0. Bold and underline denote the best and second-best displayed values.}
  \label{tab:univtac-policy-results}
  \begingroup
    \fontsize{8.5}{9.5}\selectfont
    \setlength{\tabcolsep}{3pt}
    \renewcommand{\arraystretch}{1.0}
    \begin{tabular*}{\linewidth}{@{\extracolsep{\fill}}lrrrrrrrrr@{}}
      \toprule
      Method & \shortstack{Insert\\HDMI} & \shortstack{Insert\\Hole} & \shortstack{Insert\\Tube} & \shortstack{Grasp\\Classify} & \shortstack{Lift\\Can} & \shortstack{Lift\\Bottle} & \shortstack{Pull-out\\Key} & \shortstack{Put Bottle\\in Shelf} & Avg. \\
      \midrule
      GW-Policy~\citep{ye2026gigaworldpolicy} & 0.0 & 12.0 & 9.0 & 20.0 & 0.0 & 38.0 & 32.0 & 21.0 & 16.5 \\
      LingBot-VA~\citep{li2026lingbotva} & 38.0 & 42.0 & 96.0 & 17.0 & 0.0 & 0.0 & 58.0 & 0.0 & 31.4 \\
      $\pi_{0.5}$~\citep{physicalintelligence2025pi05} & 8.0 & 25.0 & 74.0 & 49.0 & 6.0 & \textbf{100.0} & 35.0 & 34.0 & 41.4 \\
      ACT$^{\mathit{Tac}}$~\citep{zhao2023act,chen2026univtac} & 28.0 & 24.0 & 56.0 & \underline{99.0} & 29.0 & 71.0 & 46.0 & 31.0 & 48.0 \\
      Fast-WAM~\citep{yuan2026fastwam} & 19.0 & 66.0 & \textbf{98.0} & 72.0 & 0.0 & 21.0 & 73.0 & 35.0 & 48.0 \\
      Xiaomi-R0~\citep{cai2026xiaomirobotics0} & \textbf{69.0} & 96.0 & \textbf{98.0} & 45.0 & 13.0 & 21.0 & 80.0 & 12.0 & 54.3 \\
      StarVLA-$\alpha$~\citep{ye2026starvlaalpha} & 24.0 & 52.0 & 69.0 & 68.0 & 65.0 & 32.0 & 51.0 & 88.0 & 56.1 \\
      FTP-1~\citep{yuan2026ftp} & 4.0 & 64.0 & 79.0 & \underline{99.0} & 65.0 & 97.0 & 48.0 & 47.0 & 62.9 \\
      InternVLA-A1~\citep{cai2026internvlaa1} & 12.0 & 93.0 & 95.0 & 86.0 & 90.0 & 58.0 & 66.0 & 37.0 & 67.1 \\
      T-Rex~\citep{niu2026trex,liu2026physword} & 8.0 & \textbf{100.0} & 95.0 & 95.0 & 75.0 & \underline{99.0} & \textbf{99.0} & 79.0 & 81.3 \\
      TacDyn-WAM~\citep{wang2026tacdynwam} & 12.0 & 91.0 & \underline{97.0} & \underline{99.0} & 87.0 & 97.0 & \underline{97.0} & 72.0 & 81.5 \\
      DECO.p~\citep{li2026deco} & 23.0 & 59.0 & 88.0 & \underline{99.0} & \textbf{99.0} & \textbf{100.0} & 96.0 & \underline{93.0} & 82.1 \\
      VTAM~\citep{yuan2026vtam} & 45.0 & 67.0 & \underline{97.0} & \textbf{100.0} & 75.0 & 91.0 & 92.0 & \textbf{100.0} & 83.4 \\
      N$_0$-TWAM~\citep{neote2026n0twam} & \underline{68.0} & \underline{99.0} & \textbf{98.0} & 94.0 & \underline{93.0} & 58.0 & 79.0 & 87.0 & \underline{84.5} \\
      \midrule
      \textbf{Ours} & 50.0 & \underline{99.0} & \textbf{98.0} & 95.0 & 82.0 & \underline{99.0} & 96.0 & 70.0 & \textbf{86.1} \\
      \bottomrule
    \end{tabular*}
  \endgroup
\end{table}

\begin{table}[H]
  \centering
  \caption{\textbf{Policy evaluation on ten Bench2Dex tasks using composite Score (higher is better).} T denotes the official task ID; Overall recomputes Score across these ten tasks. Bold and underline denote the best and second-best scores; \textemdash{} indicates undefined per-task efficiency. Tac denotes tactile input. Baseline results reflect the leaderboard as of November~18, 2026.}
  \label{tab:bench2dex-long-horizon-results}

  \begingroup
    \fontsize{8.5}{10.5}\selectfont
    \setlength{\tabcolsep}{2.0pt}
    \renewcommand{\arraystretch}{1.12}
    \begin{tabular*}{\linewidth}{@{\extracolsep{\fill}}l*{11}{r}@{}}
      \toprule
      \multirow{2}{*}{Method} & \multicolumn{10}{c}{Task Score $\uparrow$} & \multirow{2}{*}{Overall} \\
      \cmidrule(lr){2-11}
      & T21 & T24 & T26 & T32 & T34 & T43 & T60 & T64 & T73 & T79 & \\
      \midrule
      ACT~\citep{zhao2023act} & 28.33 & 20.46 & 29.51 & 17.16 & 30.06 & 12.09 & 21.35 & 9.31 & \textemdash & 8.67 & 19.18 \\
      DP~\citep{chi2023diffusionpolicy} & 10.87 & \textemdash & \textemdash & 12.81 & \textemdash & 10.29 & 10.33 & 8.98 & \textemdash & 9.42 & 11.63 \\
      $\pi_{0.5}$~\citep{physicalintelligence2025pi05} & 30.86 & \textemdash & 32.13 & 40.06 & 31.02 & 25.00 & 22.24 & 32.49 & \textemdash & 19.40 & 25.67 \\
      GR00T N1.5~\citep{nvidia2025grootn1} & 23.35 & \textemdash & 27.79 & 33.22 & 33.55 & 17.21 & 13.90 & 19.79 & 25.46 & 31.91 & 23.58 \\
      GR00T N1.5$^{\mathit{Tac}}$~\citep{nvidia2025grootn1} & 22.15 & \textemdash & 30.30 & 34.54 & 31.85 & 19.17 & 14.74 & 21.14 & 23.43 & 28.93 & 23.60 \\
      GR00T N1.7~\citep{nvidia2025grootn1} & 37.59 & \textemdash & 26.47 & 36.54 & 32.21 & 19.15 & 19.99 & 22.47 & 15.08 & 27.77 & 24.71 \\
      GR00T N1.7$^{\mathit{Tac}}$~\citep{nvidia2025grootn1} & 41.37 & 24.67 & 30.20 & 30.70 & 42.29 & 20.10 & 29.72 & 22.72 & 21.88 & 33.90 & 29.76 \\
      Mora V0~\citep{morphi2026morabench2dex} & 52.55 & 28.28 & 49.24 & 49.31 & 39.79 & 24.66 & 37.85 & 33.13 & 25.31 & 24.36 & 36.45 \\
      MXI~\citep{ai10112026mxibench2dex} & 56.25 & 14.87 & 36.36 & 36.91 & 44.65 & 22.41 & 37.21 & 41.18 & \textemdash & 21.57 & 32.47 \\
      Phoenix-MWA~\citep{openphoenix2026bench2dex} & 65.83 & \underline{44.26} & 38.83 & 48.64 & \underline{76.35} & \textbf{45.03} & 71.20 & 37.43 & \textemdash & 13.09 & 46.18 \\
      PyRUA-Dex~\citep{pku2026pyruadex} & 37.35 & 31.14 & \textbf{65.35} & \underline{51.00} & \textemdash & 30.56 & 53.73 & \underline{55.51} & \textemdash & \textbf{86.45} & 46.35 \\
      GW-Policy-0.6~\citep{ye2026gigaworldpolicy} & \textbf{84.12} & \textbf{63.24} & 57.92 & 46.44 & \textbf{77.59} & \underline{42.15} & \underline{71.65} & 43.16 & \textbf{31.51} & 17.40 & \underline{53.52} \\
      \midrule
      \textbf{Ours} & \underline{77.86} & 20.89 & \underline{61.51} & \textbf{51.76} & 76.12 & 39.56 & \textbf{72.79} & \textbf{58.45} & \underline{26.10} & \underline{53.17} & \textbf{53.82} \\
      \bottomrule
    \end{tabular*}
  \endgroup
\end{table}

\FloatBarrier

\paragraph{Results Analysis}
On UniVTAC (Table~\ref{tab:univtac-policy-results}), the WAM achieves an average success rate of
86.1\%.
It reaches 50.0\% on Insert HDMI and 99.0\% on Lift Bottle, demonstrating its ability to handle both
precise insertion and stable grasp maintenance.
These results suggest that the learned tactile representation retains contact information useful for
different forms of manipulation.

On Bench2Dex (Table~\ref{tab:bench2dex-long-horizon-results}), the same pretrained WAM achieves an
overall score of 53.82 within the agent system, outperforming all the compared methods on the
selected ten-task subset.
It ranks first on Baking Tray Prep (T32), Breadbasket Fast-Food Loading (T60), and Sports Ball Cup
Sort (T64), with scores of 51.76, 72.79, and 58.45, respectively.
Performance across these two benchmarks suggests that the shared regional representation provides a
useful foundation for learning contact-rich control across different sensors and embodiments.
By bringing heterogeneous tactile data into a common format, tactile mid-training allows
task-specific fine-tuning to build on interaction experience collected from multiple sources.

Long-horizon tasks further illustrate the benefits of our model-agent co-design.
Fridge Wine Interhand Pour (T34) involves complex bimanual handovers that are difficult for existing
coding agents to execute, while Bimanual Piano Playing (T79) places demanding sequencing and timing
requirements on standalone WAM/VLA policies.
The co-designed system achieves competitive results on both.
The agent maintains the task structure and updates the current subgoal, while the WAM generates the
dexterous motions needed to carry it out.
These results support our co-design of task-level orchestration and learned manipulation, combining
their strengths to address tasks that challenge either component alone.
\FloatBarrier
\section{Limitations and Future Directions}
While Innata-0 has demonstrated promising progress in embodied understanding, world prediction, and robotic manipulation, challenges remain toward native embodied intelligence. Generalization across environments, tasks, and robot embodiments warrants further exploration. More extensive real-world validation is needed to assess system generalization across robot embodiments, diverse environments, and increasingly complex tasks, as well as robustness to dynamic changes and unexpected failures during physical interaction.

Looking ahead, we see opportunities to advance both model and agent capabilities. On the model side, we aim to improve generalization across tasks and robot embodiments, strengthen physical understanding and prediction, and support more complex dexterous manipulation. This requires expanding the diversity, quality, and coverage of embodied data, including egocentric videos, real-robot trajectories, tactile interactions, and dexterous manipulation data. We also aim to leverage real-world execution and recovery experiences for iterative model improvement and continual agent evolution toward more generalizable and reliable native embodied world systems. On the agent side, we aim to advance multimodal reasoning with richer visual, spatial, and tactile information. We also plan to explore in-context learning and experience retrieval to improve adaptation to unfamiliar situations. Reducing token consumption and optimizing inference efficiency will also be important for scalable, real-world deployment.
\section{Contributions and Acknowledgments}
\label{sec:author}
All contributors are listed in alphabetical order by their last names.

\textbf{Core Contributors:}
Yu Bai, Ziyi Bai, Yuxuan Cai, Chenrui He,
Wenhao Li, Zhe Lin, Heng Wang, Hai Yuan,
Zhiwei Yu, Yuanhang Zhang

\textbf{Contributors:}
Luyou Chen, Tiezhu Gao, Zheng Geng, Jinbang Guo,
Dongxu Han, Diqi He, B\"orje F. Karlsson,
Jintao Li, Zhiyu Li, Xiongkun Linghu,
Guang Liu, You Liu, Zhen Luo,
Mengsi Lyu, Yufeng Lyu, Zheng Lyu,
Quanyue Ma, Chunlei Men, Ziqiang Ni,
Tengfei Pan, Yijun Shen, Liang Sun,
Shaoxuan Xie, Xiang Yang, Guocai Yao,
Hongjun Zhang, Yingli Zhao

\textbf{Project Lead:}
Yonghua Lin,
Zhongyuan Wang\textsuperscript{\Letter},
Huan Yang\textsuperscript{\Letter}

\fancypagestyle{correspondingfooter}[fancy]{
  \fancyfoot[L]{\footnotesize
    \Letter\ Corresponding Authors: Zhongyuan Wang,
    Huan Yang (\texttt{hyang@fastmail.com})}
}
\thispagestyle{correspondingfooter}

\textbf{Acknowledgments:}
We sincerely thank Lightwheel for its contributions to egocentric data and PaXini for its
contributions to tactile data.

\clearpage
\begingroup
  \interlinepenalty=10000
  \reportreferences{innata-0}
\endgroup

\reportappendix
\numberwithin{equation}{section}
\numberwithin{table}{section}
\numberwithin{figure}{section}

\clearpage
\FloatBarrier
\section{Unified Annotation Records}
\label{app:video-caption-examples}

These examples illustrate the shared annotation layer in Section~\ref{sec:video-task-semantics}. One
reusable record combines scene descriptions, task goals, temporal intervals, execution outcomes, and
quality attributes; captions are one downstream use. It retains links to the original task text,
native trajectory channels, and sensor timestamps. Fields are populated only with source support: a
video may support semantic event boundaries without providing metric joint motion or tactile
measurements.

\subsection{Shared Context and Interval-Level Descriptions}

\begin{figure}[p]
  \centering
  \includegraphics[width=\linewidth,height=.84\textheight,keepaspectratio]{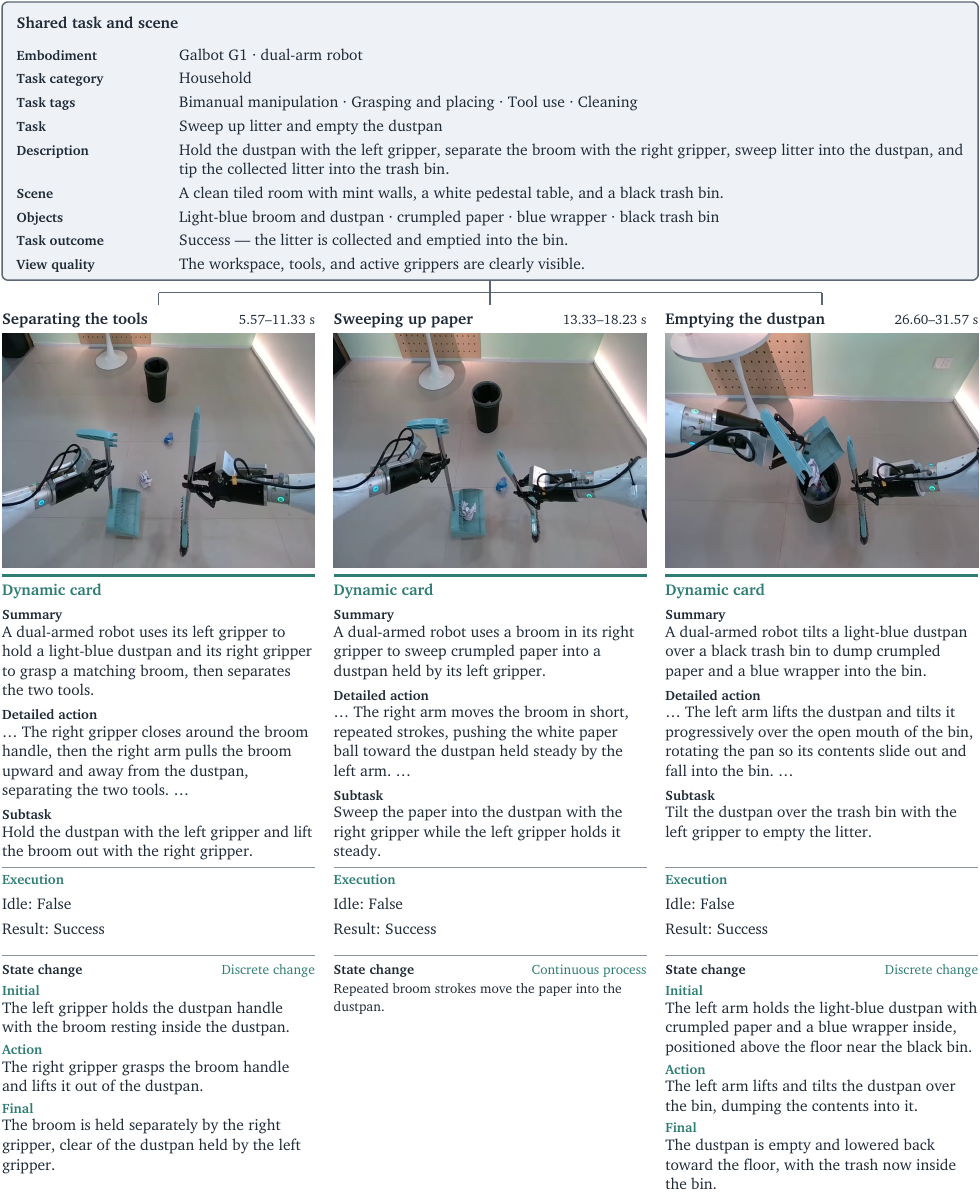}
  \caption{\textbf{Task-level and subtask-level annotations.} The shared card combines the Galbot G1 embodiment, task category and tags, full task description, scene context, outcome, and visibility. Three timed clips illustrate subtasks, execution results, and state changes. Display labels draw on the supplied frames and retained descriptions.}
  \label{fig:caption-cards}
\end{figure}

\noindent\textbf{One scene, three actions.}
The three clips in \figref{fig:caption-cards} share scene context and task-level annotations.
Dynamic cards describe separating the broom and dustpan, sweeping paper into the dustpan, and
emptying it into a bin. The sweeping description follows repeated broom strokes; the emptying
description captures a state change: the left arm tips the dustpan containing crumpled paper and a
blue wrapper over the bin, then lowers it empty. Consistent references to the left and right
grippers, tools, paper, and bin connect descriptions, task intervals, and visible relations, such as
paper resting in the dustpan before being deposited in the bin. When available, tracks, masks, or
depth from spatial modules can be linked to the same entities alongside their language descriptions.

\subsection{Semantic Subtasks, Execution Outcomes, and Evidence}
\label{app:annotation-subtasks}

\noindent\textbf{A goal defines a subtask boundary.}
For robot trajectories, the shared semantic pass produces a short \texttt{task\_title}, an ordered
\texttt{task\_description}, and timestamped subtask intervals. Each subtask groups movements toward
one manipulation goal: reaching, grasping, and lifting a piece of bread form one picking subtask;
placing and releasing it on a plate form the next. Repeated attempts at the same goal remain in one
subtask when followed by success, so two missed grasps and one successful grasp do not become three
unrelated tasks. Visible atomic actions or execution stages may be recorded within the parent
interval without becoming separate training episodes.

\noindent\textbf{Illustrative structured example.}
Table~\ref{tab:annotation-subtasks} illustrates the annotation protocol; it is not an additional
measured trajectory. The input goal is ``Place the bread on the plate,'' the duration is 11.4~s, and
the sequence is sampled at 1~FPS with explicit timestamps. The detailed description is: ``Pick up
the bread with the left gripper after two missed attempts, place it on the plate, release it, and
return the arm to its initial position ready for the next task.'' This describes the observed
procedure, not a generic goal combined with object names.

\begin{table}[htbp]
  \centering\small
  \setlength{\tabcolsep}{4pt}
  \renewcommand{\arraystretch}{1.12}
  \caption{\textbf{Illustrative subtask intervals in one shared record.} Times are episode-local seconds. Each interval has one behavior label and a separate idle flag. The terminal boundary follows the supplied duration, not the number of sampled images.}
  \label{tab:annotation-subtasks}
  \begin{tabularx}{\linewidth}{@{}l>{\raggedright\arraybackslash}Xll@{}}
    \toprule
    \textbf{Interval (s)} & \textbf{Subtask description} & \textbf{Behavior} & \textbf{Idle} \\
    \midrule
    0--5 & Pick up the bread with the left gripper after two missed attempts. & retry then success & false \\
    5--10 & Place the bread on the plate with the left gripper and release it. & success & false \\
    10--11.4 & End: Return the arm to its initial position and prepare for the next task. & end & false \\
    \bottomrule
  \end{tabularx}
\end{table}

\noindent\textbf{Separate intent, execution, and visibility.}
The record preserves the original goal alongside the corrected title and detailed description.
\texttt{task\_alignment} checks whether the text identifies the visible objects, operations, and
destinations; \texttt{outcome} assesses whether the applicable goal was achieved. A correct
instruction may accompany failure, while an incorrect object description may accompany successful
execution of the visibly supported task. Native success flags remain separate from video-based
assessments. In the example, alignment is \texttt{consistent} and the overall outcome is
\texttt{success}, despite the initial missed grasps.

Each robot interval receives one behavior label from \{\texttt{normal}, \texttt{success},
\texttt{failure}, \texttt{retry then success}, \texttt{end}\}, together with \texttt{is\_idle}. A
purposeful reset is an ending phase but is not idle. An ending segment is optional: a recording may
stop immediately after successful release, without a reset or pause. Overall outcome also permits
\texttt{uncertain} when missing context or visibility prevents assessment, distinct from a visibly
unfinished task or an annotation-processing failure.

\noindent\textbf{Timestamped input and output validation.}
Inputs include source task text, actual duration, sampling rate, camera identifiers, and
sampled-image timestamps. Sampling rate is distinct from the original video frame rate. With sparse
sampling, boundaries reflect the evidence's temporal resolution, not frame-exact contact events.
Multiple views show the same episode, not repeated attempts. Output validation checks the JSON
schema, label vocabulary, interval order, and bounds within the supplied duration. Recoverable
formatting errors can be corrected and recorded; unparseable responses are marked as annotation
failures, not task failures. Retained source-time offsets locate episode-local intervals within
longer shared video files.

\noindent\textbf{Quality and numerical evidence accompany the semantics.}
R2 Stage~1 retains a perceptual video-quality score and its rationale alongside the semantic
response. Scores range from 1 (task-critical content unusable) to 5 (clear, stable, and adequately
visible) and do not measure task success. Separate low-level measurements characterize blur,
exposure, frame changes, and decoding problems. For available native trajectories, numerical checks
collect motion, derivative, trend-residual, and stationarity evidence, with applicability and
calibration status. Sparse VLM observations cannot rule out brief freezes or high-frequency jitter,
and uncalibrated numerical rules cannot certify a trajectory as normal. These records support
filtering and review without requiring regeneration of semantic descriptions. Stage~1 extracts
evidence; downstream preparation decides whether to retain, split, exclude, or smooth a trajectory.

\subsection{Reusing One Record Across Preparation Branches}

\begin{figure}[!htbp]
  \centering
  \includegraphics[width=\linewidth]{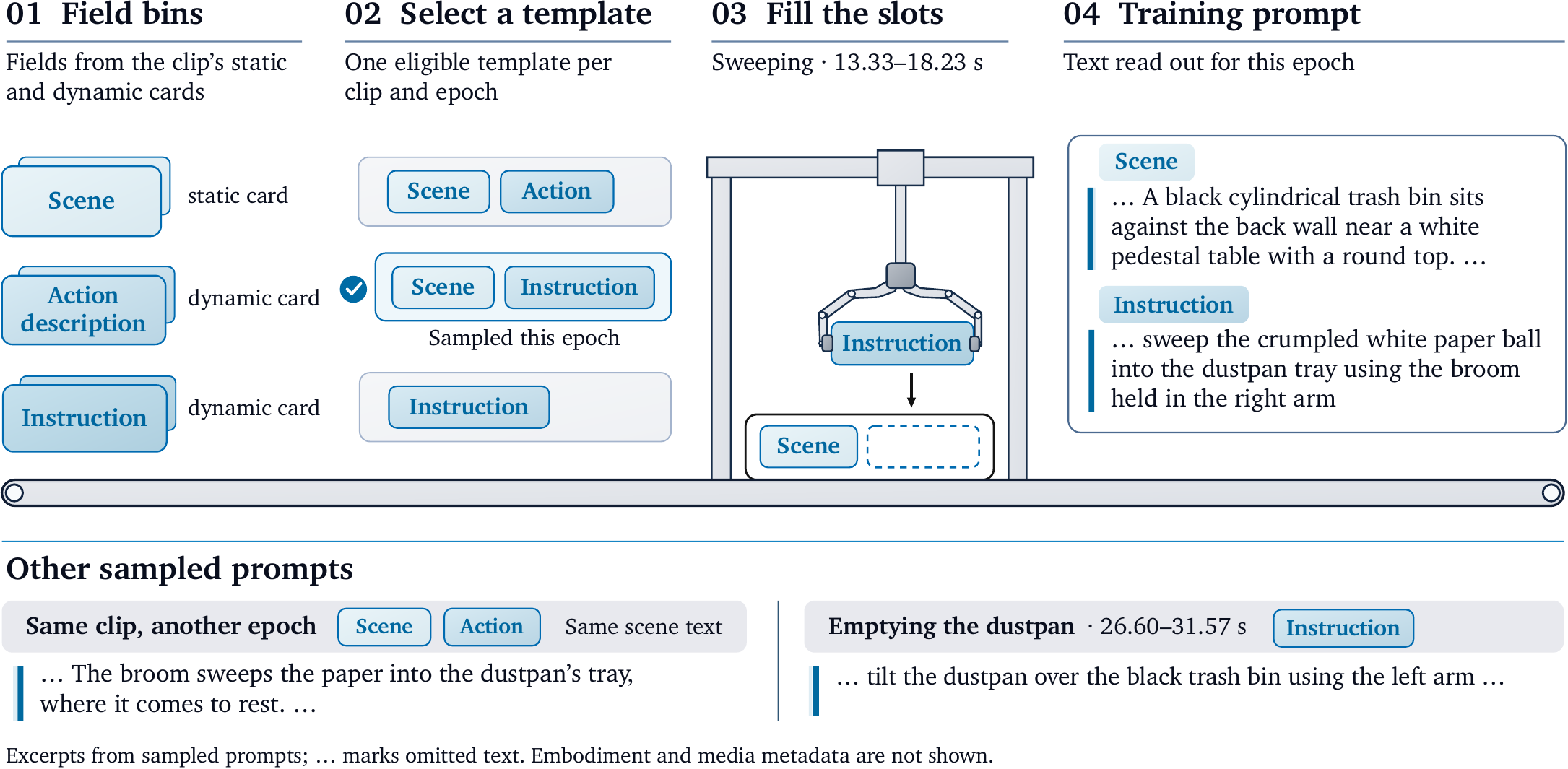}
  \caption{\textbf{Prompt assembly from reusable annotations.} Alternative templates combine scene and dynamic cards. The lower panels link assembly to trajectory fields: detailed task text and subtask instructions provide conditioning, boundaries select intervals, and outcome, idle, and quality fields support downstream selection and audit. These panels specify how fields are used; they do not report additional predictions for the displayed clips.}
  \label{fig:caption-assembly}
\end{figure}

\noindent\textbf{One clip, different prompts.}
For the sweeping clip (13.33--18.23~s), two prompts from different epochs share the same scene
description, including the sentence “A black cylindrical trash bin sits against the back wall near a
white pedestal table with a round top.” One describes the action: “The broom sweeps the paper into
the dustpan's tray, where it comes to rest.” The other instructs the robot to hold the dustpan
upright with its left arm and sweep the paper into it with the broom in its right gripper. A prompt
for the later emptying clip (26.60--31.57~s) uses an instruction without scene text
(\figref{fig:caption-assembly}).

\figref{fig:caption-assembly} shows how video generation uses the shared record, rather than a separate annotation pipeline. Other branches select different fields from the same evidence:
\begin{itemize}
  \item \textbf{Embodied understanding:} construct questions about the active gripper, object relations, subtask order, or the result of a retry, using only the observations available to the question.
  \item \textbf{Visual prediction:} combine static context with an interval-specific dynamic description or instruction, as shown in the figure.
  \item \textbf{Action prediction:} pair a selected subtask instruction and task context with the corresponding state/action interval and camera streams. Semantic boundaries guide selection but do not establish numerical synchronization or create control targets.
  \item \textbf{Tactile prediction:} link available contact observations and validity masks to the same event intervals. Visible holding or touching alone does not provide a measured tactile target.
\end{itemize}
The record retains the full task description, ordered plan, and local intervals. If filtering splits
a trajectory, each retained segment keeps its source-time correspondence and parent-plan reference.
Removing an idle interval neither erases the original task context nor connects actions across the
resulting gap. Training prompts select the appropriate context and local instruction; outcome,
quality, idle, and processing-status fields support selection and audit rather than being appended
indiscriminately to every prompt.

\FloatBarrier
\section{Understanding Training Details}
\label{app:understanding_details}

\begin{table*}[t]
  \centering
  \small
  \setlength{\tabcolsep}{5pt}
  \caption{\textbf{Data sources represented in the understanding-tower training mixtures.} M denotes mid-training and P denotes post-training. Inclusion refers to selected subsets rather than use of an entire public release. Dataset release families are listed for provenance and can share upstream media or annotations; they are not necessarily disjoint. Internally assembled collections are separated from public releases.}
  \label{tab:understanding_data}
  \renewcommand{\arraystretch}{1.05}
  \resizebox{\linewidth}{!}{%
  \begin{tabular}{p{0.29\textwidth}p{0.55\textwidth}l}
    \toprule
    Dataset / release family & Supervision in the selected subsets & Phase \\
    \midrule
    \multicolumn{3}{l}{\textit{General multimodal understanding}} \\
    Cambrian~\citep{tong2024cambrian010} & General visual instruction following & M \\
    LLaVA-OneVision~\citep{li2024llava0onevision0} & Image/video understanding and instruction following & M \\
    \addlinespace
        \midrule
    \multicolumn{3}{l}{\textit{Spatial perception and reasoning}} \\
    SPAR~\citep{zhang2025flatland} & Spatial relations, geometry, multi-view and video reasoning & M, P \\
    OpenSpatial~\citep{liu2026openspatial0} & Video-based spatial understanding & M \\
    HoliSpatial-QA~\citep{gao2026holi0spatial0} & Spatial question answering & M \\
    VST~\citep{yang2025visual} & Multi-view and video spatial understanding & M, P \\
    3D-RFT~\citep{linghu20263d0rft0} & Spatial reasoning and supervised reasoning traces & M, P \\
    VSI-590K~\citep{yang2025cambrian0s0} & Video-based spatial and metric reasoning & M, P \\
    SenseNova-SI~\citep{cai2026scalingspatialintelligencemultimodal} & Spatial understanding & M, P \\
    VLM3R-VSTI~\citep{fan2026vlm3rvisionlanguagemodelsaugmented} & Multi-view spatial reasoning & M, P \\
    EmbSpatial-SFT~\citep{du2024embspatialbenchbenchmarkingspatialunderstanding} & Embodied spatial relations & P \\
    MindCube~\citep{yin2025spatial} & Multi-view spatial reasoning & P \\
    XVR~\citep{jeong2026learningmultiviewspatialreasoning} & Cross-view spatial relations & P \\
    Viewpoint-100K~\citep{zhan2025actialactivatespatialreasoning} & Viewpoint-dependent spatial reasoning & P \\
    \addlinespace
        \midrule
    \multicolumn{3}{l}{\textit{Grounding and affordance understanding}} \\
    RefSpatial~\citep{zhou2026roboreferspatialreferringreasoning} & Spatial referring and point localization & M, P \\
    RoboAfford~\citep{Tang2025RoboAffordAD} & Functional grounding and affordance localization & M, P \\
    RoboPoint~\citep{yuan2024robopointvisionlanguagemodelspatial} & Pointing, spatial grounding and placement & M, P \\
    HoliSpatial-3D-Grounding~\citep{gao2026holi0spatial0} & 3D grounding with geometric context & M \\
    VLM-R1-RefCOCO~\citep{shen2025vlmr1stablegeneralizabler1style} & Referring-expression grounding & M, P \\
    \addlinespace
        \midrule
    \multicolumn{3}{l}{\textit{Embodied task understanding}} \\
    EO~\citep{qu2026eo1openunifiedembodied} & Embodied perception and task understanding & M, P \\
    Robo2VLM~\citep{chen2025robo2vlmvisualquestionanswering} & Robot-centric visual question answering & M \\
    ShareRobot~\citep{ji2025robobrain} & Affordance, trajectory and planning supervision & M, P \\
    Embodied-R1.5~\citep{yuan2026embodied0r1050} & ERQA-related embodied question answering & P \\
    \addlinespace
        \midrule
    \multicolumn{3}{l}{\textit{Perceptual and grounded reasoning}} \\
    LongPerceptualThoughts~\citep{liao2025longperceptualthoughtsdistillingsystem2reasoning} & Visual perceptual reasoning & P \\
    TreeVGR~\citep{wang2026traceableevidenceenhancedvisual} & Grounded visual reasoning & P \\
    Spatial-SSRL~\citep{liu2025spatialssrlenhancingspatialunderstanding} & Spatial reasoning supervision & P \\

    \midrule
    \multicolumn{3}{l}{\textit{Our constructed data}} \\
    Innata-Understanding (ours) &
    Source-grounded QA for object grounding, spatial reasoning,
    and embodied task understanding & M, P \\
    \bottomrule
  \end{tabular}
}
\end{table*}

We adapt the understanding tower through a staged supervised training pipeline that combines broad
spatial and embodied learning with task-focused refinement. Starting from
Qwen3-VL-8B~\citep{bai2025qwen30vl}, the training progressively emphasizes spatial understanding
while retaining general-purpose visual supervision. Below, we describe the data composition and
training trajectory of the evaluated checkpoint, distinguishing the available source inventory, the
prepared training mixtures, and the supervised sample exposures actually consumed.

For mid-training, our source inventory contains approximately 63.6M candidate records, of which
60.8M are eligible after source-specific preprocessing and validation. The eligible inventory
comprises approximately 11.7M general-purpose records and 49.2M embodied-oriented records, including
internally assembled collections. These figures describe source-pool records rather than globally
deduplicated examples or actual training exposure. From this inventory, we prepare a 20M-example
mid-training mixture, with approximately 13.3\% general-purpose and 86.7\% embodied-oriented data by
source category. Detailed datasets are shown in Table~\ref{tab:understanding_data}.

Training proceeds from broad mid-training to task-focused post-training. Mid-training emphasizes
large spatial and embodied datasets while retaining general-purpose supervision. Post-training first
applies a task-balanced multi-source mixture and then refines the model with the curated
question–answer collection, compact task-specific training sets, and general-purpose replay. The
final refinement mixture contains 57,565 task-focused examples, 42,737 examples from compact
task-specific sets, and 20,000 general-purpose examples. The compact sets comprise EmbSpatial-SFT,
MindCube, and the ERQA-related training subsets of Embodied-R1.5.

We fine-tune both the language and visual components using AdamW
with $\beta_1=0.9$, $\beta_2=0.95$, $\epsilon=10^{-8}$, and weight
decay of $0.1$. Training uses bfloat16 computation, fully sharded
data parallelism, activation checkpointing, and gradient-norm
clipping at $1.0$. The peak learning rate for the language model
is $10^{-5}$ throughout training; the visual component uses
$2\times10^{-6}$ during mid-training and $10^{-6}$ during both
multi-source post-training and final refinement. These three
stages use cosine learning-rate schedules with linear warm-up. The initial and minimum learning
rates are set to $1\%$ of their respective peak values.

The final checkpoint descends from a mid-training model with approximately 7.99M supervised sample
exposures, followed by approximately 0.20M exposures in multi-source post-training and 0.12M in
final refinement.

\FloatBarrier
\section{Dexterous Hands}
\subsection{Egocentric Hand Reconstruction Model}
\label{app:hand-reconstruction}

This subsection describes the hand reconstruction model used to construct egocentric motion
supervision in Section~\ref{sec:dexterous-ego-data}. Its source training mixture contains $\sim 208$
hours of video and $73.8$k static images. We build on ACE-Ego-Hand~\cite{liu2026ace}, with pinhole
camera inputs, joint-distribution supervision, and sequence-level shape and placement refinement.
This training mixture is distinct from the action-learning corpus in
Table~\ref{tab:egocentric_data}.

\subsubsection{Architecture and Camera Representation}
\label{app:hand-architecture}

We follow the clean-latent encoder and bidirectional spatiotemporal decoder of
ACE-Ego-Hand~\cite{liu2026ace}, using \texttt{Wan2.2-Fun-5B-Control}
features~\cite{aigc_apps_VideoX_Fun_2026,wan2025wan}. To impose a consistent camera model across
sources, we rectify all inputs to a pinhole domain and update intrinsics after cropping and
resizing. We then generate the standard ray map $\mathbf r_K(u,v)=K^{-1}[u,v,1]^\top$ analytically
for ray encoding and geometric decoding, replacing the freely predicted ray field. We use supplied
calibration when available; otherwise, we additionally predict pinhole parameters from the input
features and generate rays through the same mapping. Table~\ref{tab:hand-training-config} summarizes
the model and training settings.

\begin{table}[htbp]
  \centering\small
  \setlength{\tabcolsep}{5pt}
  \caption{\textbf{Hand reconstruction configuration.} The configuration describes the model used in this work. Learning rates distinguish the fully trained decoder and heads, LoRA parameters, and the trainable patch embedding.}
  \label{tab:hand-training-config}
  \begin{tabularx}{\linewidth}{@{}l X@{}}
    \toprule
    \textbf{Component} & \textbf{Configuration} \\
    \midrule
    Visual initialization & \texttt{Wan2.2-Fun-5B-Control}, clean posterior-mean latents \\
    Feature extraction & Transformer block index 15, LoRA rank / scale $64/64$ \\
    Spatiotemporal decoder & $4$ layers, width 384, $8$ attention heads \\
    Video input & 81 RGB frames, corresponding to 21 temporal latent samples \\
    Image input & $5$ repeated RGB frames, corresponding to $2$ temporal latent samples \\
    Training duration / global batch & $80,000$ optimizer steps / $48$ input examples \\
    Optimizer & AdamW, weight decay $10^{-2}$ \\
    Learning rates & Decoder / heads $2\!\times\!10^{-4}$, LoRA $10^{-4}$, patch embedding $2\!\times\!10^{-5}$ \\
    Learning-rate schedule & $200$-step linear warmup, followed by cosine decay \\
    Heatmap target & Gaussian, $\sigma=0.6$ grid intervals, JS coefficient $0.1$ \\
    Heatmap / camera loss ramp & $500$ optimizer steps \\
    \bottomrule
  \end{tabularx}
\end{table}

\subsubsection{Training Data}
\label{app:hand-training-data}
\begin{figure}[htbp]
  \centering
  \includegraphics[width=\linewidth]{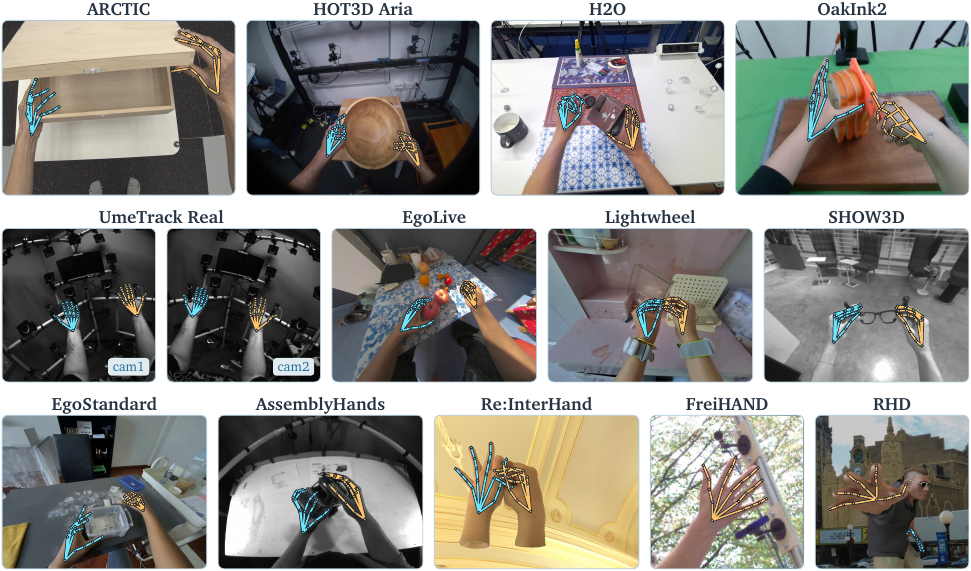}
  \caption{\textbf{Training-source diversity.} Our training set spans controlled capture, everyday manipulation, and synthetic imagery. Overlaid hand skeletons are drawn from each dataset's native annotations.}
  \label{fig:hand-training-source-images}
\end{figure}

Table~\ref{tab:hand-reconstruction-sources} reports training data and sampling probabilities. We
draw EgoLive, Lightwheel, and EgoStandard from larger pools by stratifying tasks and scenes, and
reserve unseen scenes and workers for validation. For UmeTrack, we use videos from the "real"
subset's \texttt{cam1} and \texttt{cam2}, in which the hand is visible. We deliberately exclude
EgoDex~\cite{hoque2026egodex} from the training mixture because the native ARKit annotations we
inspected have overly long skeletal bones and inconsistent 3D-to-image alignment. For action
learning, we instead re-estimate hand poses from EgoDex videos with our model, as described in
Section~\ref{sec:dexterous-ego-data}. Figure~\ref{fig:hand-training-source-images} illustrates the thirteen sources and the selected
UmeTrack views.

\begin{table}[htbp]
  \centering\small
  \setlength{\tabcolsep}{3pt}
  \renewcommand{\arraystretch}{1.12}
  \caption{\textbf{Training data for hand reconstruction.} FreiHAND and RHD contribute static images; all other rows count video frames. Duration excludes static images. Canvas dimensions are width $\times$ height, and sampling denotes source selection probability. Validation data are excluded.}
  \label{tab:hand-reconstruction-sources}
  \begin{tabularx}{\linewidth}{@{}>{\raggedright\arraybackslash}X r r r c r@{}}
    \toprule
    \textbf{Source} & \textbf{Records} & \makecell[r]{\textbf{\#Frames}} & \textbf{Hours} & \textbf{Canvas} & \textbf{Sampling} \\
    \midrule
    ARCTIC~\cite{fan2023arctic} & 267 & 184,373 & 1.71 & $672\times480$ & 8\% \\
    HOT3D Aria~\cite{banerjee2025hot3d} & 126 & 444,649 & 4.12 & $480\times480$ & 10\% \\
    H2O~\cite{kwon2021h2o} & 114 & 68,929 & 0.64 & $832\times480$ & 5\% \\
    \textsc{OakInk2}~\cite{zhan2024oakink2} & 550 & 935,260 & 8.66 & $832\times480$ & 8\% \\
    UmeTrack Real~\cite{han2022umetrack} & 1,992 & 876,018 & 8.11 & $480\times480$ & 5\% \\
    EgoLive~\cite{li2026egolive} & 6,823 & 5,390,474 & 50.28 & $480\times480$ & 11\% \\
    Lightwheel & 6,452 & 5,309,132 & 49.99 & $832\times480$ & 11\% \\
    SHOW3D~\cite{rim2026show3d} & 3,162 & 3,026,568 & 28.02 & $384\times480$ & 10\% \\
    EgoStandard~\cite{lightwheel2026egosuite} & 6,493 & 5,309,250 & 49.99 & $832\times480$ & 11\% \\
    AssemblyHands~\cite{ohkawa2023assemblyhands} & 800 & 697,054 & 6.46 & $640\times480$ & 6\% \\
    Re:InterHand~\cite{moon2023dataset} & 43 & 21,315 & 0.59 & $480\times480$ & 5\% \\
    FreiHAND~\cite{zimmermann2019freihand} & 32,560 & 32,560 & --- & $224\times224$ & 5\% \\
    Rendered Handpose Dataset (RHD)~\cite{zimmermann2017learning} & 41,258 & 41,258 & --- & $320\times320$ & 5\% \\
    \midrule
    \textbf{Total} & \textbf{100,640} & \makecell[r]{\textbf{22,263,022 frames}\\\textbf{+73,818 images}} & \textbf{208.57} & --- & \textbf{100\%} \\
    \bottomrule
  \end{tabularx}
\end{table}

\paragraph{Preprocessing}
We remap fisheye inputs from HOT3D Aria, Re:InterHand, and UmeTrack to pinhole canvases, and apply
fixed photometric processing to Aria recordings (de-vignetting and color correction). We standardize
joints, units, handedness, and MANO conventions, using validity and existence/visibility masks to
represent each source's annotation coverage.

\paragraph{Crop and scale augmentation}
To vary apparent hand size and position, we sample coverage-aware crops and apply one spatial
transform throughout each sequence before VAE encoding. We update intrinsics, two-dimensional
targets, visibility, and rays together, preserving the camera-frame 3D geometry and articulation. We
use the base canvas for validation.

\subsubsection{Supervision and Optimization}
\label{app:hand-supervision}
We supervise direct and MANO geometry, articulation, 2D joint localization and reprojection, hand
existence and visibility, and temporal smoothness. In particular, we fit MANO parameters for
UmeTrack Real, AssemblyHands, Lightwheel and EgoStandard with episode-consistent shapes. We mask
each objective by target availability, average valid residuals per source, and combine sources using
the probabilities in Table~\ref{tab:hand-reconstruction-sources}.
Table~\ref{tab:hand-loss-coefficients} lists the coefficients. We focus below on the additional
distribution and camera supervision.

\paragraph{Joint-distribution supervision}
A coordinate mean alone cannot distinguish a localized joint prediction from diffuse or separated
spatial support.
We therefore supervise the attention-based joint distribution $P$ with a Gaussian target $Q$
centered on the annotated joint~\cite{nibali2018numerical,sun2018integral}:
\begin{equation}
  Q_i=\frac{\exp[-d_i^2/(2\sigma^2)]}{\sum_j\exp[-d_j^2/(2\sigma^2)]},
  \qquad
  \mathcal L_{\mathrm{dist}}=\operatorname{JS}(P,Q).
  \label{eq:hand-heatmap-js}
\end{equation}
Here $d_i$ measures distance from the target in grid intervals on the decoder's endpoint grid,
$\sigma=0.6$, and JS denotes Jensen--Shannon divergence. We apply this objective at native decoder
times to joints with valid image and geometry annotations, including reliably annotated occluded
joints. We also supervise coordinate means because boundary truncation can shift the Gaussian's
expectation.
\FloatBarrier
\paragraph{Geometry-aware camera supervision}
Equal calibration-parameter errors can produce different ray errors across cameras. We therefore
scale log-focal and normalized principal-point penalties by their bearing sensitivities. For one
axis, with normalized coordinates and bearing $b=(u-c)/f$, we use
\begin{equation}
  \begin{aligned}
    \mathcal L_{\mathrm{camera}}={}&r_n\left[
    3.8\left\langle s_f\rho_{0.01}(\log\widehat f-\log f^\star)\right\rangle
    +4.1\left\langle s_c\rho_{0.005}(\widehat c-c^\star)\right\rangle\right],\\
    s_f={}&\left\langle\left|\frac{u-c^\star}{f^\star}\right|\right\rangle_{\mathrm{ray\ cells}},
    \qquad s_c=\frac{1}{f^\star}.
  \end{aligned}
  \label{eq:hand-camera-objective}
\end{equation}
We compute sensitivities from calibrated intrinsics and apply them outside SmoothL1, denoted by
$\rho_\delta$ with transition $\delta$, then average over valid cameras and both axes. We also
supervise the resulting unit rays with cosine distance. RHD contributes only joint and image
supervision. For both camera and distribution losses, we use the ramp $r_n=\min(1,(n+1)/500)$ at
zero-based optimizer step $n$.

\begin{table}[H]
  \centering\small
  \setlength{\tabcolsep}{4pt}
  \renewcommand{\arraystretch}{1.1}
  \caption{\textbf{Objective coefficients.} We average each component over its valid targets per source. The camera and distribution terms use the 500-step ramp $r_n$.}
  \label{tab:hand-loss-coefficients}
  \begin{tabularx}{\linewidth}{@{}>{\raggedright\arraybackslash}p{.34\linewidth} X l@{}}
    \toprule
    \textbf{Target} & \textbf{Residual / branch} & \textbf{Coefficient} \\
    \midrule
    Wrist-relative joints & Direct / MANO mean absolute error & $10/10$ \\
    Camera-frame joints & Direct / MANO mean absolute error & $2.5/2.5$ \\
    Camera-frame wrist & Direct / MANO mean absolute error & $1/1$ \\
    MANO rotations & Geodesic / rotation-matrix error & $1/1$ \\
    MANO shape / translation & Parameter mean absolute error & $0.1/1$ \\
    Image joints / wrist / projection & Normalized coordinate absolute error & $1/0.5/1$ \\
    Existence / visibility & Binary cross-entropy & $0.5/0.25$ \\
    Temporal joints & Valid-triple second difference & $0.5$ \\
    Unit-ray direction & Cosine distance & $1$ \\
    Log focal / principal point & Jacobian-scaled SmoothL1 & $3.8r_n/4.1r_n$ \\
    Auxiliary translation & Bearing-path $XY$ absolute error & $1$ \\
    Out-of-sight orientation & Global geodesic + matrix error & $1/32$ \\
    Joint readout distribution & Gaussian-target JS divergence & $0.1r_n$ \\
    \bottomrule
  \end{tabularx}
\end{table}

We optimize LoRA, patch embedding, decoder, and heads with AdamW while keeping the VAE and base
backbone frozen. Table~\ref{tab:hand-training-config} gives the optimization and loss-ramp
schedules. During inference, to reduce repeated computation in long recordings, we encode each video
once and batch overlapping reconstruction windows with a resident model. We use calibrated FP8
kernels for selected backbone and VAE operations and FP16 for the VAE input, output, and latent
projection.

\subsubsection{Sequence Reconstruction and Global Consistency}
\label{app:hand-outputs}

Independent window estimates can introduce shape and scale drift as image support changes. We extend
ACE's window-level shape prior to the full continuous segment, estimating one shape per hand and a
shared camera before regenerating geometry. We first fuse overlapping predictions at their source
timestamps, combining rotations in rotation space.

To reduce the influence of unreliable frames, we weight each shape estimate by hand existence
$e_{th}$, visibility $v_{th}$, and a gate $g_{th}$ for confidence, image support, and projection
consistency. We then refine the shared shape by Huber reweighting:
\begin{equation}
  \begin{aligned}
    w_{th}&=e_{th}^{2}v_{th}^{2}g_{th},
    &\widetilde w_{th}^{(k)}&=w_{th}\frac{1.5}{\max(1.5,d_{th}^{(k)})},\\
    \overline{\boldsymbol\beta}^{h,(k+1)}
    &=\frac{\sum_t\widetilde w_{th}^{(k)}\boldsymbol\beta_{th}}
    {\sum_t\widetilde w_{th}^{(k)}}.
  \end{aligned}
  \label{eq:hand-global-shape}
\end{equation}
We initialize each parameter with its weighted median and standardize residuals using the fixed
per-parameter scale $\max(1.4826\,\mathrm{MAD},0.05)$, using weighted median absolute deviations
from that initial estimate. The distance $d_{th}^{(k)}$ is the root-mean-square standardized
residual from the current shared shape. If gating supplies no reliable frames, we use equal base
weights in the same robust estimate.

For predicted cameras, we apply the same procedure to log focal lengths and normalized principal
points on the source canvas, using the maximum left/right hand weight per frame and respective scale
floors of 0.01 and 0.002. We use supplied intrinsics for calibrated sequences.

\paragraph{Geometry regeneration and translation refinement}
Updating shape and calibration changes the hand's image projection. We regenerate MANO geometry from
the shared shape and refine translation against two-dimensional anchors $\mathbf a_{thj}$, balancing
image alignment with a prior around the original placement of regenerated points $\mathbf Q_{thj}$:
\begin{equation}
  \min_{\mathbf t_{th}}
  \sum_j\omega_{thj}
  \rho\!\left(
  \left\|\pi_{\overline K}
  (\mathbf Q_{thj}+\mathbf t_{th})-\mathbf a_{thj}\right\|_2
  \right)
  +\tfrac12\left\|\mathbf t_{th}-\mathbf t_{th}^{\mathrm{prior}}\right\|_{\Sigma^{-1}}^2.
  \label{eq:hand-translation-refinement}
\end{equation}
Here $\rho$ is Huber loss, $\omega$ combines anchor support and hand reliability, and $\Sigma$
specifies lateral and depth prior scales. We optimize translation while fixing articulation and
rotation, enforce positive depth, and bound wrist-depth changes. With insufficient anchors, we
preserve the original projected wrist and depth.

\begin{figure}[htbp]
  \centering
  \includegraphics[width=0.95\linewidth]{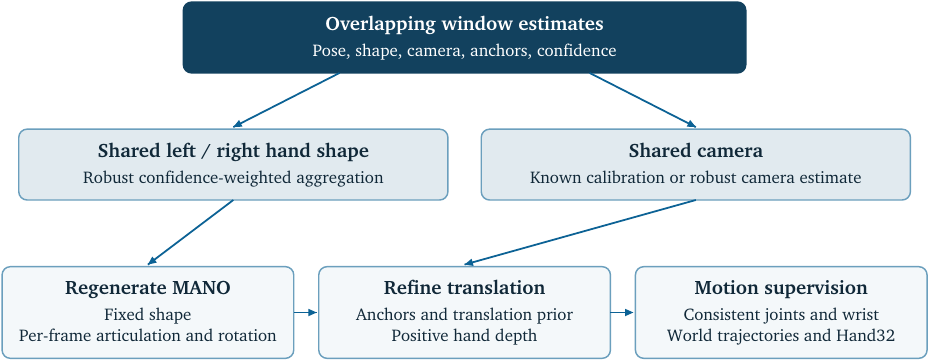}
  \caption{\textbf{Sequence-level consistency.} We aggregate reliable window estimates into shared hand shapes and calibration, regenerate geometry, and refine translation against image anchors.}
  \label{fig:ego-global-consistency}
\end{figure}

We export timestamp-aligned joints, MANO parameters, intrinsics, and confidence scores for
action-data preparation in Section~\ref{sec:dexterous-ego-data}.
Figure~\ref{fig:ego-global-consistency} summarizes the refinement.

\subsubsection{Qualitative Examples}
\label{app:hand-examples}

\begin{figure}[!htbp]
  \centering
  \includegraphics[width=0.95\linewidth]{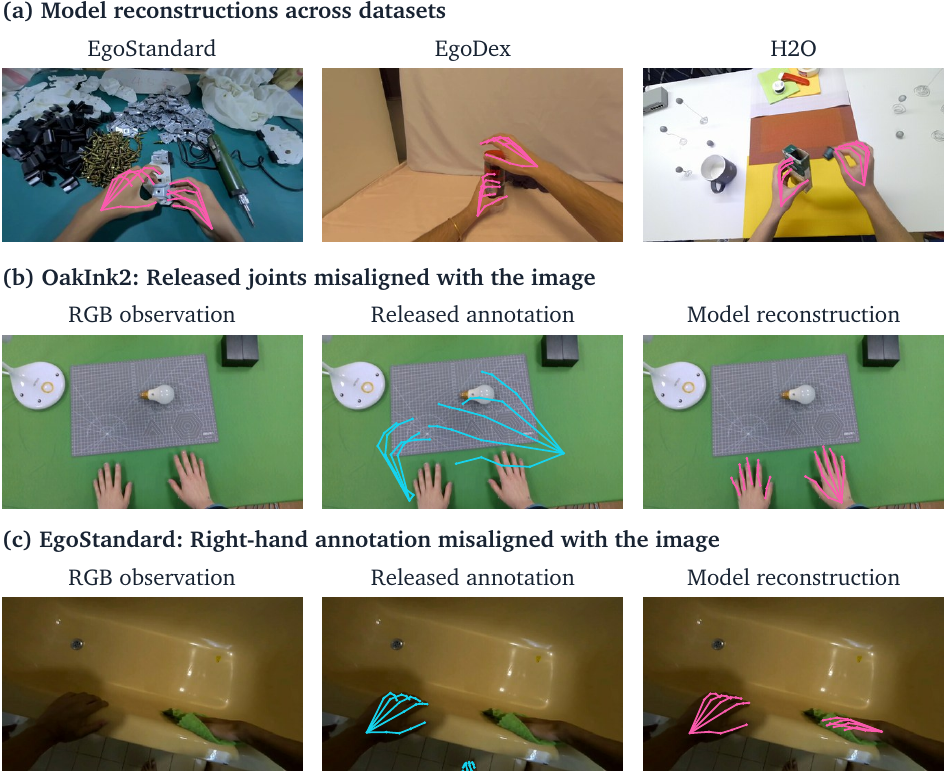}
  \caption{\textbf{Qualitative hand reconstruction across datasets.}
    (a) Model reconstructions on EgoStandard, EgoDex, and H2O.
    (b,c) RGB observations, released joint annotations (cyan), and model reconstructions (pink), shown from left to right for \textsc{OakInk2} and EgoStandard, respectively. Joint overlays use the provided camera calibration.}
  \label{fig:hand-qualitative-cases}
\end{figure}

Figure~\ref{fig:hand-qualitative-cases}(a) shows model reconstructions of bimanual manipulation in
EgoStandard, EgoDex, and H2O. Panels (b) and (c) compare the reconstructions with the released
annotations. In the \textsc{OakInk2} example, the released joints project across the tabletop, away
from the visible hands. In the EgoStandard example, the released right-hand joints project below the
partially occluded hand. In both cases, the reconstructed joints align more closely with the visible
hands. These examples motivate applying dataset-specific thresholds to annotation--image agreement
to flag and filter noisy supervision during data curation.

\subsection{Dexterous State and Action Details}
\label{app:dexterous-details}

\subsubsection{\texttt{Hand32} Definition}
\label{app:hand32}

Following the functional alignment principle of FAAS~\cite{zhang2026unidex},
\texttt{Hand32} assigns slots by finger, joint level, and motion function.
Different hands can therefore share an articulation format without identical
mechanical structures: index-finger flexion, for example, can occupy the same
slot despite different mechanisms.
Table~\ref{tab:app-hand32-slots} gives the slot definitions.

\begin{table}[htbp]
  \centering\footnotesize
  \setlength{\tabcolsep}{5pt}
  \caption{\textbf{\texttt{Hand32} slot definitions.} Entries are zero-based slot indices;
    angles use radians. \textbf{FE} denotes flexion/extension and \textbf{AA} abduction/adduction.
    \textbf{CMC}: carpometacarpal; \textbf{MCP}: metacarpophalangeal; \textbf{IP}: interphalangeal;
    \textbf{PIP}/\textbf{DIP}: proximal/distal interphalangeal.
    Slot 31 is reserved for a declared embodiment-specific extension.}
  \label{tab:app-hand32-slots}
  \begin{tabular*}{0.5\linewidth}{@{\extracolsep{\fill}}llccc@{}}
    \toprule
    Finger & Joint & FE & AA & Twist \\
    \midrule
    \multirow{3}{*}{Thumb} & CMC & 1 & 2 & 26 \\
    & MCP & 3 & 25 & 0 \\
    & IP & 4 & --- & --- \\
    \cmidrule(lr){1-5}
    \multirow{3}{*}{Index} & MCP & 7 & 6 & 5 \\
    & PIP & 8 & --- & 27 \\
    & DIP & 9 & --- & --- \\
    \cmidrule(lr){1-5}
    \multirow{3}{*}{Middle} & MCP & 12 & 11 & 10 \\
    & PIP & 13 & --- & 28 \\
    & DIP & 14 & --- & --- \\
    \cmidrule(lr){1-5}
    \multirow{3}{*}{Ring} & MCP & 17 & 16 & 15 \\
    & PIP & 18 & --- & 29 \\
    & DIP & 19 & --- & --- \\
    \cmidrule(lr){1-5}
    \multirow{3}{*}{Little} & MCP & 22 & 21 & 20 \\
    & PIP & 23 & --- & 30 \\
    & DIP & 24 & --- & --- \\
    \bottomrule
  \end{tabular*}
\end{table}

\FloatBarrier
\subsubsection{Mapping Dexterous Hands to \texttt{Hand32}}
\label{app:dexterous-adapters}

We map each hand's articulation to \texttt{Hand32} by motion function rather
than native joint numbering.
Several mechanical joints may contribute to one reference coordinate.
Validity masks mark unsupported motions, and special mappings retain motions
specific to each hand.
Table~\ref{tab:app-device-hand32} lists the device mappings.

\begin{table}[htbp]
  \centering\footnotesize
  \setlength{\tabcolsep}{4pt}
  \renewcommand{\arraystretch}{1.13}
  \caption{\textbf{\texttt{Hand32} mappings for representative dexterous hands.}
    Slot lists give State coverage; Shadow Action additionally excludes slot 25.
    Special mappings apply only to the listed hand.}
  \label{tab:app-device-hand32}
  \begin{tabularx}{\linewidth}{@{}l >{\raggedright\arraybackslash}p{.38\linewidth} X@{}}
    \toprule
    Hand & Mapped slots & Special mapping \\
    \midrule
    Sharpa Wave & 1--4, 6--9, 11--14, 16--19, 21--25, 31 & 31: little CMC \\
    DexHand021 & 1--4, 6--9, 11--14, 16--19, 21--24 & --- \\
    Allegro & 1--4, 6--9, 11--14, 16--19 & Four functional digits \\
    Orca & 1--4, 6--8, 11--13, 16--18, 21--23, 31 & 31: internal wrist \\
    Revo2 & 2--4, 7--8, 12--13, 17--18, 22--23 & --- \\
    RH56DFX & 2--4, 7--8, 12--13, 17--18, 22--23, 26 & 26: extra thumb flexion \\
    RH5DG2 & 2--4, 6--9, 11--14, 17--19, 22--24, 26 & 26: thumb DIP \\
    Schunk & 2--4, 6--9, 12--14, 16--19, 21--24, 26, 31 & 26: thumb hinge; 31: coupled palm-side motion \\
    Shadow & 1--4, 6--9, 11--14, 16--19, 21--25, 30--31 & 30/31: internal wrists \\
    Wuji & 1--4, 6--9, 11--14, 16--19, 21--24 & --- \\
    Ability & 2--3, 7--8, 12--13, 17--18, 22--23 & Retain recorded mimic commands \\
    LEAP & 1--4, 6--9, 11--14, 16--19 & Four functional digits \\
    \bottomrule
  \end{tabularx}
\end{table}

\subsubsection{Action-Learning Sources and the \texttt{Ego128} Layout}
\label{app:ego-action-data}

Table~\ref{tab:egocentric_data} lists the data sources used for action learning in
Section~\ref{sec:dexterous-action-data} and Section~\ref{sec:dexterous-ego-data}. Instrumented human
and glove recordings supply additional hand measurements and, where available, tactile observations.

\begin{table}[htbp]
  \centering\small
  \setlength{\tabcolsep}{3pt}
  \renewcommand{\arraystretch}{1.10}
  \caption{\textbf{Human and dexterous hand demonstration sources for action learning.} "Before filtering" counts local input episodes. "After filtering" counts retained training fragments in the indicated releases, including the effect of split selection. Hours use the original recording timeline, counting synchronized views once. EgoVerse is restricted to human recordings with a shorter side strictly above 480 pixels in our September 8, 2026 snapshot.}
  \label{tab:egocentric_data}
  \begin{tabularx}{\linewidth}{@{}>{\raggedright\arraybackslash}X rrrr@{}}
    \toprule
    & \multicolumn{2}{c}{Before filtering} & \multicolumn{2}{c}{After filtering} \\
    \cmidrule(lr){2-3}\cmidrule(l){4-5}
    Source & Episodes & Hours & Fragments & Hours \\
    \midrule
    \multicolumn{5}{@{}l}{\textit{Egocentric human demonstrations}} \\
    EgoDex~\cite{hoque2026egodex} & 314,839 & 721.93 & 303,391 & 700.80 \\
    EgoLive~\cite{li2026egolive} & 971,045 & 1,675.46 & 915,403 & 1,592.18 \\
    EgoPro~\cite{lightwheel2026egosuite} & 175,284 & 1,035.85 & 170,476 & 1,011.70 \\
    EgoStandard~\cite{lightwheel2026egosuite} & 1,185,239 & 5,687.81 & 1,160,797 & 5,589.46 \\
    EgoVerse~\cite{punamiya2026egoverse} & 89,690 & 101.74 & 80,406 & 94.47 \\
    Lightwheel (in-house) & 2,002,511 & 4,235.84 & 1,322,146 & 2,819.34 \\
    \midrule
    \multicolumn{5}{@{}l}{\textit{Instrumented human and glove demonstrations}} \\
    Xspark-HumanTouch~\cite{qin2026dextouch} & 17,608 & 132.75 & 17,576 & 132.30 \\
    PaXini OmniSharing~\cite{PXOmniSharingDB} & 44,176 & 335.34 & 41,508 & 317.44 \\
    SynData (glove)~\cite{SynData} & 36,780 & 145.46 & 35,303 & 120.39 \\
    \midrule
    \multicolumn{5}{@{}l}{\textit{Dexterous robot demonstrations}} \\
    FTP-1~\cite{yuan2026ftp} & 9,328 & 73.16 & 8,098 & 66.49 \\
    SharpaDex~\cite{sharpadex_v1_2026} & 28,993 & 299.96 & 28,403 & 294.04 \\
    RoCo Task Board Assembly~\cite{roco_task_board_assembly_2026} & 562 & 22.79 & 571 & 22.65 \\
    Robotic Origami Challenge~\cite{robotic_origami_challenge_fold_plane_2026,robotic_origami_challenge_2026} & 1,845 & 101.39 & 1,821 & 100.12 \\
    T-Rex~\cite{niu2026trex} & 5,464 & 50.68 & 5,254 & 48.53 \\
    HRDexDB (robot)~\cite{lim2026hrdexdb} & 1,630 & 9.20 & 1,592 & 8.15 \\
    Bench2Dex~\cite{yang2026bench2dex} & 1,300 & 12.55 & 1,281 & 12.37 \\
    \midrule
    \textbf{Total} & 4,886,294 & 14,641.91 & \textbf{4,094,026} & \textbf{12,930.43} \\
    \bottomrule
  \end{tabularx}
\end{table}

\begin{table}[htbp]
  \centering\small
  \setlength{\tabcolsep}{5pt}
  \renewcommand{\arraystretch}{1.13}
  \caption{\textbf{Ego128 State and Action.} Indices use zero-based, half-open intervals. Wrist and camera poses use the nine-dimensional position--rotation encoding. Hand articulation targets are absolute angles at the next contiguous observation.}
  \label{tab:ego128-layout}
  \begin{tabularx}{\linewidth}{@{}l l X X@{}}
    \toprule
    \textbf{Channels} & \textbf{Group} & \textbf{State at $t$} & \textbf{Action from $t$} \\
    \midrule
    $[0,9)$ & Left wrist & $\operatorname{pose9}(X_t^L)$ & $\operatorname{pose9}(D_t^L)$ \\
    $[10,19)$ & Right wrist & $\operatorname{pose9}(X_t^R)$ & $\operatorname{pose9}(D_t^R)$ \\
    $[20,52)$ & Left \texttt{Hand32} & $q_t^L$ & $q_{t+1}^L$ \\
    $[52,84)$ & Right \texttt{Hand32} & $q_t^R$ & $q_{t+1}^R$ \\
    $[84,93)$ & Main camera & $\operatorname{pose9}(S_t^C)$ & $\operatorname{pose9}(D_t^C)$ \\
    $9$, $19$, $[93,128)$ & Reserved & Zero, inactive mask & Zero, inactive mask \\
    \bottomrule
  \end{tabularx}
\end{table}
The \texttt{Ego128} State and Action layouts in Table~\ref{tab:ego128-layout} use the
camera-relative wrist and anchored camera quantities defined in Equation~\ref{eq:ego-state-action};
wrist targets $D_t^h$ follow the local-increment convention in
Equation~\ref{eq:canonical-tcp-action} at the canonical wrist. Each pose is encoded as
$\operatorname{pose9}(T)=[\mathbf p;\mathbf R_{:,1};\mathbf R_{:,2}]$, with translation in meters;
hand articulation is in radians. State records current articulation and Action records absolute
articulation at the next contiguous observation.

\FloatBarrier
\section{Tactile Sensing}
\subsection{Tactile Data Processing Details}
\label{app:tactile-data-details}

\subsubsection{Surface Calibration and Source Preparation}
\label{app:tactile-calibration}
\label{app:tactile-cleaning}

\paragraph{Calibration tool}
The operator first aligns source geometry to neutral MANO by a whole-hand
similarity transform, then adjusts each sensor module's position,
orientation and geometric scale before surface snapping. Review checks
finger/phalanx identity, palmar/dorsal placement and local directions
in neutral and articulated poses. PaXini uses 15 modules and 1,005
samples/hand; HumanTouch uses 360 assigned atlas samples/hand.
Source pixels or taxels retain their original identifiers.

Each attachment stores a triangle $f_i$, barycentric weights
$\boldsymbol\beta_i$, and neutral contact frame $Q_i^0$. Posed locations are
\begin{equation}
  x_i(t)=\sum_{k=1}^3\beta_{ik}V_{f_i,k}(t),\qquad
  \sum_k\beta_{ik}=1,\quad\beta_{ik}\geq0.
  \label{eq:app-tactile-barycentric}
\end{equation}
The tangent is transported as $A_tA_0^+q_x^0$, using the neutral/posed
triangle edge matrices, then orthogonalized against the posed normal.
The resulting $Q_i$ has tangent-X, tangent-Y and normal-Z columns;
local vectors map as $Q_i g_i$. Automated checks enforce unique IDs,
valid triangles, barycentric reconstruction and $Q_i\in SO(3)$.
Left/right bindings are reviewed separately. Geometry, topology and
calibration hashes identify caches.

\paragraph{Source preparation}
Table~\ref{tab:app-tactile-coverage} lists source-specific observations.
Normal-only response $a_i$ forms $a_i n_i$ along its mapped normal.
The Sharpa Step4f-v2 cache subtracts the first-frame baseline, retains
positive residuals of at least 8, requires 0.5\% active area, and confirms
contact after four frames with offline onset backfill; release is immediate.
These settings are specific to that cache version. Coverage, contact and
feature validity are stored separately; only covered no-contact regions
supply negative presence labels.

\begin{table}[htbp]
  \centering\small
  \setlength{\tabcolsep}{4pt}
  \caption{\textbf{Representative tactile-source adapters and coverage rules.}
    Signal interpretation, surface correspondence, and coverage are
    recorded independently.}
  \label{tab:app-tactile-coverage}
  \begin{tabularx}{\linewidth}{@{}lXX@{}}
    \toprule
    Source / layout & Mapped observation & Coverage handling \\
    \midrule
    PaXini glove & Local directional array responses; calibrated surface anchors & Separate left/right IDs; palm and finger bindings retained \\
    HumanTouch atlas & Relative normal response & 360 assigned samples/hand; padding excluded \\
    Sharpa FTP & Optical fingertip response & Thumb/index/middle; ring/little invalid \\
    RoCo / Origami / T-Rex & Source fingertip deformation streams & Declared hand/finger tiles and mapping version \\
    Bench2Dex & Surface-aligned geometric contact signal~\cite{yang2026bench2dex} & Hand-specific sensing surface and contact validity \\
    Single tactile pad & Source-supported pad observation & Assigned index-region role \\
    UniVTac two-jaw layout & Two separately mapped sensing faces & Right thumb/index roles; other regions masked \\
    \bottomrule
  \end{tabularx}
\end{table}

\subsubsection{Proxy Construction and Encoding}
\label{app:tactile-proxy}

For each region, freeze origin $o_r$, right-handed basis $B_r$ and scale
$\ell_r>0$. Positions, UV coordinates and loads are
\begin{equation}
  y_i=B_r^\top(x_i-o_r),\qquad
  u_i=(y_{i,x},y_{i,y})^\top/\ell_r,\qquad f_i=B_r^\top f_i^M.
  \label{eq:app-region-chart}
\end{equation}
UV is dimensionless; area and moment arms use three-dimensional geometry.
Table~\ref{tab:app-proxy-fields} gives the complete layout.

\begin{table}[htbp]
  \centering\small
  \setlength{\tabcolsep}{5pt}
  \caption{\textbf{Regional proxy layout.} Index ranges are zero-based and
    half-open. Feature validity is recorded separately.}
  \label{tab:app-proxy-fields}
  \begin{tabularx}{\linewidth}{@{}l c l X@{}}
    \toprule
    Indices & Width & Symbol & Definition \\
    \midrule
    $[0,2)$ & 2 & $c$ & Geometric UV center \\
    $[2,4)$ & 2 & $\bar c$ & Load-weighted UV center \\
    $[4,7)$ & 3 & $F$ & Local resultant load \\
    $[7,10)$ & 3 & $\tau$ & Moment about the load-weighted 3D center \\
    $[10,11)$ & 1 & $A$ & Active support area \\
    $[11,14)$ & 3 & $b$ & Stable oriented extent and anisotropy encoding \\
    $[14,20)$ & 6 & $\operatorname{symvec}(C)$ & Traction covariance \\
    $[20,23)$ & 3 & $\operatorname{symvec}(S)$ & Load-weighted UV covariance \\
    $[23,26)$ & 3 & $t_{\mathrm{peak}}$ & Peak traction vector \\
    $[26,28)$ & 2 & $u_{\mathrm{peak}}$ & Peak UV location \\
    \bottomrule
  \end{tabularx}
\end{table}

For valid contact samples $\mathcal I$, support areas $a_i>0$, and
$w_i=\max(\lVert f_i\rVert_2,\epsilon)$, compute
\begin{equation}
  \begin{aligned}
    c&=\frac{1}{|\mathcal I|}\sum_i u_i,&
    \bar y&=\frac{\sum_iw_i y_i}{\sum_iw_i},&
    \bar c&=(\bar y_x,\bar y_y)^\top/\ell_r,\\
    F&=\sum_i f_i,&
    \tau&=\sum_i(y_i-\bar y)\times f_i,& A&=\sum_i a_i.
  \end{aligned}
  \label{eq:app-proxy-macro}
\end{equation}
Here $f_i$ is a sample load; calibrated pressure/traction is multiplied
by physical support area first. Vertex support can use one third of
incident triangle areas, or calibrated native areas when available.
Uncalibrated inputs retain source-response units.

Principal UV axes give extents $L\geq W$ and angle $\theta$ modulo $\pi$.
The stable shape descriptor is
\begin{equation}
  b=[L+W,\;\eta\cos(2\theta),\;\eta\sin(2\theta)],\qquad
  \eta=(L-W)/(L+W).
  \label{eq:app-proxy-shape}
\end{equation}
Singleton or zero-extent contacts use zero shape descriptors.
With traction $t_i=f_i/a_i$ and $\mu_t=\sum_i a_it_i/A$,
\begin{equation}
  \begin{aligned}
    C&=A^{-1}\sum_i a_i(t_i-\mu_t)(t_i-\mu_t)^\top,\\
    S&=(\sum_iw_i)^{-1}\sum_iw_i(u_i-\bar c)(u_i-\bar c)^\top.
  \end{aligned}
  \label{eq:app-proxy-covariance}
\end{equation}
Packing orders are $(xx,yy,zz,xy,xz,yz)$ for $C$ and $(uu,uv,vv)$ for $S$.
The peak is $i^*=\arg\max_i\lVert t_i\rVert_2$, storing
$(t_{i^*},u_{i^*})$. Empty contacts use zero proxy storage and disabled
contact-attribute losses.

\paragraph{Numerical encoding}
Offline source normalization uses train-split statistics and shared
scales for coupled XYZ components. Proxy encoding additionally uses
full-region sample count, spatial scale and support area to scale
resultant, moment and geometric fields. Normalizer identity and inverse
parameters are recorded. Proxies computed after dense-field clipping
need not be affine-invertible to raw proxies, so both views are retained.
The training reader validates published values and timestamps without
repeating calibration or normalization fitting.

\FloatBarrier
\subsection{Tactile Modeling Details}
\label{app:tactile-model-details}

\subsubsection{Inputs and Temporal Alignment}
\label{app:tactile-tokenization}
\label{app:tactile-time}

The input combines independent proxy/presence projections with source
and anatomical metadata:
\begin{equation}
  \begin{aligned}
    H^{(0)}={}&E_p(p_\sigma\odot M_p^{\mathrm{in}})
    +g_zE_z(z_\sigma\odot M_z)\\
    &+e_{\mathrm{sensor}}+e_{\mathrm{normalizer}}+e_{\mathrm{part}}
    +e_{\mathrm{role}}+E_s(s,M_s)+e_{\mathrm{modality}}.
  \end{aligned}
  \label{eq:app-tactile-codec}
\end{equation}
The learnable gain starts at $g_z=0.1$; role distinguishes history/target.
Part IDs are $r+64h$, with $r=0,\ldots,4$ from thumb to little finger and
$h=0,1$ for left/right. Projection routes and normalizer IDs are separate.

The configuration uses nine real observations per available modality
and a 1.2-second target horizon. Target rows follow a half-open interval;
the anchor row can appear in both context and target under distinct roles.
Missing history is rejected and padded targets are invalid.
Different native rates yield different history durations.
The loader uses Eq.~\eqref{eq:wam-fps-rope} with floating-point timestamps
and spatial rotary coordinates $(r,h)$. Historical pose context uses
the corresponding State; target context uses current State.

\subsubsection{Loss Implementation}
\label{app:tactile-loss-details}

Proxy and presence reduce over separate valid-element counts:
\begin{equation}
  \operatorname{MSE}_M(\hat v,v^*)=
  \frac{\sum_j M_j(\hat v_j-v_j^*)^2}{\max(1,\sum_j M_j)}.
  \label{eq:app-tactile-masked-mse}
\end{equation}
Invalid entries remain zero before and after noise interpolation.
The auxiliary BCE readout uses the presence decoder with detached hidden states:
\begin{equation}
  \ell_\sigma=2\big[z_\sigma-\sigma D_z(\operatorname{stopgrad}(H_\sigma))\big].
  \label{eq:app-presence-aux}
\end{equation}
Thus, BCE updates that decoder, while flow losses train the ordinary
encoder/Transformer/decoder paths. In Eq.~\eqref{eq:tactile-prediction-loss},
$\lambda_z=0.25$ and $\lambda_{\mathrm{BCE}}=0.02$.
An additional $0.05L_{\mathrm{geo}}$ penalizes reconstructed UV centers:
\begin{equation}
  L_{\mathrm{geo}}=
  \frac{\sum_{j\in\mathcal U}M_{p,j}
    [\max(|\hat p_{0,j}|-1,0)]^2}
  {\max(1,\sum_{j\in\mathcal U}M_{p,j})},\quad
  \hat p_0=p_\sigma-\sigma\hat v_p,\quad\mathcal U=\{0,1,2,3\}.
  \label{eq:app-tactile-geometry}
\end{equation}
The penalty is soft. The presence gate is applied once at final readout,
with raw outputs retained for diagnostics.

\FloatBarrier
\section{Infrastructure Details}\phantomsection\label{app:infra}

\subsection{Experimental Conditions}\phantomsection\label{app:infra:experiments}

The diagrams show four token types (U/V/A/T) across three towers. The Action and Tactile Generation
Tower places A before T. The historical measurements use no tactile tokens ($T=0$) and span
different workloads and implementation snapshots; their gains cannot be combined into a speedup for
the released model. \tabref{tab:infra:experiments} lists the training comparisons,
\tabref{tab:infra:configuration} the utilization configurations, and
\tabref{tab:infra:inference-protocol} the inference protocols. Throughput windows exclude startup
and compilation; training time excludes offline preprocessing and storage.

\begin{table}[htbp]
  \centering\small
  \setlength{\tabcolsep}{4pt}
  \renewcommand{\arraystretch}{1.08}
  \caption{\textbf{Training studies and implementation scope.} All GPUs have 80~GB memory. Optional experiments do not establish that these paths trained the released checkpoints. U/V/A/T denote understanding/video/action/tactile capacities; T is zero in these historical studies.}\label{tab:infra:experiments}
  \begin{tabularx}{\linewidth}{@{}>{\raggedright\arraybackslash}p{.21\linewidth} >{\raggedright\arraybackslash}p{.364\linewidth} >{\raggedright\arraybackslash}X@{}}
    \toprule
    \textbf{Study / role} & \textbf{Workload and implementation} & \textbf{Measurement scope} \\
    \midrule
    Static execution /\newline configured training path & Three-tower DROID, \texttt{width\_only\_1b}, without state conditioning. Frozen World Understanding Tower; SDPA, routed projections; PyTorch~2.10/CUDA~13.0 & Eight GPUs, online VAE, BF16, full~recomputation, global batch size~32; three~runs/configuration, updates~11--40;~\tabref{tab:infra:static-config} \\
    Grouped FA3 / experiment & T2V, eight GPUs, shard8/replica1, GA2, U3,072/V98,304; PyTorch~2.10/Triton~3.6. \href{https://github.com/Hchnr/flash-attention/blob/6c9c171a74a7a251f2cab8466d33d1f672e157ad/hopper/flash_attn_interface.py\#L1250}{Kernel source \texttt{6c9c171a}}. Tested~revision~\texttt{f6447a5d} & Same-fork split/grouped comparison, one~run~each, ten warmup and thirty timed~updates; rank-0 kernel counts from two separate profiled updates \\
    ODC / experiment & T2V, eight GPUs, shard8, GA2, online VAE, 64K; snapshot \texttt{016ac4e6} & One 100-update run per arm; timing over 51--94, traces over 96--98 across eight~GPUs \\
    Reassignment / prototype & T2V, eight GPUs, replica4/shard2, GA4, offline~VAE; U512/V32,768 & Four packs/rank/window, 0.2~s search budget; seventeen cohort-matched updates, three~excluded \\
    Shared loader /\newline adopted in tested recipe & Action, eight GPUs, 23 sources, same i8800 weights/Wan2.2 VAE, seed42; 74,368-token budget, shard8/GA2 & Complete updates; W\&B and checkpoint writes disabled. Compile, repeat, all-column and background controls as~in~\figref{fig:infra:loader} \\
    Offline latents /\newline T2V path & Four GPUs, shard4, BF16; U19,456/V25,600; online/offline catalogs & Two captured rank-0 updates; shows VAE-path removal, not a matched end-to-end speedup \\
    \bottomrule
  \end{tabularx}
\end{table}

\textbf{96K configurations.} The production configuration uses shard16/replica8/GA2 on 128 GPUs. Historical rank-0 summaries report roughly 18.5 samples/47 s for 64K offline and 15 samples/35 s for 96K offline: approximately $0.394 \rightarrow 0.429$ samples/s/GPU, or 8.9\% higher throughput. These stage summaries are rounded; exact averaging windows were not retained. The later FA3 benchmark adapts the 96K recipe to eight GPUs, shard8/replica1, a read-only checkpointer, and the recorded kernel build. At the same U/V capacity, peak allocated memory is 64.13--64.22 GiB; this is not a matched shard8/shard16 capacity ablation. A separate compiler fix sets the reduction split size to 32: the failing tile needed 278,688 bytes, exceeding the 232,448-byte shared-memory limit.

\begin{table}[htbp]
  \centering\small
  \setlength{\tabcolsep}{5pt}
  \caption{\textbf{Static-execution ablation in \figref{fig:infra:static}.} The global batch size is fixed at 32. Fixed-capacity rows use U/V/A/T capacities of 512/2,688/128/0 tokens per pack. Times are the median and range of three run-mean complete-update times. Workers are counted per rank. Prefetch is the factor per worker. The variant name does not specify total model parameter count.}\label{tab:infra:static-config}
  \begin{tabularx}{\linewidth}{@{}lcc>{\raggedright\arraybackslash}Xccr@{}}
    \toprule
    \textbf{Stage} & \textbf{Pack} & \textbf{GA} & \textbf{Capacity} & \textbf{Compile} & \textbf{Workers / prefetch} & \textbf{s/update [range]} \\
    \midrule
    Baseline & 1 & 4 & Dynamic & Off & 8/2 & 3.320 [3.305, 3.323] \\
    Packing & 4 & 1 & Dynamic & Off & 8/2 & 1.429 [1.420, 1.429] \\
    + Fixed & 4 & 1 & Fixed & Off & 8/2 & 1.042 [1.037, 1.045] \\
    + Compile & 4 & 1 & Fixed & Static & 8/2 & 0.871 [0.866, 0.873] \\
    + Workers & 4 & 1 & Fixed & Static & 18/2 & 0.819 [0.811, 0.822] \\
    \bottomrule
  \end{tabularx}
\end{table}

\begin{table}[htbp]
  \centering\small
  \setlength{\tabcolsep}{4pt}
  \renewcommand{\arraystretch}{1.08}
  \caption{\textbf{Historical three-tower inference protocols.}
    These exclude state and tactile inputs and differ from the current
    Ray backend. The CFG parallelism study uses equally weighted input/block p50 summaries with an interpolated baseline; GPU cost uses mean duration. Its twenty-observation, two-repeat quality check is separate from timing.}\label{tab:infra:inference-protocol}
  \begin{tabularx}{\linewidth}{@{}l>{\raggedright\arraybackslash}X>{\raggedright\arraybackslash}X@{}}
    \toprule
    \textbf{Study} & \textbf{Inputs and execution} & \textbf{Timer / repetitions} \\
    \midrule
    CFG & 80~GB GPU; \texttt{width\_only\_1b}; $960 \times 640$, 41~frames/40~actions, 6,600 video tokens; five sampling steps at a CFG scale of~1.5; action-only & From image preparation to action delivery on CPU; three ABA blocks, two inputs, three warmup and twenty timed requests per input/profile; excludes~startup/HTTP \\
    FFN reuse & Three timing inputs; separate six-task quality set; one GPU, dual CFG branches, ten denoiser~evaluations & Synchronous execution after preparation, excluding preparation/HTTP; two sessions, twenty-four AB/BA pairs/input/session \\
    Queue & Four one-GPU replicas; 128~requests/configuration/rate; fixed-interval arrivals, client concurrency eight, eight-second~deadline & Timely responses per allocated GPU per wall time from launch to final drain; one finite replay per~configuration/rate \\
    \bottomrule
  \end{tabularx}
\end{table}
\FloatBarrier

\subsection{Execution Mechanisms and Tradeoffs}\phantomsection\label{app:infra:execution}

\subsubsection{Grouped Attention}\phantomsection\label{app:infra:fa3}

\begin{figure}[htbp]
  \centering
  \includegraphics[width=\linewidth]{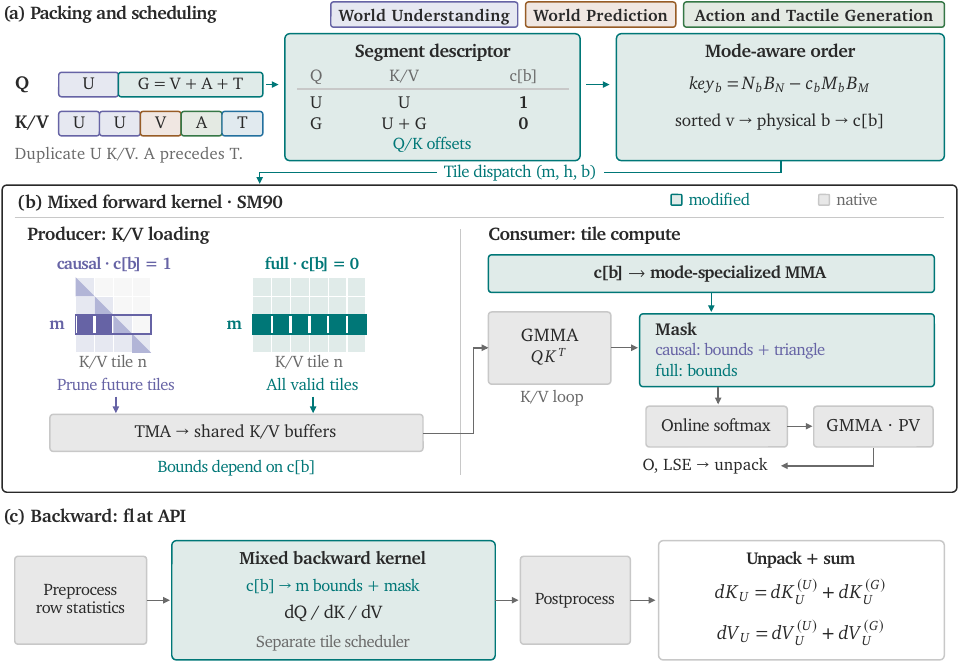}
  \caption{\textbf{Grouped FA3 internals.} (a) U and G = V + A + T form causal and full-attention segments; A/T share the Action and Tactile Generation Tower's attention projections, with A first. Flag c[b] identifies physical segment b. Q/K offsets are retained during mode-aware sorting; $M$/$N$ count query/key tiles, and $B_M$/$B_N$ give tile sizes. (b) The persistent scheduler dispatches query tile m, head h, and segment b; one forward main kernel uses mode-specific bounds and masks. (c) The all-gradient flat API uses a mixed backward main kernel and a separate scheduler. Pre/postprocessing remain separate; framework autograd sums gradients for duplicated U keys and values. Tile grids are schematic.}\label{fig:infra:fa3}
\end{figure}

The upstream FA3 variable-length interface takes separate cumulative query and key offsets and one
causal setting per call~\citep{infra_fa-upstream}. Causal and full attention therefore require
separate calls. We retain the offsets and pass a device-resident vector of causal flags, one per
segment. U queries attend causally to U, and generation queries attend to all valid context in the
same sample. The descriptor packs $G=V+A+T$ and keeps A before T in the Action and Tactile
Generation Tower. Mode-aware sorting changes tile dispatch order while preserving segment identity
(\figref{fig:infra:fa3}).

The forward pass uses one main kernel with mode-specific tile bounds and masks. The backward pass
uses a separate mixed scheduler and accumulates gradients for duplicated U keys/values; packing,
preparation, and gradient conversion remain separate. In the historical T2V trace, forward
main-kernel launches decrease $576 \rightarrow 288$ and backward launches $288 \rightarrow 144$. One
unprofiled comparison reports median $41.410 \rightarrow 41.311$ s/update: long generation attention
dominates runtime despite fewer dispatches. 

\FloatBarrier
\subsubsection{On-Demand Communication}\phantomsection\label{app:infra:odc}\label{sec:infra:communication}

\begin{figure}[htbp]
  \centering
  \includegraphics[width=\linewidth]{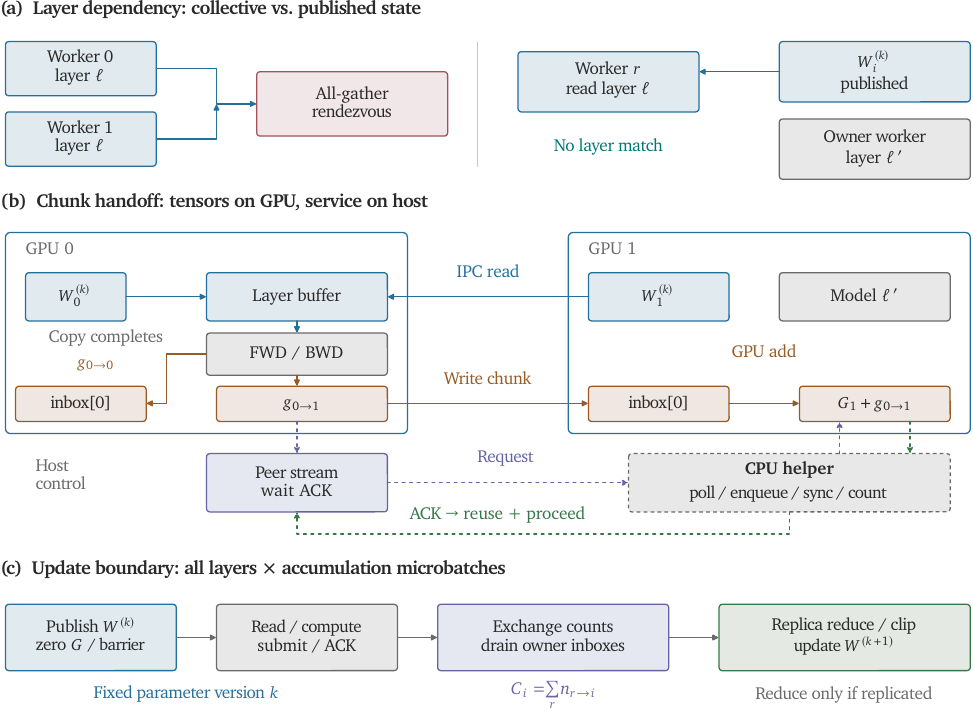}
  \caption{\textbf{ODC dependencies at two scales.} (a) A collective requires both workers at layer $\ell$; a peer read requires the owner's published shard, not its worker at $\ell$. (b) Same-node handoff separates GPU buffers from the CPU helper. $W_i^{(k)}$/$G_i$ are owner $i$'s parameter/gradient shards; $g_{r\to i}$ is sender $r$'s contribution. Local contributions use the same inbox protocol (helper 0 omitted). Requests/ACKs carry control; solid arrows carry tensors. (c) Update $k$ holds $W^{(k)}$ fixed across layers and GA microbatches; submitted counts $n_{r\to i}$ must match completed count $C_i$ before the optimizer. Geometry is schematic.}\label{fig:infra:odc}
\end{figure}

FSDP parameter all-gather requires peers to enter the same collective. We integrate On-Demand
Communication (ODC)~\citep{infra_odc} with FSDP2 and synchronous gradient accumulation, allowing a
ready worker to read published shards without waiting for each owner's model worker. Node-local CUDA
IPC transfers still require copy completion. Gradient senders write to inboxes and signal CPU
helpers, which schedule accumulation and acknowledge when buffers can be reused; NVSHMEM provides
control signaling~\citep{infra_nvshmem}. The optimizer still waits for all expected contributions,
replica reduction, and clipping (\figref{fig:infra:odc}).

\tabref{tab:infra:odc-events} summarizes communication changes over updates 96--98. In a separate unprofiled window on eight GPUs, shard8/GA2 with online VAE takes $25.21 \rightarrow 27.36$ s/update over updates 51--94 (throughput $-7.84\%$; one run/arm). Inbox handling, requests, and completion add overhead; overlapping trace durations cannot be subtracted from wall time to estimate net savings. ODC removes layerwise rendezvous, but this workload runs slower and has not passed numerical acceptance (\appref{app:infra:correctness}).

\begin{table}[htbp]
  \centering\small
  \setlength{\tabcolsep}{6pt}
  \caption{\textbf{ODC communication changes.} Eight-GPU trace-event totals over updates 96--98. These counts describe protocol changes, not saved critical-path time.}\label{tab:infra:odc-events}
  \begin{tabularx}{\linewidth}{@{}Xrr@{}}
    \toprule
    \textbf{Event} & \textbf{Native FSDP} & \textbf{ODC} \\
    \midrule
    Parameter all-gather & 3,536 & 80 \\
    Shard reduce-scatter & 1,776 & 0 \\
    All-reduce & 824 & 841 \\
    NVSHMEM request / wait & 0 / 0 & 41,856 / 41,856 \\
    \bottomrule
  \end{tabularx}
\end{table}

\FloatBarrier
\subsubsection{Workload Reassignment}\phantomsection\label{app:infra:balance}\label{sec:infra:balance}

\begin{figure}[htbp]
  \centering
  \includegraphics[width=\linewidth]{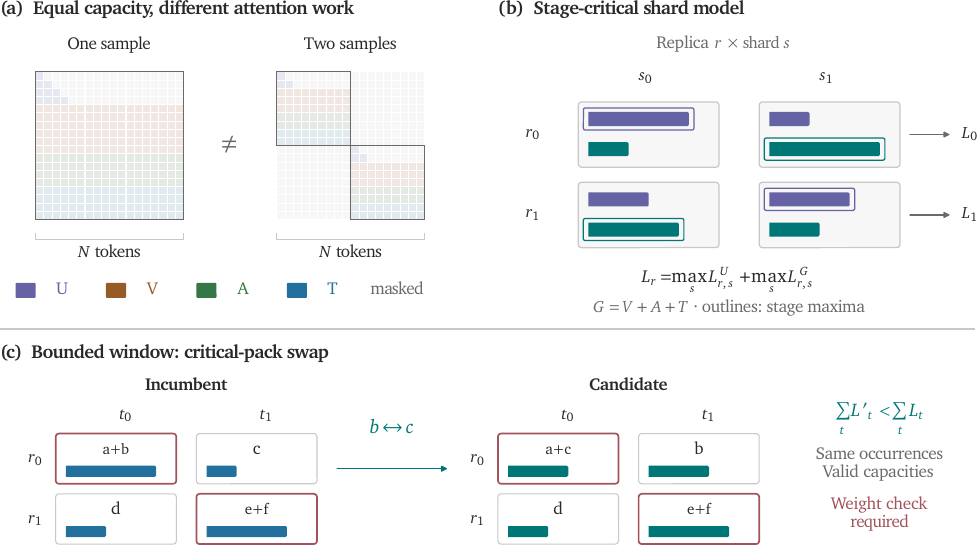}
  \caption{\textbf{Cost-aware reassignment.} (a) Packs with equal capacity have different visible U/V/A/T attention areas, in that sequence order. (b) Stage-cost model: each replica $r$ is limited by its slowest shard $s$ in each cost component; $G=V+A+T$ denotes generation work, not a separate attention operation. (c) Two replicas with one shard each: columns are microbatches $t$, rows are ranks, a--f are samples, and bars show predicted costs. Swapping b/c lowers the sum of column maxima $L_t$ while preserving occurrences and capacities; red outlines mark maxima. Loss-weight invariance requires a separate check. All geometry is schematic.}\label{fig:infra:balance}
\end{figure}

At fixed physical capacity, valid attention work depends on sample layout
(\figref{fig:infra:balance}). For sample $p$, define $G_p=V_p+A_p+T_p$; with causal U and full
generation queries, the total number of visible query--key pairs is
\[
  \mathcal A=\sum_p\left[\frac{U_p(U_p+1)}{2}+G_p(U_p+G_p)\right].
\]
At fixed buffer size, dense projections and FFNs perform largely fixed work. Our prototype holds the
HSDP mesh fixed and fits stage costs from $\sum U_p^2$, $\sum G_p^2$, and $\sum U_pG_p$,
complementing parallelism and scheduling methods such as Hydraulis~\citep{infra_hydraulis},
ByteScale~\citep{infra_bytescale}, and DistTrain~\citep{infra_disttrain}.

Starting from sequential packing, ranks all-gather complete Python sample objects and repeat a
deterministic swap search within a time limit. With microbatch $t$, replica $r$, and shard $s$, the
fitted variable-cost objective is
\[
  \sum_t\max_r\left[\max_s L^U_{t,r,s}+\max_s L^G_{t,r,s}\right].
\]
Pack latency, summed replica load, and reordering distance break ties. Checks verify capacity and
sample occurrences; equivalent sample/loss weighting across ranks and GA remains unvalidated.

The T2V experiment uses $A=T=0$, eight GPUs, replica4/shard2, GA4, offline VAE, and four packs per
rank in each window. Across seventeen cohort-matched updates, core time decreases
$20.7519 \rightarrow 20.4973$ s ($+1.24\%$ throughput). Coordination takes 0.2104 s/window, leaving
$+0.18\%$ complete-update throughput ($p=0.468$; time-saving 95\% CI $[-0.0685,0.1426]$ s). Three
updates are excluded because their cohorts differ; held-out cost prediction gives $R^2=-0.097$.
Coordination overhead and assignment quality limit adoption.

\FloatBarrier
\subsubsection{Compute and Memory Policies}\phantomsection\label{app:infra:compute}

Attention-only recomputation avoids repeated FFN computation but retains more
activations~\citep{infra_recompute}. \tabref{tab:infra:compute} reports this tradeoff alongside
experimental FP8 computation in the World Prediction Tower MLPs and VAE compilation. The studies use
different workloads and timing windows; their gains are not cumulative, and FP8 model-quality
validation remains incomplete.

\begin{table}[htbp]
  \centering\small
  \setlength{\tabcolsep}{4pt}
  \renewcommand{\arraystretch}{1.10}
  \caption{\textbf{Independent studies on eight GPUs.} Recomputation uses matched input/RNG fingerprints, ten warmup and thirty timed updates per run. FP8 quality validation remains incomplete.}\label{tab:infra:compute}
  \begin{tabularx}{\linewidth}{@{}ll>{\raggedright\arraybackslash}X@{}}
    \toprule
    \textbf{Policy / measured workload} & \textbf{Result} & \textbf{Cost and scope} \\
    \midrule
    Attention-only recomputation / T2V & $29.821 \rightarrow 25.979$ s/update & Peak allocated memory\newline $47.30 \rightarrow 64.50$~GiB; three~runs/mode \\
    FP8 MLP in World Prediction Tower / T2V & $45.944 \rightarrow 42.964$ s/update & Allocated memory\newline $64.28 \rightarrow 66.90$~GiB; one~run/mode \\
    Ahead-of-time VAE compilation / Action & $12.173 \rightarrow 11.007$ s compute & Four timed updates;\newline excludes input, transfer, compilation \\
    \bottomrule
  \end{tabularx}
\end{table}

\subsection{Supplementary Measurements}\phantomsection\label{app:infra:measurements}

\textbf{Input supply.} The compiled loader comparison uses a per-source baseline with 23 workers per rank and prefetch factor~2 (46 legacy credits per rank). The shared pool uses four workers per rank and a budget of $K=16$. Throughput changes by $+0.63\%$. Independent repeats reduce memory by 52.13\%/51.94\%, with throughput changes of $+2.73\%$/$-0.21\%$. Reading all columns reduces memory by 27.71\% at $+2.74\%$ throughput. The background-delivery variants use 23 workers per rank and $K=8$. Both pin memory; the newer 58-source recipe disables pinning to avoid locking the prepared window. Column projection, bounded reads, seek-prefix reuse, and selective RGB conversion further limit materialization.

\begin{figure}[htbp]
  \centering
  \includegraphics[width=\linewidth]{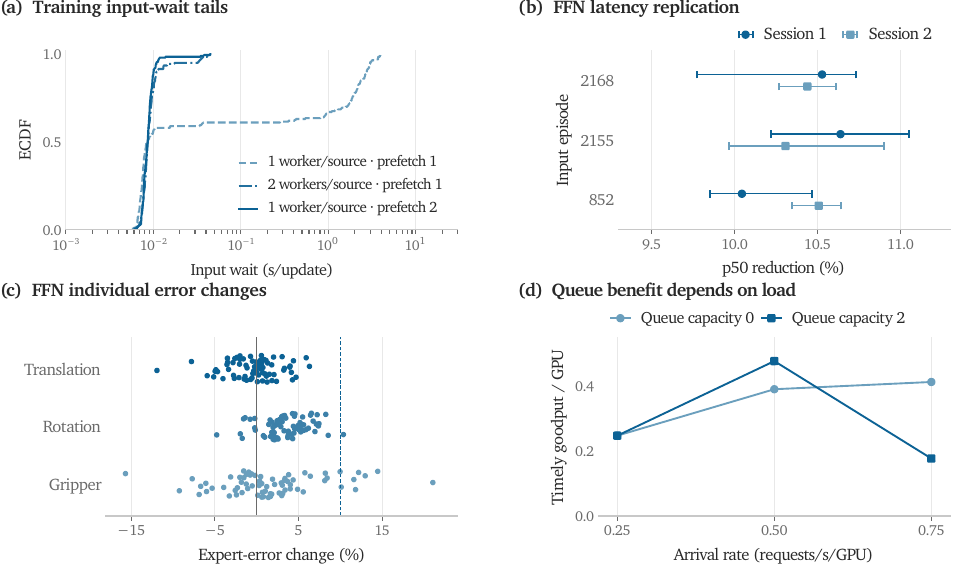}
  \caption{\textbf{Supplementary evidence.} (a) Twelve-source, eight-GPU loader study: one run/configuration, twenty warmup/200 timed updates. Wait is summed over GA2 microbatches per rank, then maximized across ranks. The legend counts workers per source and gives the prefetch factor per worker. (b) FFN latency in Sessions 1 and 2, with three identical inputs and twenty-four pairs per input per session. Intervals are 95\% paired circular block-4 bootstrap intervals. (c) Seventy-two paired generations/metric from six tasks, twelve episodes, three seeds; $+10\%$ is a task-level reference. (d) Queue capacities of zero or two pending requests per replica, with four replicas and 128~requests/configuration/rate.}\label{fig:infra:supplement}
\end{figure}

\figref{fig:infra:supplement}(a) extends the short loader comparisons. With twelve sources, the configurations with workers per source and prefetch factor $(1,1)$, $(2,1)$, and $(1,2)$ average 21.7407, 20.9629, 20.9606 s/update, with input-wait p95 of 3.0040, 0.0182, 0.0108 s, respectively. Large tail reductions accompany only about 3.6\% improvement in mean update time. These are three runs, not 600 independent repeats. Background-delivery variants have matching input-trace hashes; their memory cost accompanies the main result.

\textbf{FFN policy and cost.} The World Prediction Tower caches FFN outputs in layers 1--18 and 31--36 at denoiser evaluations 2/4/6 and reuses them at 3/5/7 (layers numbered 1--36, evaluations 0--9). Other layers and evaluations 0/1/8/9 run exactly. This skips 144 of 720 dual-CFG FFN calls (20\%). Attention and the Action and Tactile Generation Tower still run. With separate BF16 buffers by layer and branch, each request uses $6{,}600\times4{,}096\times2\times24\times2$ bytes of cache (2.42 GiB), excluding allocator overhead and other activations. \figref{fig:infra:supplement}(b--d) and \tabref{tab:infra:inference-cost} report timing, open-loop error, and queue effects separately. Their speedups cannot be added.

\begin{table}[htbp]
  \centering\small
  \setlength{\tabcolsep}{4pt}
  \renewcommand{\arraystretch}{1.08}
  \caption{\textbf{Inference resource and quality costs.} Quality intervals resample episodes within each fixed task, preserving observation/seed dependence; only two episodes/task are available. Open-loop error does not establish task success.}\label{tab:infra:inference-cost}
  \begin{tabularx}{\linewidth}{@{}>{\raggedright\arraybackslash}p{.25\linewidth} >{\raggedright\arraybackslash}X >{\raggedright\arraybackslash}X@{}}
    \toprule
    \textbf{Quantity} & \textbf{Baseline / population} & \textbf{Alternative / change} \\
    \midrule
    CFG latency / GPU cost & One GPU\newline 2.004~s, 2.004 allocated~GPU-s/request & Two-GPU CFG parallelism\newline 1.054~s, 2.108 allocated~GPU-s/request \\
    FFN expert-error change & Equal-task mean; six-task episode-cluster~bootstrap & Translation $-0.51\%$; rotation $+3.42\%$ [2.98\%, 3.87\%]; gripper opening $+0.31\%$ \\
    FFN individual tail & Seventy-two generation pairs/metric & Gripper MAE: seven worsen >10\%; worst~$+21.03\%$ \\
    \bottomrule
  \end{tabularx}
\end{table}

\begin{table}[htbp]
  \centering\small
  \setlength{\tabcolsep}{4pt}
  \caption{\textbf{FlagGems operator benchmarks on trace-derived shapes.}
    VLM denotes vision--language modeling and T2V denotes text-to-video generation. In World Action Model (WAM) mode, the model generates action chunks.
    Launches and time shares refer to compute kernels on rank~0 after warmup.
    Speedups are reference-to-FlagGems latency ratios on the historical FlagTree/PyTorch~2.8.0a0 stack, with ranges across tested shapes where available.
    T2V used four WorldArena prompts per GPU on four GPUs ($736\times544$, 50 UniPC steps, CFG scale~5).
    Its attention count includes 7,200 setup launches. Representative query and KV shapes were $(55{,}131,32,128)$ and $(66{,}503,8,128)$.}\label{tab:infra:flaggems}
  \begin{tabularx}{\linewidth}{@{}lXrrrr@{}}
    \toprule
    \textbf{Workload} & \textbf{Operator} & \textbf{Launches} & \textbf{Time share} & \textbf{Speedup} & \textbf{Across shapes} \\
    \midrule
    VLM & Mean reduction & 4,640 & 3.443\% & 1.25$\times$ & 1.23--1.27$\times$ \\
        & Forward attention & 1,296 & 0.244\% & 1.424$\times$ & --- \\
        & Backward attention & 1,620 & 0.554\% & 1.397$\times$ & --- \\
        & Indexed write (\texttt{index\_put\_}) & 1,125 & 0.048\% & 2.170$\times$ & --- \\
    \midrule
    T2V & Forward attention & 14,400 & 34.043\% & 1.40$\times$ & 1.39--1.41$\times$ \\
        & Indexed write (\texttt{index\_put\_}) & 800 & 0.112\% & 2.97$\times$ & 2.95--3.00$\times$ \\
        & Sum reduction & 400 & $<0.001\%$ & 1.44$\times$ & 1.02--1.86$\times$ \\
        & Scatter add & 100 & 0.040\% & 1.007$\times$ & --- \\
    \midrule
    WAM  & Forward attention & 1,728 & 26.022\% & 1.425$\times$ & 1.38--1.47$\times$ \\
         & Fused RMSNorm (forward) & 864 & 0.236\% & 1.03$\times$ & 1.01--1.04$\times$ \\
         & Chunk and concatenate (\texttt{chunk\_cat}) & 222 & 0.052\% & 1.16$\times$ & 1.06--1.26$\times$ \\
    \bottomrule
  \end{tabularx}
\end{table}
\FloatBarrier

\textbf{Structured JSON generation for data captioning.} The FlagOS inference framework~\citep{flaggems2024} batches
grammar-mask computation during speculative decoding and uses optimized mixture-of-experts kernels.
Across the evaluated platforms, it achieved 1.4--2.2 times the per-device throughput of the corresponding baselines.

\subsection{FLOP Accounting}\phantomsection\label{app:infra:accounting}

The audited T2V decoder has 36 layers, width 4,096, 32 query/8 KV heads, head width 128, and FFN
width 12,288. For the audited online T2V run,
useful FLOPs include decoder forward/backward passes on real tokens and the required frozen VAE
forward pass. Activation recomputation and padding are excluded. Using rank-0 analytical FLOPs
$F_i$, complete-update time $t_i$, and dense BF16 device peak $P=989\times10^{12}$ FLOP/s,
\[
  \mathrm{MFU}_{\mathrm{useful}}=\frac{\sum_i F_{i,\mathrm{useful}}}{P\sum_i t_i}.
\]
\tabref{tab:infra:flop-accounting} reconstructs the result over updates 52--728. Only the last rank-0 microbatch is sampled and scaled by GA4; this is neither an all-rank average nor exhaustive microbatch accounting. Backward FLOPs use the analytical $2\times$ forward approximation. The numerator includes the online VAE forward pass required within the measured update; the Transformer-only result excludes it. Complete inter-update time includes data, communication, and optimizer overhead.

\begin{table}[htbp]
  \centering\small
  \setlength{\tabcolsep}{5pt}
  \caption{ \textbf{Useful-work MFU for the audited online T2V configuration in \tabref{tab:infra:configuration}: 677 updates, 35,350.814 s.} Both rows exclude recomputation and padding and are analytical estimates, not hardware-counter measurements.}\label{tab:infra:flop-accounting}
  \begin{tabularx}{\linewidth}{@{}Xrr@{}}
    \toprule
    \textbf{Required model work} & \textbf{FLOPs per GPU over window} & \textbf{MFU} \\
    \midrule
    Real-token Transformer forward/backward & $8.83882\times10^{18}$ & 25.28\% \\
    Above + online frozen VAE forward & $1.08533\times10^{19}$ & \textbf{31.04\%} \\
    \bottomrule
  \end{tabularx}
\end{table}

\FloatBarrier

\subsection{Correctness and Recovery}\phantomsection\label{app:infra:correctness}

\tabref{tab:infra:correctness} reports kernel and complete-update agreement. \tabref{tab:infra:recovery} reports saved-state restoration and numerical continuation. Candidate failures remain unresolved even when native repeats vary. Neither grouped FA3 nor ODC has established long-run training quality in these studies.

\begin{table}[htbp]
  \centering\small
  \setlength{\tabcolsep}{4pt}
  \renewcommand{\arraystretch}{1.08}
  \caption{\textbf{Numerical validation.} FA3 update audits sample layers 0/17/35 and input/output heads across all eight FSDP shards, with matched inputs/RNG and sampled initial parameters; the reported norms do not cover the full model.}\label{tab:infra:correctness}
  \begin{tabularx}{\linewidth}{@{}>{\raggedright\arraybackslash}p{.20\linewidth} >{\raggedright\arraybackslash}X >{\raggedright\arraybackslash}X@{}}
    \toprule
    \textbf{Comparison} & \textbf{Result} & \textbf{Interpretation} \\
    \midrule
    Grouped FA3 operator & Real QKV replay: maximum output/source-gradient relative L2 0.002299, below 0.03 & Operator agreement passes for the tested cases, including GQA, padding and source-gradient restoration \\
    FA3 native repeat & Clipped-gradient / update relative L2: 0.8798~/~0.6825 & Baseline update variability remains~unresolved \\
    FA3 split to grouped & Clipped-gradient / update relative L2: 1.301271~/~0.838888 & Update-level numerical acceptance not~passed \\
    ODC / native & Fixed-input gradient relative L2 1.2007\%; native repeat 1.2847\%; separate backend comparison 2.1583\% & Backend difference exceeds 2\% threshold; strict elementwise checks also fail \\
    \bottomrule
  \end{tabularx}
\end{table}

\begin{table}[htbp]
  \centering\small
  \setlength{\tabcolsep}{4pt}
  \renewcommand{\arraystretch}{1.08}
  \caption{\textbf{Recovery validation.} Saved cursors track consumed data, not completed prefetches. CP/TP/PP denote context, tensor and pipeline parallelism.}\label{tab:infra:recovery}
  \begin{tabularx}{\linewidth}{@{}>{\raggedright\arraybackslash}p{.22\linewidth} >{\raggedright\arraybackslash}X >{\raggedright\arraybackslash}X@{}}
    \toprule
    \textbf{Boundary} & \textbf{Evidence} & \textbf{Scope} \\
    \midrule
    Exact restoration & 10,149 saved tensor/metadata items per rank and nine input/plan/RNG records match their reference & Validated at the saved-state boundary; subsequent numerical equivalence not~established \\
    $160 \rightarrow 144$ GPUs & 58 sources \ensuremath{\times} 192 original logical shards =~11,136 states retained;\newline each of 144 ranks passes its first ten microbatch~sequence~checks & Offline ownership remapping; CP/TP/PP=1, unchanged fingerprints/processing.\newline Logical shards remain independent of~physical~rank~count \\
    Numerical continuation & Three-update sampled FP32 maximum difference $1.24\times10^{-5}$--$2.40\times10^{-5}$ & $10^{-5}$ threshold not met; in-memory rollback also fails \\
    \bottomrule
  \end{tabularx}
\end{table}

Resuming on fewer GPUs restores the saved boundary; work after that checkpoint may replay, and the
global batch or interleaving may change. Separate state-conditioning tests verify clean-input
routing and the distinction between input and loss masks; correcting this contract does not validate
previously trained weights.

\FloatBarrier
\section{Innata-0-Agent Details}

\subsection{Implementation of the Embodied Tool Set}
\label{app:toolset}

We implement a general and extensible embodied tool set shared by verification and recovery. The
tools are organized into four functional groups according to their role in closed-loop interaction:
\emph{Observation}, \emph{Perception}, \emph{Motion adjustment} and \emph{Atomic skills}.
Observation and perception tools do not advance the environment, whereas motion-adjustment tools and
atomic skills act through the native robot or simulator interface and consume simulation steps. The
full tool set is displayed in Table~\ref{tab:embodied_toolset}.

\begin{table*}[t]
  \centering
  \caption{\textbf{Embodied tool set.}
    The tools are shared by the simulators.}
  \label{tab:embodied_toolset}
  \small
  \setlength{\tabcolsep}{5pt}
  \renewcommand{\arraystretch}{1.12}
  \begin{tabularx}{\textwidth}{
      @{}
      >{\raggedright\arraybackslash}p{0.17\textwidth}
      >{\ttfamily\raggedright\arraybackslash}p{0.19\textwidth}
      >{\raggedright\arraybackslash}X
      @{}
    }
    \toprule
    \textbf{Category} & \textbf{Tool} & \textbf{Function} \\
    \midrule

    \multirow{7}{*}{Observation}
    & task\_info
    & Returns the task instruction, remaining execution budget, and current task context. \\

    & robot\_state
    & Reads the current arm, end-effector, and gripper states. \\

    & scene\_objects
    & Lists scene entities that have already been detected, localized, or measured. \\

    & rgb\_view
    & Acquires the raw RGB image from a specified camera. \\

    & depth\_view
    & Acquires depth observations from a specified camera. \\

    & scene\_view
    & Presents synchronized multiview observations to the agent, optionally with visual overlays. \\

    & camera\_info
    & Returns camera intrinsics, extrinsics, and pose information. \\
    \midrule

    \multirow{4}{*}{Perception}
    & target\_find
    & Localizes and segments a target specified by a semantic name, image point, or bounding box. \\

    & target\_judge
    & Estimates the target's three-dimensional center, spatial extent, orientation, and measurement uncertainty. \\

    & pixel\_unproject
    & Unprojects an image pixel into a three-dimensional point in the world coordinate frame. \\

    & point\_project
    & Projects a three-dimensional world point onto a specified camera image plane. \\
    \midrule

    \multirow{4}{*}{Motion Adjustment}
    & pose\_move
    & Moves the end-effector to a parameterized target pose. \\

    & arc\_move
    & Executes a parameterized arc motion around a specified rotation center and axis. \\

    & arc\_push
    & Maintains physical contact while pushing a control element along a constrained arc. \\

    & base\_move
    & Adjusts the position and orientation of the mobile base. \\
    \midrule

    \multirow{4}{*}{Atomic Skills}
    & gripper\_set
    & Opens or closes the gripper, or sets it to a specified opening width. \\

    & arm\_reset
    & Returns the robot arm to a predefined safe configuration. \\

    & motion\_hold
    & Maintains the current robot configuration for a bounded number of control steps. \\

    & rim\_grasp
    & Executes a specialized grasp using the estimated rim geometry of a container. \\

    \bottomrule
  \end{tabularx}
\end{table*}

\begin{figure*}[htpb]
  \centering
  \includegraphics[width=\linewidth]{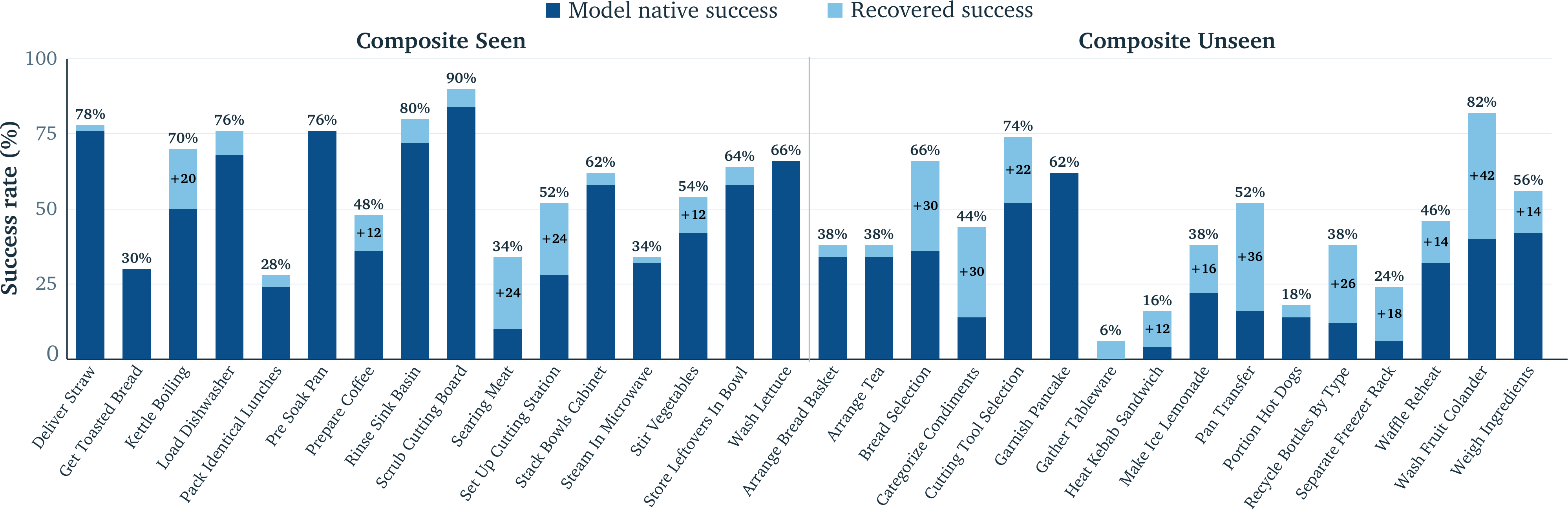}
  \caption{
    \textbf{Per-task recovery gains on the 32 RoboCasa365 composite tasks.}
    Each stacked bar reports the success rate of Innata-0. Dark blue denotes episodes completed natively by the Innata-0-WAM, while light blue denotes additional successes obtained through online recovery. Tasks are grouped into the Composite-Seen and Composite-Unseen splits.
    Percentages above the bars indicate the final success rate after recovery, and values inside the light-blue segments indicate the
    absolute percentage-point improvement contributed by recovery.
  }
  \label{fig:robocasa_per_task_recovery}
\end{figure*}

\subsection{Detailed Result Analysis}

\paragraph{Per-task recovery analysis}
Figure~\ref{fig:robocasa_per_task_recovery} shows that the benefits of recovery are strongly
task-dependent and concentrate on those challenging tasks for native Innata-0-WAM. In the
Composite-Seen split, recovery provides substantial gains on tasks such as \textit{SearingMeat} and
\textit{SetUpCuttingStation}. The improvements are even more pronounced on several Composite-Unseen
tasks, including \textit{BreadSelection}, \textit{CategorizeCondiments}, \textit{PanTransfer},
\textit{RecycleBottlesByType}, and  \textit{WashFruitColander}. In contrast, tasks with already
strong native performance, such as \textit{PreSoakPan}, \textit{ScrubCuttingBoard}, and
\textit{GarnishPancake}, require little or no recovery intervention. These results suggest that
recovery primarily compensates for WAM failures under difficult or unfamiliar conditions, rather
than uniformly altering otherwise successful executions.

\end{document}